\documentclass{article}
\usepackage[preprint, nonatbib]{nips_2018}
\usepackage[utf8]{inputenc} 
\usepackage[T1]{fontenc}    
\usepackage{hyperref}       
\usepackage{url}            
\usepackage{booktabs}       
\usepackage{amsfonts}       
\usepackage{nicefrac}       
\usepackage{microtype}      
\usepackage{multirow}
\usepackage[pdftex]{graphicx}
\graphicspath{{../pdf/}{../jpeg/}}
\DeclareGraphicsExtensions{.pdf,.jpeg,.png}
\usepackage[cmex10]{amsmath}
\usepackage{amssymb}
\usepackage{algorithmic}
\usepackage{array}
\usepackage{mdwmath}
\usepackage{mdwtab}
\usepackage{eqparbox}
\usepackage{caption}
\usepackage[font=footnotesize]{subfig}
\usepackage{url}
\usepackage{color}

\newcommand\T{\rule{0pt}{2.9ex}}       
\newcommand\B{\rule[-1.2ex]{0pt}{0pt}} 

\usepackage{cite}

\DeclareMathOperator\arctanh{arctanh}

\title{Exploring the Semantic Content of Unsupervised Graph Embeddings: An Empirical Study}

\author{
  Stephen Bonner\\
  Department of Computer Science\\
  Durham University\\
  Durham, UK \\
  \And
  Ibad Kureshi\\
  InlecomSystems\\
  Brussels, Belgium \\
  \And
  John Brennan\\
  Department of Computer Science\\
  Durham University\\
  Durham, UK \\
  \And
  Georgios Theodoropoulos\\
  School of Computer Science and Engineering\\
  SUSTech\\
  Shenzhen, China \\
  \And
  Andrew Stephen McGough\\
  School of Computing\\
  Newcastle University\\
  Newcastle, UK \\
  \And
  Boguslaw Obara\\
  Department of Computer Science\\
  Durham University\\
  Durham, UK \\
}

\begin{document}

\maketitle

\begin{abstract}
  Graph embeddings have become a key and widely used technique within the field of graph mining, proving to be successful across a broad range of domains including social, citation, transportation and biological. Graph embedding techniques aim to automatically create a low-dimensional representation of a given graph, which captures key structural elements in the resulting embedding space. However, to date, there has been little work exploring exactly which topological structures are being learned in the embeddings process. In this paper, we investigate if graph embeddings are approximating something analogous with traditional vertex level graph features. If such a relationship can be found, it could be used to provide a theoretical insight into how graph embedding approaches function. We perform this investigation by predicting known topological features, using supervised and unsupervised methods, directly from the embedding space. If a mapping between the embeddings and topological features can be found, then we argue that the structural information encapsulated by the features is represented in the embedding space. To explore this, we present extensive experimental evaluation from five state-of-the-art unsupervised graph embedding techniques, across a range of empirical graph datasets, measuring a selection of topological features. We demonstrate that several topological features are indeed being approximated by the embedding space, allowing key insight into how graph embeddings create good representations. 
\end{abstract}

\section{Introduction}

Representing the complex and inherent links and relationships between and within datasets in the form of a graph is a widely performed practised across scientific disciplines \cite{newman2010networks}. One reason for the popularity is that the structure or topology of the resulting graph can reveal important and unique insights into the data it represents. Recently, analysing and making predictions about graph using machine learning has shown significant advances in a range of commonly performed tasks over traditional approaches \cite{Goyal2017}. Such tasks include predicting the formation of new edges within the graph and the classification of vertices \cite{Moyano2017}. However, graph are inherently complex structures and do not naturally lend themselves as input into existing machine learning methods, many of which operate on vectors of real numbers.

Graph embeddings \footnote{In this work, we focus on vertex representation learning approaches.} are a family of machine learning models which learn latent representations for the vertices within a graph. The goal of all graph embedding techniques is broadly the same; to transform a complex graph, with no inherent representation in vector space, into a low-dimension vector representation of the graph or its elements. More concretely, the objective of a graph embedding technique is to learn some function $f:V \rightarrow \mathbb{R}^d$ which is a mapping from the set of vertices $V$ to a set of embeddings for the vertices, where $d$ is the required dimensionality of the resulting embedding. This results in $f$ being a matrix of dimensions $|V| $ by $ d$, i.e. an embedding of size $d$ for each vertex in the graph. It should be noted that this mapping is intended to capture the latent structure from a graph by mapping similar vertices together in the embedding space. Many of the recent approaches are able to produce low-dimensional graph representations without the need for labelled datasets. These representations can then be utilised as input to secondary supervised models for downstream prediction tasks, including classification \cite{Perozzi2014} or link prediction \cite{Grover2016}. Thus, graph embeddings are becoming a key area of research as they act as a translation layer between the raw graph and some desired machine learning model.

However, to date, there has been little research performed into why graph embedding approaches have been so successful. They all aim to capture as much topological information as possible during the embedding process, but how this is achieved, or even exactly what structure is being captured, is currently not known. In previous work \cite{bonner2017embedding}, we provided a framework which could be used to directly measure the ability of graph embeddings to capture a good representation of a graph's topology. In this paper, we expand upon this work by attempting to provide insight into the graph embedding process itself. We attempt to explore if the known and mathematically understood range of topological features \cite{newman2010networks} are being approximated in the embedding space. To achieve this, we investigate if a mapping from the embedding space to a range of topological features is possible. We hypothesise that if such a mapping can be found, then the topological structure represented by that feature, is also approximated in the embedding space. Such a discovery could begin to provide a theoretical framework for the use of graph embeddings, by experimentally demonstrating which topological structures are used to create the representations. Our methodology employs a combination of supervised and unsupervised models to predict topological features directly from the embeddings. The features we are investigating, first go through a binning process to transform them into classes to enable the classification. We make the following contributions whilst exploring this area:

\begin{itemize}
  \item We propose to investigate if graph embeddings are learning something analogous with traditional vertex level graph features. If this is the case, is there a particular type of feature which is being learned best. 

  \item We empirically show, to the best of our knowledge for the first time, that several known topological features are present in graph embeddings. This can be used to help explain the graph embedding process, by detailing which graph features are key in creating high quality representation. 

  \item We provide detailed experimental evidence, with five state-of-the-art unsupervised graph embeddings approaches, across seven topological features and six empirical graph datasets for our claims.

\end{itemize}

Reproducibility - we make all the experiments performed in this paper, reproducible by open-sourcing our code\footnote{https://github.com/sbonner0/unsupervised-graph-embedding/}, reporting key model hyper parameters and presenting results on public benchmark graph datasets. 

In Section \ref{sec:PW} we explore prior work, in Section \ref{sec:EEQ} we detail our approach for providing an experimental methodology for assessing know topological features approximated by graph embeddings, Section \ref{sec:E} details the experiment setup, in Section \ref{sec:R} we present our results and in Section \ref{sec:C} we conclude the paper along with suggesting further expansions of this work. 

\subsection{Notation}

We adopt here the commonly used notation for representing a graph or network\footnote{To avoid confusion with neural networks we will use the term graph throughout the remainder of the paper without loss of generality.} $G = (V,E)$ as an undirected graph which comprises a finite set of vertices (sometimes referred to as nodes) $V$ and a finite set of edges $E$. The elements of $E$ are an unordered tuple $\{u,v\}$ of vertices $u,v \in V$. Here $G$ could be a graph-based representation of a social, citation or biological network \cite{obara2012}. The adjacency matrix $\mathbf{A} = (a _i,_j)$ for a graph is symmetric matrix of size $|V|$ by $|V|$, where $(a _i,_j)$ is 1 if an edge is present and 0 otherwise.

\section{Previous Work}
\label{sec:PW}

This section explores the prior research regarding graph embedding techniques and previous approaches measuring known features in embeddings. We firstly introduce the notation of graph embeddings, detail supervised and factorization based approaches, explore in detail state-of-the-art unsupervised approaches which we be used throughout the rest of the paper and finally review past attempts to provide a theoretical understanding of there functionality.

\subsection{Graph Embeddings}

The ability to automatically learn some descriptive numerical based representation for a given graph is a attractive goal, and could provide a timely solution to some common problem within the field of graph mining. Traditional approaches have relied upon extracting features -- such as various measures of a vertices' centrality \cite{Page1998} -- capturing the required information about a graph's topology, which could then be used in some down-stream prediction task \cite{Li2012} \cite{bonner2016} \cite{Berlingerio2012} \cite{bonner2016gfp}. However, such a feature extraction based approach relies solely upon the hand-crafted features being a good representation of the target graph. Often a user must use extensive domain knowledge to select the correct features for a given task, with a change in task often requiring the selection of new features \cite{Li2012}. 

Graph embedding models are a collection of machine learning techniques which attempt to learn key features from a graph's topology automatically, in either a supervised or un-supervised manner, removing the often cumbersome task of end users manually selecting representative graph features \cite{Perozzi2014}. This process, known as feature selection \cite{guyon2003introduction} in the machine learning literature, has clear disadvantages as certain features many only be useful for a certain task. It even could negatively affect model performance if utilised in a task for which they are not well suited. Arguably, many of the recent exciting advances seen in the field of Deep Learning have been driven by the removal of this feature selection process \cite{Grover2016}, instead allowing models to learn the best data representations themselves \cite{Goodfellow2016}. For a selection of recent review papers covering the complete family of graph embedding techniques, readers are referred to \cite{Hamilton2017} \cite{cai2017} \cite{zhang2017network} \cite{cui2017survey}. The work presented in this paper focuses on neural network based approaches for graph embedding (as these have demonstrated superior performance compared with traditional approaches \cite{Goyal2017}).

The study of Neural Networks (NNs) is a field within machine learning inspired by the human brain \cite{Goodfellow2016}. NNs model problems via the use of connected layers of artificial neurons, where each network has an input layer, at least one hidden layer and an output layer. The activation of each neuron in a layer is given by a pre-specified function, with each neuron taking a weighted sum of all the outputs of those neurons to which it is connected. These weights are learned through training examples which are fed through the network, with modifications made to the weights via back-propagation to increase the probability of the NN producing the desired result \cite{Goodfellow2016}. 

\subsubsection{Supervised Approaches}

Within the field of machine learning, approaches which are supervised are perhaps the most studied and understood \cite{Goodfellow2016}. In supervised learning, the datasets contain labels which help guide the model in the learning process. In the field of graph analysis, these labels are often present at the vertex level and contain, for example, the meta-data of a user in a social network. 

Perhaps the largest area of supervised graph embeddings is that of Graph Convolutional Neural Networks (GCNs) \cite{Bruna2013}, both spectral \cite{Defferrard2016} \cite{kipf2017semi} and spatial \cite{NiepertMATHIASNIEPERT2016} approaches. Such approaches pass a sliding window filter over a graph, in a manner analogous with Convolutional Neural Networks from the computer vision field \cite{Goodfellow2016}, but with the neighbourhood of a vertex replacing the sliding window. Current GCN approaches are supervised and thus require labels upon the vertices. This requirement has two significant disadvantages: Firstly, it limits the available graph data which can be used due to the requirement for labelled vertices. Secondly, it means that the resulting embeddings are specialised for one specific task and cannot be generalised for a different problem without costly retraining of the model for the new task. 

\subsubsection{Factorization Approaches}

Before the recent interest in learning graph embeddings via the use of neural networks, a variety of other approaches were explored. Often these approaches took the form of matrix factorization, in a similar vain to classical dimensionality reduction techniques such as Principal Competent Analysis (PCA) \cite{Hamilton2017} \cite{wold1987principal}. Such approaches first calculate the pair wise similarity between the vertices of a graph, then find a mapping to a lower dimensional space, such that the relationships observed in the higher dimensions are preserved. An early example of such an approach is that of the Laplican eigenmaps, which attempts to directly factorize the Laplacian matrix of a given graph \cite{belkin2002laplacian}. Other approaches, often using the adjacency matrix, define the relationship in low dimension space between two vertices in the graph as being determined by the dot product of their corresponding embeddings. Such approaches include Graph Factorization \cite{ahmed2013distributed}, GraGrep \cite{cao2015grarep} and HOPE \cite{ou2016asymmetric}. Such dimensionality reduction based approaches are often quadratic in complexity \cite{zhang2017network} and the predictive performance of the embeddings has largely been superseded by the recent neural network based methods \cite{Goyal2017}.

\subsection{Unsupervised Stochastic Embeddings}
\label{sec:randomwalk}

DeepWalk \cite{Perozzi2014} and Node2Vec \cite{Grover2016} are the two main approaches for random walk based embedding. Both of these approaches borrow key ideas from a technique entitled Word2Vec \cite{Mikolov2013} designed to embed words, taken from a sentence, into vector space. The Word2Vec model is able to learn an embedding for a word by using surrounding words within a sentence as targets for a single hidden layer neural network model to predict. Due to the nature of this technique, words which co-occur together frequently in sentences will have positions which are close within the embedding space. The approach of using a target word to predict neighbouring words is entitled Skip-Gram and has been shown to be very effective for language modelling tasks \cite{Mikolov2013b}.

\subsubsection{DeepWalk}

The key insight of DeepWalk is to use random walks upon the graph, starting from each vertex, as the direct replacement for the sentences required by Word2Vec. A random walk can be defined as a traversal of the graph rooted at a vertex $v_t \in V$, where the next step in the walk is chosen uniformly at random from the vertices incident upon $v_t$ \cite{Backstrom2010}, these walks are recorded as $w_0^{t}, ..., w_n^{t}$ (where $t$ is the walk starting from $v_t$ of length $n$, and $w_i^{t} \in V$), i.e. a sequence of the vertices visited along the random walk starting from $v_t = w_0^t$. DeepWalk is able to learn unsupervised representations of vertices by maximising the average log probability $\mathbf{P}$ over the set of vertices $V$:
\begin{equation}
  \label{eq:logprob}
 \frac{1}{|V|} \displaystyle\sum\nolimits_{t=1}^{|V|} \ \ \sum\nolimits_{i=0}^n \sum\nolimits_{-c \leq j \leq c, j \neq 0} \log \mathbf{P}(w_{i+j}^t | w_i^t),
\end{equation}
where $c$ is the size of the training context of vertex $w_n^t$.\footnote{Note if $i+j < 0$ then we skip these from the sum we are past the start of the current work.}

The basic form of Skip-Gram used by DeepWalk defines the conditional probability $\mathbf{P}(w_{i+j}^t | w^t_i)$ of observing a nearby vertex $w_{i+j}^t$, given the vertex $w_i^t$ from the random walk $t$, can be defined via the softmax function over the dot-product between their features \cite{Perozzi2014}:
\begin{equation}
  \label{eq:softmax}
 \mathbf{P}(w_{i+j}^t | w^t_i) = \frac{ \exp{ (\mathbf{W}_{w^t_i}^\intercal \mathbf{W}_{w_{i+j}^t}^\prime) }}
 {\sum_{t=1}^{|V|} \exp{ (\mathbf{W}_{w^t_i}^\intercal \mathbf{W}_{v_t}^\prime) }  },
\end{equation}
where $\mathbf{W}_{w^t_{i}}$ and $\mathbf{W}_{w_{i+j}^t}^\prime$ are the hidden layer and output layer weights of the Skip-Gram neural network respectively.

\subsubsection{Node2Vec}

Whilst DeepWalk uses a uniform random transition probability to move from a vertex to one of its neighbours, Node2Vec biases the random walks. This biasing introduces two user controllable parameters which dictate how far from, or close to, the source vertex the walk progresses. This is done to capture either the vertex's role in its local neighbourhood (homophily), or alternatively its role in the global graph structure (structural equivalence) \cite{Grover2016}. Changing the random walk means that Node2Vec has a higher accuracy over DeepWalk for a selection of vertex classification problems \cite{Grover2016}. 

\subsection{Unsupervised Hyperbolic Embeddings}

Recently, a new family of graph embedding approaches has been introduced which embed vertices into hyperbolic, rather than Euclidean space \cite{Nickel2017} \cite{Chamberlain2017}. Hyperbolic space has long been used to analyse graphs which exhibit high levels of hierarchical or community structure \cite{munzner1998}, but it also has properties which could make it an interesting space for embeddings \cite{Chamberlain2017}. Hyperbolic space can be considered ``larger'' than Euclidean with the same number of dimensions, as the space is curved, its total area grows exponentially with the radius \cite{Chamberlain2017}. For graph embeddings, this key property means that one effectively has a much larger range of possible points into which the vertices can be embedded. This property allows for closely correlated vertices to be embedded close together, whilst also maintaining more distance between disparate vertices, resulting in an embedding which has the potential to capture more of the latent community structure of a graph.

The hyperbolic approach we focus on was introduced by Chamberlain \cite{Chamberlain2017}, and uses the Poincar\'{e} Disk model of 2D hyperbolic space \cite{epstein1988}. In their model, the authors use polar coordinates $ x = (r,\theta)$, where $r \in [0,1]$ and $\theta \in [0, 2 \pi]$ to describe a point in space for each vertex $v$ in the Poincar\'{e} Disk, which allows for the technique to be significantly simplified \cite{Chamberlain2017}. Similar to DeepWalk, an inner-product is used to define the similarity between two points within the space. The inner-product of two vectors in a Poincar\'{e} Disk can be defined as follows \cite{Chamberlain2017}:
\begin{equation}
  \label{eq:hyper}
<x, y> = ||x|| ||y||\cos(\theta_x-\theta_y),
\end{equation}
\begin{equation}
  \label{eq:hyper2}
  =4\,\arctanh\,r_x\,\arctanh\, r_y\, \cos(\theta_x-\theta_y),
\end{equation}
where $x=(r_x,\theta_x)$ and $y=(r_y,\theta_y)$ are the two input vectors representing two vertices and \emph{arctanh} is the inverse hyperbolic tangent function \cite{Chamberlain2017}. 

To create their hyperbolic graph embedding, the authors use the softmax function of Equation \ref{eq:softmax}, used by DeepWalk and others, but importantly replacing the Euclidean inner-products with the hyperbolic inner-products of Equation \ref{eq:hyper}. Aside from this, hyperbolic approaches share many similarities with the stochastic approaches in regards to their input data and training procedure. For example, the hyperbolic approaches are still trained upon pairs of vertex IDs, taken from sequences of vertices generated via random walks on graphs.

\subsection{Unsupervised Auto-Encoder Based Approaches}

A different approach for graph embeddings which does not use random walks for input, is entitled Structural Deep Network Embedding (SDNE) \cite{Wang2016a}. Instead of a technique based upon capturing the meaning of language, SDNE is designed specifically for creating graph embeddings using Deep Learning \cite{Goodfellow2016} -- deep auto-encoders \cite{Hinton2011}. Auto-encoders are an un-supervised neural network, where the goal of the technique is to accurately reconstruct the input data through explicit encoder and decoder stages \cite{Salakhutdinov2009}.

The authors of SDNE argue that a deep neural network, versus the shallow Skip-Gram model used by both DeepWalk and Node2Vec, is much more capable of capturing the complex structure of graphs. In addition the authors argue that for a successful embedding, it must capture both the first and second order proximity of vertices. Here the first order proximity measures the similarity of the vertices which are directly incident upon one another, whereas the second order proximity measures the similarly of vertices neighbourhoods. To capture both of these elements SDNE has a dual objective loss function for the model to optimise. The input data to SDNE is the adjacency matrix $\mathbf{A}$, where each row $a$ represents the neighbourhood of a vertex.

The objective function for SDNE comprises two distinct terms, the first term captures the second order proximity of the vertices neighbourhood, whilst the second captures the first order proximity of the vertices by iterating over the set of edges $E$:

\begin{equation}
 \label{eq:finalloss}
L_{SDNE} = \sum_{i=1}^{|V|} ||(q_i^\prime - q_i) \odot b_i||_2^2 +\alpha \sum_{u,v=1}^{|E|} a_{u,v}|| (w_u^{(k)} - w_v^{(k)}) ||_2^2,
\end{equation}

where $q_i$ and $q_i^\prime$ are the input and reconstructed representation of the input, $\odot$ is the element wise Hadamard product and $b_i$ is a scaling factor to penalise the technique if it predicts zero too frequently, $w^{(k)}$ is the weights of the $k^{th}$ layer in the auto-encoder technique and where $\alpha$ is a user-controllable parameter defining the importance of the second term in the final loss score \cite{Wang2016a}.

To initialise the weights of the deep auto-encoder used for this approach, an additional neural network must be trained to find a good starting region for the parameters. This pre-training neural network is called a Deep Belief Network, and is widely used within the literature to form the initialisation step of deeper models \cite{erhan2010}. However, this pre-training step is not required by either the stochastic or hyperbolic approaches as random initialisation is used for the weights, and adds significant complexity. 

\subsection{Observing Features Preserved in Embeddings}

\subsubsection{Graph Embeddings Features}

To date, there has been little research performed exploring a theoretical bases as to why graph embeddings are able to demonstrate such power in graph analytic tasks, or if something approximating traditional graph features are being captured during the embeddings process. Recently Goyal and Ferrar~\cite{Goyal2017} presented a experimental review paper on a selection of graph embedding techniques. The authors use of range of tasks including vertex classification, link prediction and visualization to measure the quality of the embeddings. However the authors do not provide any theoretical basis as to why the embedding approaches they test are successful, or if know features are present in the embeddings. In addition, the authors do not consider embeddings taken from promising unsupervised techniques -- such as the family of hyperbolic approaches, nor; do they explore performance across imbalanced classes during the classification.

Some recent work has speculated on the use of a graph's topological features as a way to improve the quality of vertex embeddings by incorporating them into a supervised GCN based model \cite{Hamilton2017a}. They show how aggregating a vertex feature -- even one as simple as its degree -- can improve the performance of their model. Further, they present theoretical analysis to validate that their approach is able to learn the number of triangles a vertex is part of, arguing that this demonstrates the model is able to learn topological structure. We take inspiration from this work, but consider unsupervised approaches as well as exploring if richer and more complicated topological features are being captured in the embedding process. In a similar vain, an approach for generating supervised graph embeddings using heat-kernel based methods is validated by visualizing if a selection a topological features can be seen in a two-dimensional projection of the embedding space \cite{li2016deepgraph}.

Research has investigated the use of a graph's topological features as a way of validating the accuracy of a neural network based graph generative model \cite{Liu2017}. With the presented model, the authors aim to generate entirely new graph datasets which mimic the topological structure of a set of target graphs -- a common task within the graph mining community \cite{Albert2002}. To validate the quality of their model, they investigate if a new graph created from their generative procedure has a similar set of topological features to the original graph. 

Perhaps most closely related to our present research is work exploring the use of random walk based graph embeddings as an approximation for more complex vertex level centrality measures on social network graphs \cite{salehi2017}. The authors argue that graph embeddings could be used as a replacement for centrality measures as they potentially have a lower computational complexity, thus taking less time to compute. The work explores the use of linear regression to try to directly predict four centrality measures from the vertices of three graph datasets, with limited success \cite{salehi2017}. Our own work differs significantly as we attempt to provide insight into what exactly graph embeddings are learning with a view to explain there success, explore a wider range of embeddings approaches, use datasets from a wider range of domains, explore more topological features, use classification rather than regression as the basis for the analysis and address the inherent unbalanced nature of most graph datasets.

\subsubsection{Feature Learning in Other Domains}

A large quantity of the successful unsupervised graph embedding approaches have adapted models originally designed for language modelling \cite{Grover2016} \cite{Perozzi2014}. Some recent research investigated how best to evaluate a variety of unsupervised approaches for embedding words into vectors \cite{schnabel2015}. They choose a variety of Natural Language Processing (NLP) tasks, which capture some known and understood aspects of the structure of language, and investigate how well the chosen embedding models perform for these tasks. They conclude that no single word embedding model performs the best across all the tasks they investigated, suggesting there is not a single optimal vector representation for a word. What features are used to help word embeddings achieve compositionality --  constructing the meaning of an entire sentence from the component words, has also been explored \cite{li2015visualizing}. Further research has investigated the use of word embeddings to create representations for the entire sentence using word features \cite{conneau2017}. The work suggests that word features learned by the embeddings for natural language inference can be transferred to other tasks in NLP. 

Outside of NLP, there has been work in the field of Computer Vision (CV) investigating what known features, already commonly used for image representation, are captured by deep convolutional neural network - potentially being used to explain how they work. For example, it has been shown that convolotuional networks, when trained for image classification, often detect the presence of edges in the images \cite{zeiler2014}. The same work also shows how the complexity of the detected edges increases as the depth of the network increases. 

In this present work, we take inspiration from these approaches and attempt to provide insight and a potential theoretical basis for the use of graph embeddings by exploring which known graph features can be reconstructed from the embedding space.

\section{Semantic Content of Graph Embeddings} 
\label{sec:EEQ}

Despite extensive prior work in unsupervised graph embedding, performing well for the tasks they were proposed for (such as vertex classification and link prediction \cite{Goyal2017}), there has been little work in exploring why these approaches are successful. Inspired by recent work in Computer Vision and Natural Language Processing which examine if traditional features (The edges detected in images for example) are captured by deep models, we explore, in this paper, the following research question: 

\textbf{Problem Statement - } \emph{Are graph embedding techniques capturing something similar to traditional topological features as part of the embedding process?} 

Topological features are a known and mathematically understood way to accurately identify graphs and vertices \cite{Li2012} \cite{bonner2016}. We hypothesis that if graph embeddings are shown to be learning approximations of existing features, than this could be used to begin to provide a theoretical basis for the functionality of graph embeddings. We hypothesise that if topological structures within a graph, similar to the known and mathematically understood examples from the literature, are being captured in the embedding space, then this could be used to begin to provide a theoretical basis for the functionality of graph embeddings. This would suggest that graph embedding are automatically learning detailed and known graph structures in order to create the representations. This could explain how they have been so successful in a variety of graph mining tasks. Effectively the graph embeddings techniques would be acting as an automated way of selecting the most representative topological feature(s) for a given objective function. 

If graph embeddings are shown to be learning topological features, then other interesting research questions arise. For example, do competing embedding approaches learn different topological structures, do different graph datasets each require different features to be approximated in order to create a good representation, what is the structural complexity of the features approximated by the embeddings or even are the embeddings capable of approximating multiple features simultaneously.

In order to explore these questions, we attempt to predict a selection of topological features directly from graph embeddings computed from a range of state-of-the-art approaches across a series of empirical datasets. We suggest that if a second mapping function $f:\mathbb{R}^d \rightarrow \Lambda$ can be found which accurately maps the embedding space to a given topological feature $\Lambda$, then this is strong evidence that something approximating the structural information represented by $\Lambda$ is indeed present in the embedding space. Here the mapping function could take the form of a linear regression, but for this work we investigate a range of classification algorithms -- this is explored more in Section \ref{sec:methodology}. We assess a range of known topological features, from simple to complex, to gain a better understanding of the expressive capabilities of the embedding techniques. 



\subsection{Predicting Topological Features}
\label{sec:topoligicalfeatures}

Numerous topological features have been identified in the literature, measuring various aspects of a graph's topology, at the vertex, edge and graph level \cite{Li2012}. As we are focusing our work here upon methods for creating vertex embedding, we will focus on features which are measured at the \emph{vertex level} of a given graph. We have selected a range of vertex level features from the graph mining literature, which capture information about a vertex's local and global role within a graph \cite{Grover2016}. This selection of features, range from ones which are simple to compute from vertices directly adjacent to each other, to more complex features which can require information from many hops\footnote{Hops represent the length of the sequences of vertices that must traversed to get from vertices $i$ to $j$.} further along with the graph. This will allow us to explore if embedding models learn complex topological features, or are they able to learn good representations of only simple features. The topological vertex level features we are predicting are detailed below, listed approximately by their complexity:

These features are defined in terms of a graph $G=(V,E)$ with it's corresponding adjacency matrix $\mathbf{A}$, where $|V|$ is the total number of vertices in the graph, $|E|$ the total number of edges. For each vertex $v \in V$, we also define $d(v)$ to be the total number of neighbours for $v$, $d^+(v)$ to be the number of connections $v$ has to other vertices, $\Gamma^{-}(v)$ to be the subset of vertices in $V$ with edges to $v$ and $\sigma_{st}(v)$ is the total number of shortest paths from vertices $s$ and $t$ which also pass through $v$.

\textbf{Total Degree} $DG(v) = d(v)$: The total number of edges from $v$ to other vertices.

\textbf{Degree Centrality} $DC(v) = \frac{1}{|V|} d(v)$: The degree for the vertex $v$ over the total number of vertices in the graph, providing a normalised centrality score \cite{bonner2016}.

\textbf{Number Of Triangles} $TC(v) = \Phi$: The number of triangles containing the vertex $v$, where $\Phi$ is the number of vertices in $\Gamma^{-}(v)$ which are also connected via an edge \cite{bonner2016}. 

\textbf{Local Clustering Score} $ CLU(v) =\frac{ 2\Phi }{ d(v)(d(v) - 1) }$: Represents the probability of two neighbours of $v$ also being neighbours of each other \cite{Watts1998}.

\textbf{Eigenvector Centrality} $EC(v) = \mathbf{A}\mathbf{x} = \lambda\mathbf{x}$: Used to calculate the importance of each vertex within a graph, where $\lambda$ is the largest eigenvalue and $ \mathbf{x}$ is the eigenvector centrality \cite{bonacich2007}.

\textbf{PageRank Centrality} $PR(v) = \frac{1 - \Omega}{|V|} + \Omega \displaystyle\sum_{ u \in \Gamma^{-}(v) } \frac{ PR(u) }{ d^+(u) }$: PageRank centrality is commonly used to measure the local influence of a vertex within a graph \cite{Page1998} \cite{Han2014}. Where  $\Omega$ is a constant damping factor (0.85 for this work).

\textbf{Betweenness Centrality} $BC(v)= \sum_{s \neq v \neq t \in V \atop s \neq t} \frac{\sigma_{st}(v)}{\sigma_{st}}$: The Betweenness centrality of a vertex depends upon the frequency which it acts as a bridge between two additional vertices \cite{Han2014}, where $\sigma_{st}$ is the total number of shortest paths from $s$ to $t$.

\subsection{Power-Law Feature Distribution}
\label{sec:feature-imbalance}

\begin{figure*}
  \centering
  \subfloat[Degree]{%
  \includegraphics[width=0.25\linewidth]{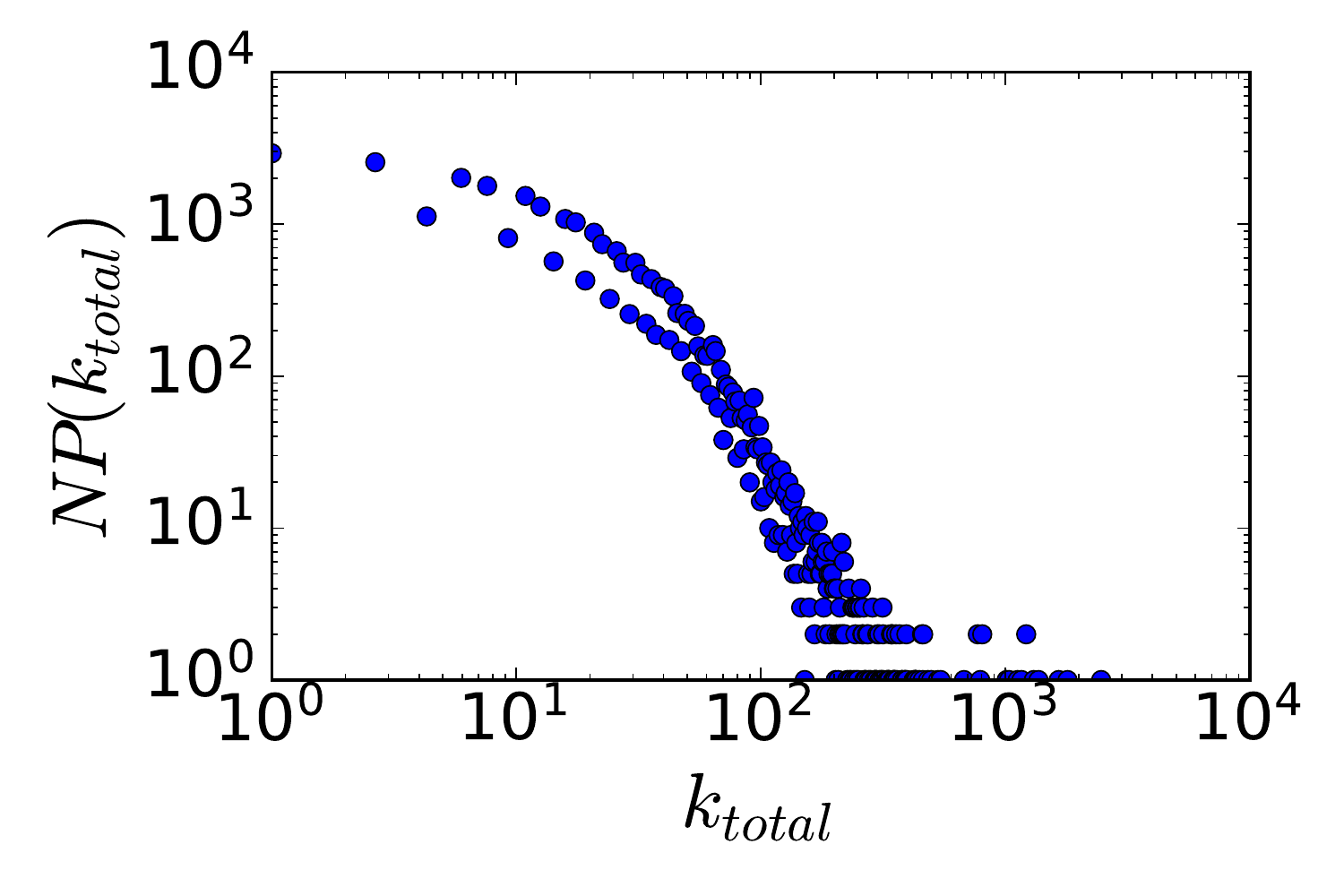}}
\label{DEG}\hfill
  \subfloat[Triangle Count]{%
     \includegraphics[width=0.25\linewidth]{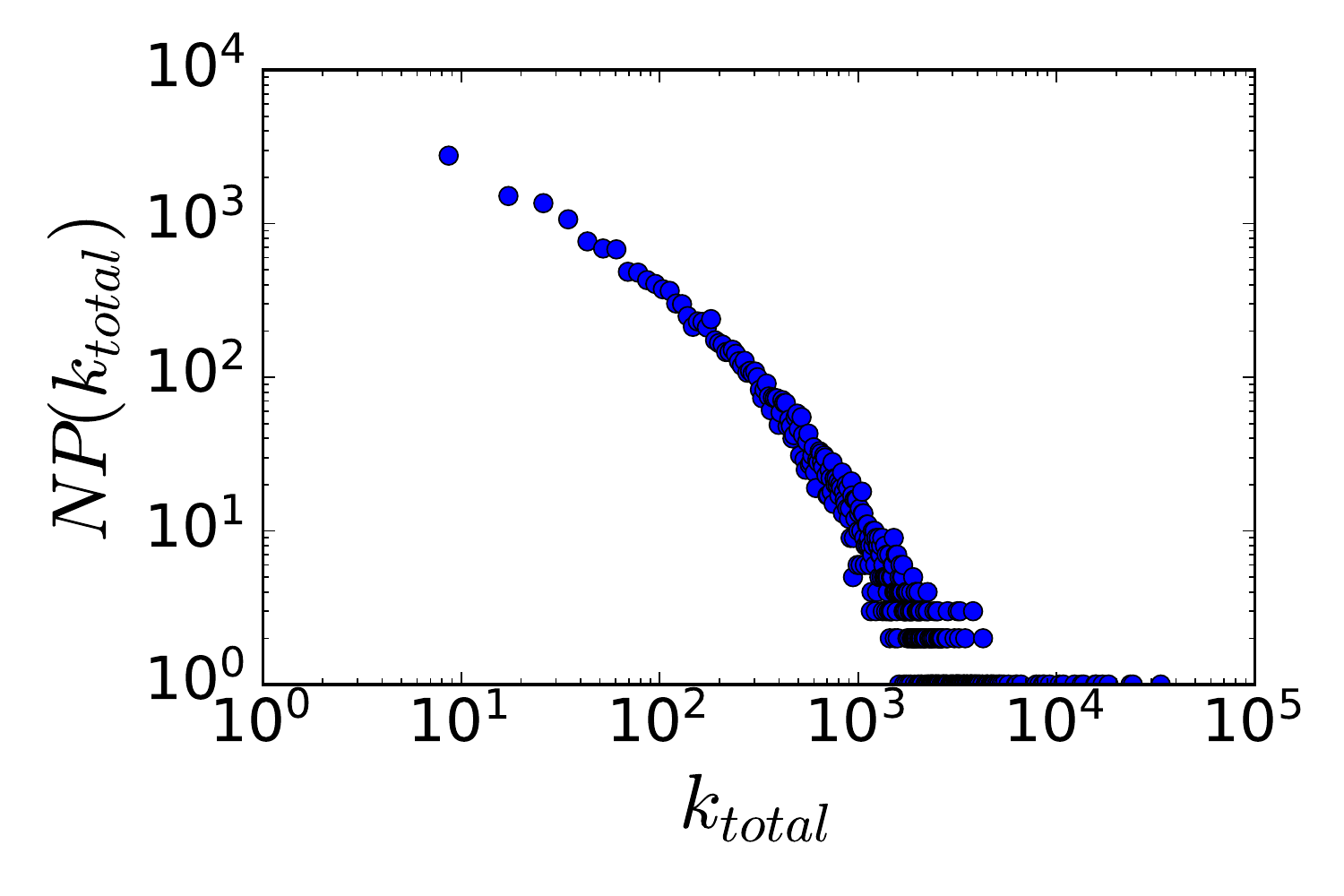}}
  \label{TC}\hfill
  \subfloat[Eigenvector Cent]{%
      \includegraphics[width=0.25\linewidth]{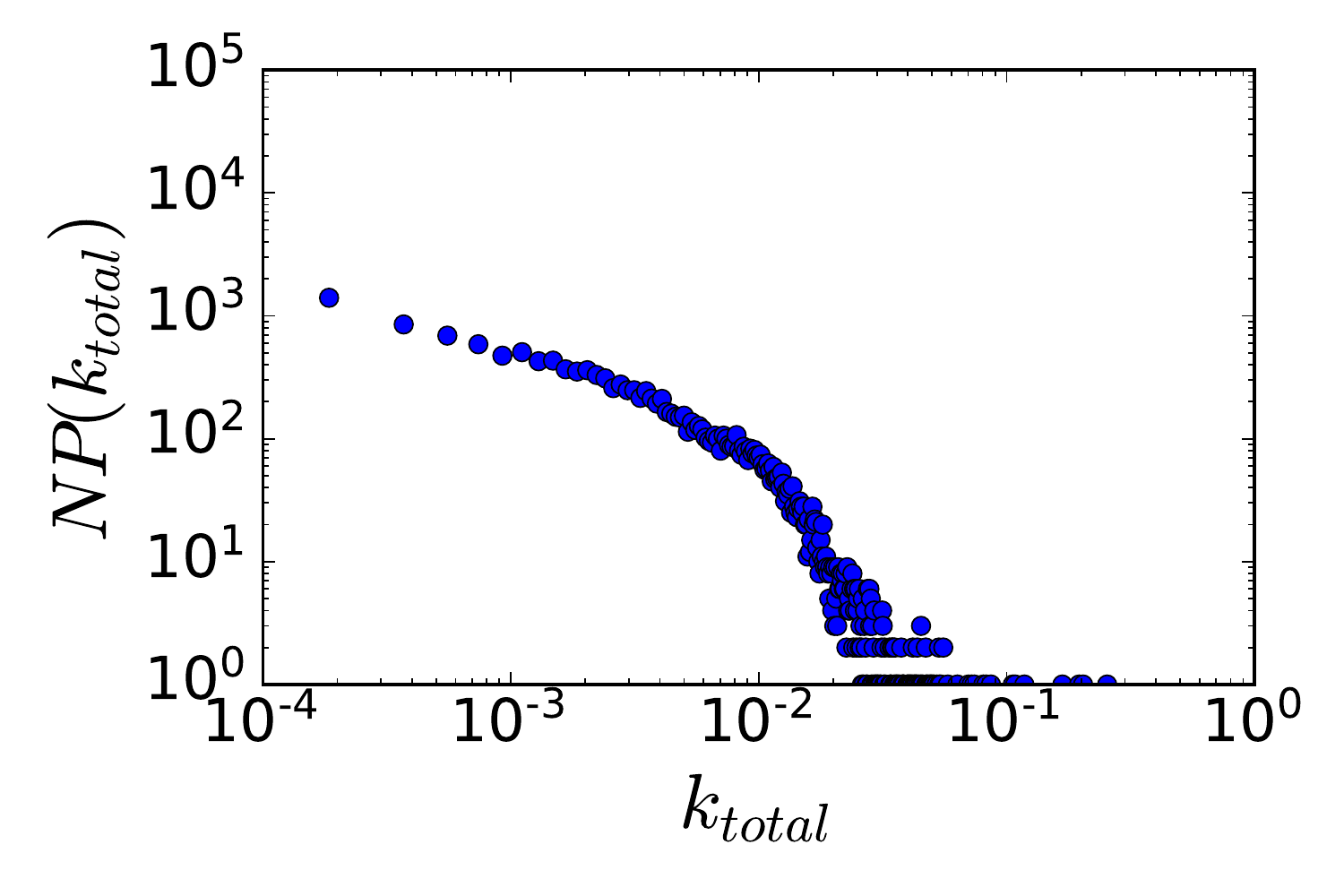}}
  \label{EC}\hfill
  \subfloat[Betweenness Cent]{%
      \includegraphics[width=0.25\linewidth]{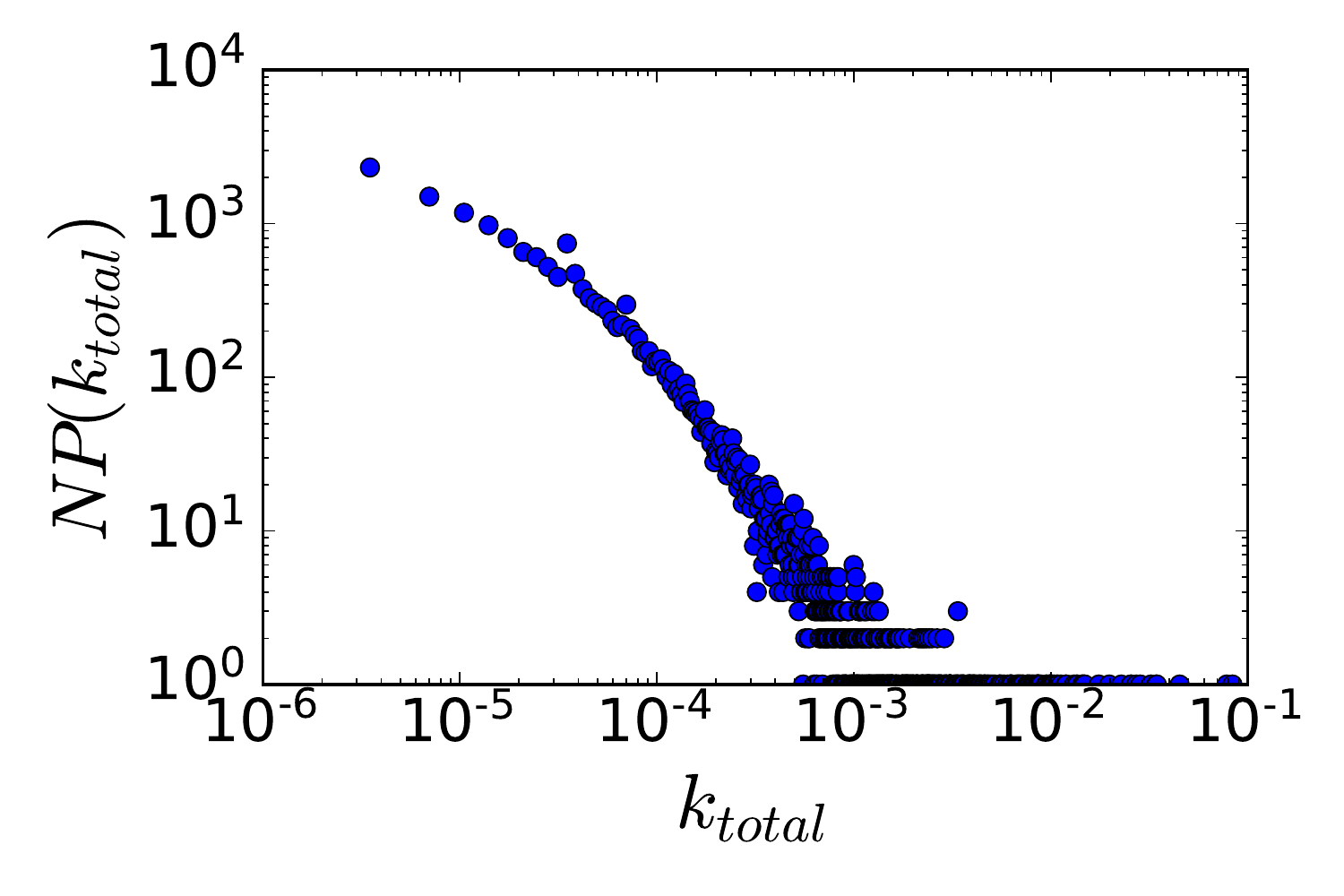}}
   \label{BC} 
  \caption{Distribution of topological feature values from the cit-HepTh dataset in log scale: (a) total vertex degree distribution, (b) distribution complete triangles for each vertex, (c) Eigenvector centrality distribution and (d) Betweenness centrality score distribution.}
  \label{fig:powerlaw} 
  \end{figure*}

Many empirical graphs, especially those representing social, hyper-link and citation networks, have been shown to have an approximately power-law distribution of degree values \cite{faloutsos1999}. This power-law distribution poses a challenge for machine learning models, as it means the features we are trying to predict are extremely unbalanced, with a heavy skew towards the lower range of features. Imbalanced class distribution creates difficulties for machine learning models, as there are fewer examples of the minority classes for the model to learn, which can often lead to poor predictive performance on these classes \cite{Goodfellow2016}. It has been shown that the distribution of other topological features can also follow a power-law distribution in many graphs \cite{Albert2002}. To demonstrate this phenomenon, Figure \ref{fig:powerlaw} shows the distribution of a range of topological feature values for the cit-HepTh dataset. The Figure shows that indeed, all the topological feature values tested largely follow an approximately power-law distribution. This fact has the potential to make predicting the value of a certain topological feature challenging, as the datasets will not be balanced and any model attempted to find the mapping $f:\mathbb{R}^d \rightarrow \Lambda$, will be prone to over fitting to the majority classes. Our approach for tackling this issue is outlined in the following section. 

\subsection{Methodology}
\label{sec:methodology}

Unlike previous studies \cite{salehi2017}, we use classification, rather than regression, as a way to explore the embedding space. Predicting topological features directly via the use of regression has proven challenging in prior work \cite{salehi2017}, owing largely to the imbalance problem explored in the previous section. With such an imbalanced dataset, using a classification based approach is often advantageous \cite{oord2016} as techniques exist to over sample minority examples. However, the features we are attempting to predict are continuous, so must go through some transformation stage before classification can be performed. For our transformation stage, we follow a procedure similar to \cite{oord2016}. We bin the real-valued features into a series of classes via the use of a histogram, where the bin a particular features is placed becomes it class label. One can consider each of these newly created classes as representing a range of possible values for a given feature. As an example, we could transform a vertex's continuous PageRank score \cite{Page1998} into a series of discrete classes via the use of a histogram with a bin size of three, where each of the newly created classes represented a low, medium or high PageRank score. 

In order to allow for a good distribution of feature values, for our experiments we utilise a bin size of six for the histogram function, meaning that six discrete classes were created for each of the features we are exploring. This value was chosen empirically from our datasets as it fully covered the numerical range of the topological features we were measuring. Although this binning process helps with the feature imbalance, it still produces a skew in number of features assigned to each class. To further address this issue, we take the logarithm of each feature value before it is passed to the binning function. Essentially, this will mean that features within the same order of magnitude will be assigned the same label, for example vertices with degrees in the range of 0 to $10^1$ would be assigned into one class, whilst degree values between $10^2$ to $10^3$ would be assigned to another class. This was performed as it dramatically improved the balance of the datasets, and as we are only attempting to discover if something approximating the topological features is present in the embedding space, we found that predicting the order of magnitude to be sufficient. 

\subsection{Embedding Approaches Compared}
\label{sec:approachescompared}

In this paper, we evaluate five state-of-the-art unsupervised graph embedding approaches as a way of exploring what semantic content is extracted from graph to create it's embedding. The approaches are as follows: DeepWalk, Poincar\'{e} Disk, Structural Deep Network Embedding and Node2Vec \footnote{Please note, we explore two variations of Node2Vec, bringing the total number of approaches to five}, which are detailed in Table \ref{tab:approaches}. These approaches were chosen as they represent a good cross-section of the current competing methodologies and all either exploit a different method of sampling the graph, use different geometries for the embedding space or use competing methods of comparing vertices. This selection of approaches will allow explorations of interesting research questions. Such questions include if any differences between the approaches can be explained by what graph structures they learn and do methods which promote local exploration around the vertices, mean that only vertex features which capture local structural information are present in the resulting embeddings. To explore this second question in more detail, we actually created two versions of Node2Vec: Firstly, Node2Vec-Structural, which biases the random walks used to create training pairs for the model to explore vertices further away from the target vertex and Node2Vec-Homophily, which biases the random walks to stay closer to the target vertex. 

\begin{table}[!h]
  \centering
  \begin{tabular}{c c c c c}
  \hline
  \textbf{Approach} &  \textbf{Year} & \textbf{Type} & \textbf{Published} & \textbf{Complexity}\\
  \hline \hline

  DeepWalk & 2014 & stochastic & KDD \cite{Perozzi2014} & $O(|V|)$\\

  Node2Vec & 2016 & stochastic & KDD \cite{Grover2016} & $O(|V|)$\\

  SDNE & 2016 & auto-encoder & KDD \cite{Wang2016a} & $O(|V||E|)$\\

  Poincar\'{e} Disk & 2017 & hyperbolic & MLG \cite{Chamberlain2017} & $O(|V|)$\\
  \hline
  \end{tabular}
  \caption{Graph Embedding Approaches being Compared.}
  \label{tab:approaches}
\end{table}

\section{Experimental Setup and Classification Algorithm Selection} 
\label{sec:E}

In the following section we detail the setup of the experiments and evaluate potential classification algorithms.

\subsection{Metrics}

\subsubsection{Presented Results} 

All the reported results are the mean of five replicated experiment runs along with confidence intervals. For the runtime analysis, the presented results are the mean runtime for job completion, presented in minutes. For the classification results, all the accuracy scores presented are the mean accuracy after $k$-fold cross validation -- considered the gold standard for model testing \cite{Arlot2010}. For $k$-fold cross validation, the original dataset is partitioned into $k$ equally sized partitions. $k-1$ partitions are used to train the model, with the remaining partition being used for testing. The process is repeated $k$ times using a unique partition for each repetition and a mean taken to produce the final result.

\subsubsection{Precision Metrics}
\label{sec:precision}

For reporting the results of the vertex feature classification tasks, we report the macro-f1 and micro-f1 scores with varying percentages of labelled data available at training time. This is a similar setup to previous works \cite{Grover2016} \cite{Goyal2017}. 

The micro-f1 score calculates the f1-score for the dataset globally by counting the total number of true positives (TP), false positives (FP) and false negatives (FN) across a labelled dataset $|L|$. Using the notation from \cite{Goyal2017}, micro-f1 is defined as:
\begin{equation}
  \label{eq:microf1}
 microf1 = \frac{2 \cdot P \cdot R}{P + R},
\end{equation}
\noindent where:
$$
 Precision (P) = \frac{ \sum_{l=1}^{|L| } TP(l) } {\sum_{l=1}^{|L|}TP(l)+FP(l) },
$$
$$
 Recall (R) = \frac{ \sum_{l=1}^{|L| } TP(l) } {\sum_{l=1}^{|L|}TP(l)+FN(l) },
$$
and $TP(l)$ denotes the number of true positives the model predicts for a given label $l$, $FP(l)$ denotes the number of false positives and $FN(l)$ the number of false negatives.

The macro-f1 score, when performing multi-label classification, is defined as the average micro-f1 score over the whole set of labels $L$:
\begin{equation}
  \label{eq:macrof1}
 macrof1 = \frac{1}{|L|} \sum_{l \in L}micro-f1(l),
\end{equation}
where $microf1(l)$ is the $micro-f1$-score for the given label $l$.

\subsection{Experimental Setup}

\begin{table}[!t]
  \centering
  \begin{tabular}{l  c c c}
  \hline
  \textbf{Approach} &\textbf{Optimiser} & \textbf{Learning Rate} & \textbf{Specific Parameters} \\
  \hline \hline
  SNDE & RMSProp & 0.01 & $\alpha$=500, $b$=10, epochs=500 \\
  Node2Vec-S & SGD & 0.1 & p=0.5, q=2, epochs=15 \\
  Node2Vec-H & SGD & 0.1 & p=1.0, q=0.5, epochs=15 \\
  DeepWalk & SGD & 0.1 & epochs=15 \\
  Poincar\'{e} Disk (PD) & SGD & 0.1 & p=0.5, q=2, epochs=15 \\
  \hline
  \end{tabular}
  \caption{Key Hyper-Parameter Settings}
  \label{tab:hyperparms}
\end{table}

\subsubsection{Implementation Details}

The approaches used for experimentation were reimplemented in Tensorflow \cite{abadi2016tensorflow}, as the author-provided versions were not all available using the same framework. We also attempted to ensure the same Tensorflow-based optimisations were used across all the approaches \cite{shi2016}. Neural Networks contain many hyper-parameters a user can control to improve the performance, both of the predictive accuracy and the runtime, of a given dataset. This process can be extremely time consuming and often requires users to perform a grid search over a range of possible hyper-parameter values to find a combination which performs best \cite{Goodfellow2016}. For setting the required hyper-parameters for the approaches, we took the default values provided by the authors in their respective papers \cite{Grover2016} \cite{Chamberlain2017} \cite{Wang2016a}  keeping them constant across all datasets. The key hyper-parameters used for each approach are detailed in Table \ref{tab:hyperparms}. We have open sourced our implementations of these approaches and made them available online\footnote{https://github.com/sbonner0/unsupervised-graph-embedding/}.

\subsubsection{Experimental Environment} 

\begin{table}[!h]
  \centering
  \begin{tabular}{c c c c c}
  \hline
  \textbf{Dataset} & $|V|$ & $|E|$ & \textbf{Domain} & \textbf{Source} \\
  \hline \hline

  fly-drosophila-medulla & 1,800 & 33,500 & Biological & \cite{rossi2015} \\
  cit-HepTh & 27,770 & 352,807 & Citation & \cite{snapnets} \\
  email-Eu-core & 1,005 & 25,571 & Communication & \cite{snapnets} \\
  inf-openflights & 2,900 & 30,500 & Infrastructure & \cite{rossi2015} \\
  soc-sign-bitcoinotc & 5,881 & 35,592 & Blockchain & \cite{snapnets} \\
  ego-Facebook & 4,039 & 88,234 & Social & \cite{snapnets} \\
  \hline
  \end{tabular}
  \caption{Empirical Graph Datasets}
  \label{tab:datasets}  
\end{table}

Experimentation was performed on a compute system with 2 NVIDIA Tesla K40c's, 2.3GHz Intel Xeon E5-2650 v3, 64GB RAM and the following software stack: Ubuntu Server 16.04 LTS, CUDA 9.0, CuDNN v7, TensorFlow 1.5, scikit-learn 0.19.0, Python 3.5 and NetworkX 2.0.

\subsubsection{Experimental Datasets}

The empirical datasets used for evaluation were taken from the Stanford Network Analysis Project (SNAP) data repository \cite{snapnets} and the Network Repository \cite{rossi2015} and are detailed in Table \ref{tab:datasets}. The domain label provided is taken from the listings of the graphs domain provided by SNAP \cite{snapnets} and Network Repository.

\subsection{Classification Algorithm Selection}
\label{sec:classalgsec}

\begin{table}[t!]
  \centering
  \resizebox{\textwidth}{!}{
  \begin{tabular}{lllllll}
  \hline
  \textbf{Feature}   & \textbf{Classifier} & \textbf{F1-Micro} & \textbf{F1-Macro} & \textbf{Uniform} & \textbf{Strat} & \textbf{Freq} \T\B \\
  \hline \hline
  \multirow{5}{*}{$DG$} & LR  &  $0.336(\pm0.015)$  & $0.190(\pm0.012)$ &  $+65.09\%$ & $+33.85\%$  &  $+12.07\%$ \T \\
                        & SVM(Lin)  &  $\mathbf{0.339(\pm0.017)}$  & $0.164(\pm0.013)$ & $\mathbf{+66.57\%}$  & $\mathbf{+35.03\%}$ & $\mathbf{+13.07\%}$  \\
                        & SVM(RBF)  &  $0.336(\pm0.021)$  & $0.158(\pm0.013)$ & $+65.09\%$  & $+33.84\%$ & $+12.07\%$  \\
                        & NN    &  $0.329(\pm0.013)$  & $\mathbf{0.200(\pm0.018)}$ & $+61.65\%$  & $+31.05\%$ & $+9.73\%$  \\
                        & NN-2   &  $0.326(\pm0.016)$  & $0.192(\pm0.019)$ & $+60.18\%$  & $+29.85\%$ & $+8.73\%$   \B \\
  \hline
  \multirow{5}{*}{$TC$} & LR    &  $0.340(\pm0.011)$  & $0.154(\pm0.014)$ & $+109.34\%$  & $+37.19\%$ & $+12.38\%$  \T \\
                        & SVM(Lin)  &  $\mathbf{0.344(\pm0.015)}$  & $0.139(\pm0.006)$ & $\mathbf{+111.8\%}$  & $\mathbf{+38.8\%}$ & $+\mathbf{13.7\%}$   \\
                        & SVM(RBF)   &  $0.335(\pm0.018)$  & $0.130(\pm0.010)$ & $+106.26\%$  & $+35.17\%$ & $+10.73\%$   \\
                        & NN        &  $0.331(\pm0.019)$  & $0.157(\pm0.013)$ & $+103.8\%$  & $+33.56\%$ & $+9.4\%$   \\
                        & NN-2     &  $0.326(\pm0.017)$  & $\mathbf{0.163(\pm0.015)}$ & $+100.72\%$  & $+31.54\%$ & $+7.75\%$ \B  \\
  \hline
  \multirow{5}{*}{$EC$}  & LR     &  $0.590(\pm0.013)$  & $0.474(\pm0.010)$ & $+195.66\%$  & $+144.16\%$ & $+92.18\%$ \T  \\
                         & SVM(Lin)  &  $0.591(\pm0.012)$  & $0.480(\pm0.011)$ & $+196.16\%$  & $+144.58\%$ & $+92.51\%$   \\
                         & SVM(RBF)   &  $0.552(\pm0.012)$  & $0.446(\pm0.011)$ & $+176.62\%$  & $+128.44\%$ & $+79.8\%$   \\
                         & NN      &  $0.629(\pm0.012)$  & $0.512(\pm0.017)$ & $+215.2\%$  & $+160.3\%$ & $+104.89\%$   \\
                         & NN-2    &  $\mathbf{0.630(\pm0.019)}$  & $\mathbf{0.513(\pm0.021)}$ & $\mathbf{+215.7\%}$  & $\mathbf{+160.72\%}$ & $\mathbf{+105.21\%}$  \B \\
  \hline
  \end{tabular}}
  \caption{Total Degree, Triangle Count and Eigenvector Centrality classification results for DeepWalk embeddings on the ego-Facebook dataset. Results for Micro and Macro-F1 scores are the mean after 5-fold cross validation, with standard deviations. Lift over Uniform, Stratified and Frequency predictors are presented as percentages.}
  \label{tab:class-dw}
\end{table}

\begin{table}[h!]
  \centering
  \resizebox{\textwidth}{!}{
  \begin{tabular}{lllllll}
  \hline
  \textbf{Feature}       & \textbf{Classifier} & \textbf{F1-Micro} & \textbf{F1-Macro} & \textbf{Uniform} & \textbf{Strat} & \textbf{Freq} \T\B \\
  \hline \hline
  \multirow{5}{*}{$DG$}  & LR    &  $0.284(\pm0.013)$  & $0.177(\pm0.008)$ & $+53.15\%$  & $+21.0\%$ & $-5.28\%$ \T   \\
                         & SVM(Lin)   &  $\mathbf{0.295(\pm0.020)}$  & $0.167(\pm0.012)$ & $\textbf{+59.08\%}$  & $+25.69\%$ & $\textbf{-1.61\%}$   \\
                         & SVM(RBF)   & $0.289(\pm0.017)$  & $0.142(\pm0.006)$ & $+55.85\%$  & $\textbf{+23.13\%}$ & $-3.61\%$   \\
                         & NN         & $0.253(\pm0.012)$  & $0.187(\pm0.012)$ & $+36.43\%$  & $+7.79\%$ & $-15.62\%$   \\
                         & NN-2       & $0.247(\pm0.018)$  & $\mathbf{0.193(\pm0.019)}$ & $+33.2\%$  & $+5.24\%$ & $-17.62\%$  \B \\
  \hline
  \multirow{5}{*}{$TC$} & LR         & $0.284(\pm0.015)$  & $0.138(\pm0.011)$ & $+99.15\%$  & $+18.87\%$ & $-6.13\%$ \T  \\
                        & SVM(Lin)   & $0.296(\pm0.016)$  & $0.125(\pm0.008)$ & $+107.56\%$  & $+23.89\%$ & $-2.16\%$   \\
                        & SVM(RBF)   & $\mathbf{0.300(\pm0.018)}$  & $0.124(\pm0.006)$ & $\mathbf{+110.37\%}$  & $\textbf{+25.57\%}$ & $\textbf{-0.84\%}$   \\
                        & NN         & $0.264(\pm0.020)$  & $0.161(\pm0.018)$ & $+85.12\%$  & $+10.5\%$ & $-12.74\%$   \\
                        & NN-2       & $0.247(\pm0.018)$  & $\mathbf{0.162(\pm0.016)}$ & $+73.2\%$  & $+3.38\%$ & $-18.36\%$ \B  \\
  \hline
  \multirow{5}{*}{$EC$} & LR         & $0.297(\pm0.008)$  & $0.166(\pm0.004)$ & $+70.4\%$  & $+12.85\%$ & $-3.26\%$ \T  \\
                        & SVM(Lin)   & $\mathbf{0.316(\pm0.010)}$  & $0.156(\pm0.006)$ & $\mathbf{+81.3\%}$  & $\mathbf{+20.07\%}$ & $+\mathbf{2.93\%}$   \\
                        & SVM(RBF)   & $0.309(\pm0.017)$  & $0.149(\pm0.008)$ & $+77.28\%$  & $+17.41\%$ & $+0.65\%$   \\
                        & NN         & $0.286(\pm0.013)$  & $0.198(\pm0.018)$ & $+64.08\%$  & $+8.67\%$ & $-6.84\%$   \\
                        & NN-2       & $0.272(\pm0.018)$  & $\mathbf{0.201(\pm0.014)}$ & $+56.05\%$  & $+3.35\%$ & $-11.4\%$  \B \\
  \hline
  \end{tabular}}
  \caption{Degree, Triangle Count and Eigenvector Centrality classification results for SDNE embeddings on the ego-Facebook dataset. Results for Micro and Macro-F1 scores are the mean after 5-fold cross validation, with standard deviations. Lift over Uniform, Stratified and Frequency predictors are presented as percentages.}
  \label{tab:class-ae}
\end{table}

As highlighted throughout the paper, we are focusing our research on unsupervised graph embedding approaches. In order to be able to use the embeddings for further analysis, they must be classified using a supervised classification model. Traditionally in the embedding literature, a simple Logistic Regression is used in any classification task \cite{Perozzi2014} \cite{Mikolov2013}, with seemingly little work exploring the use of more sophisticated models to perform the classification. 

\begin{figure*}[!t]
  \centering
\subfloat[Macro Drosophila]{%
    \includegraphics[width=0.25\linewidth]{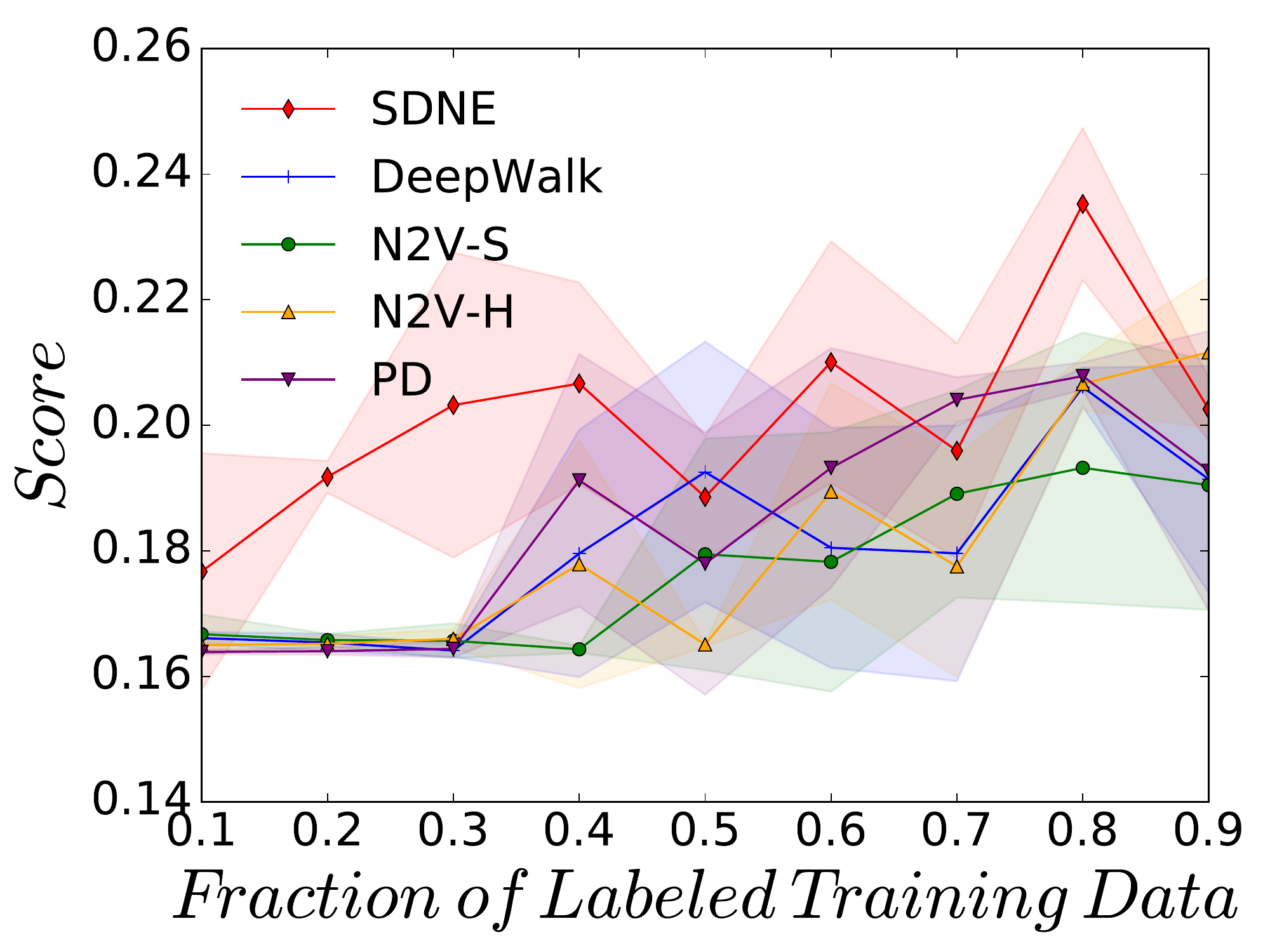}}
  \label{DCmiCA}\hfill
\subfloat[Micro Drosophila]{%
    \includegraphics[width=0.25\linewidth]{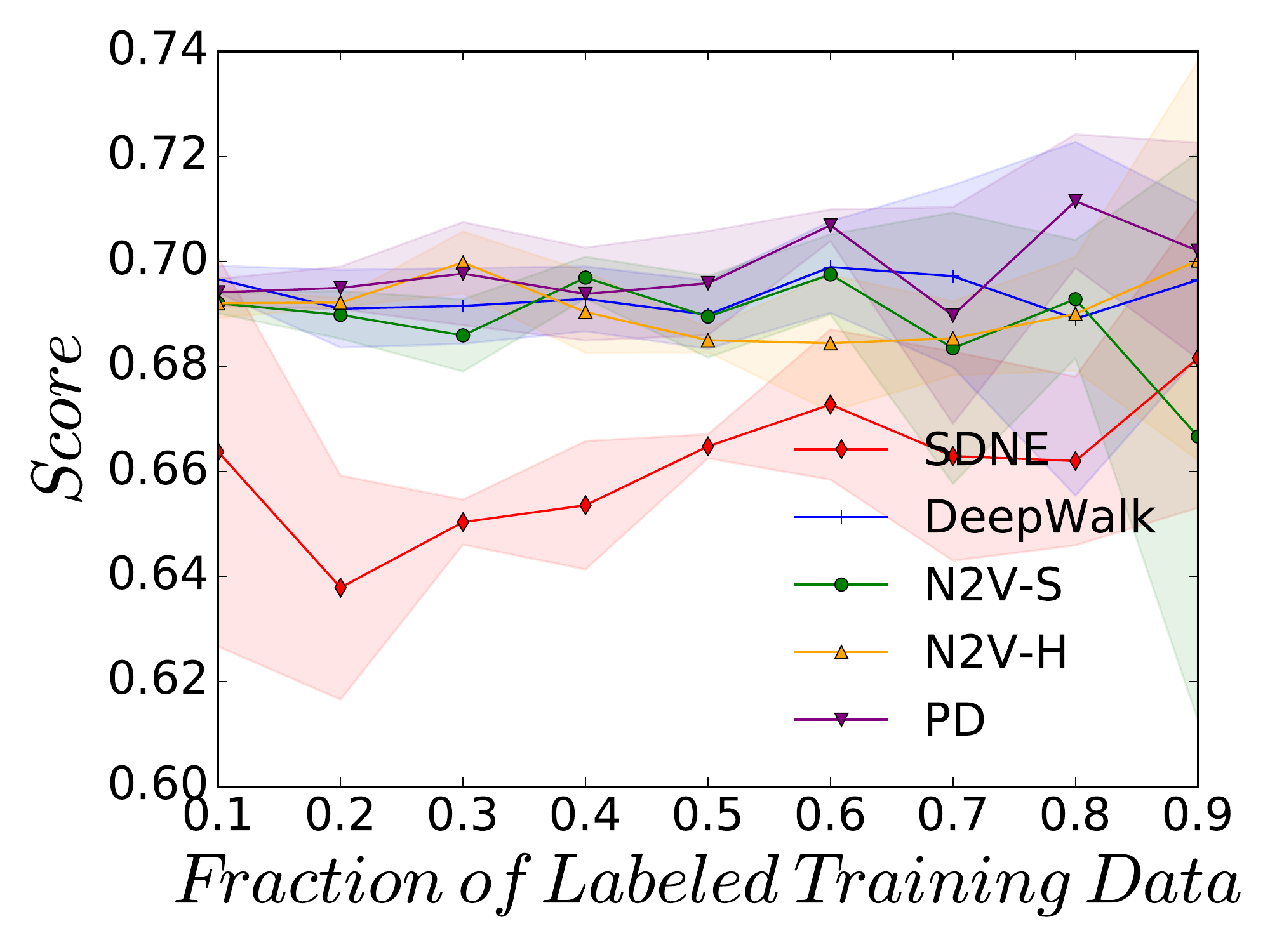}}
  \label{DCmiFB}\hfill
\subfloat[Macro HepTh]{%
    \includegraphics[width=0.25\linewidth]{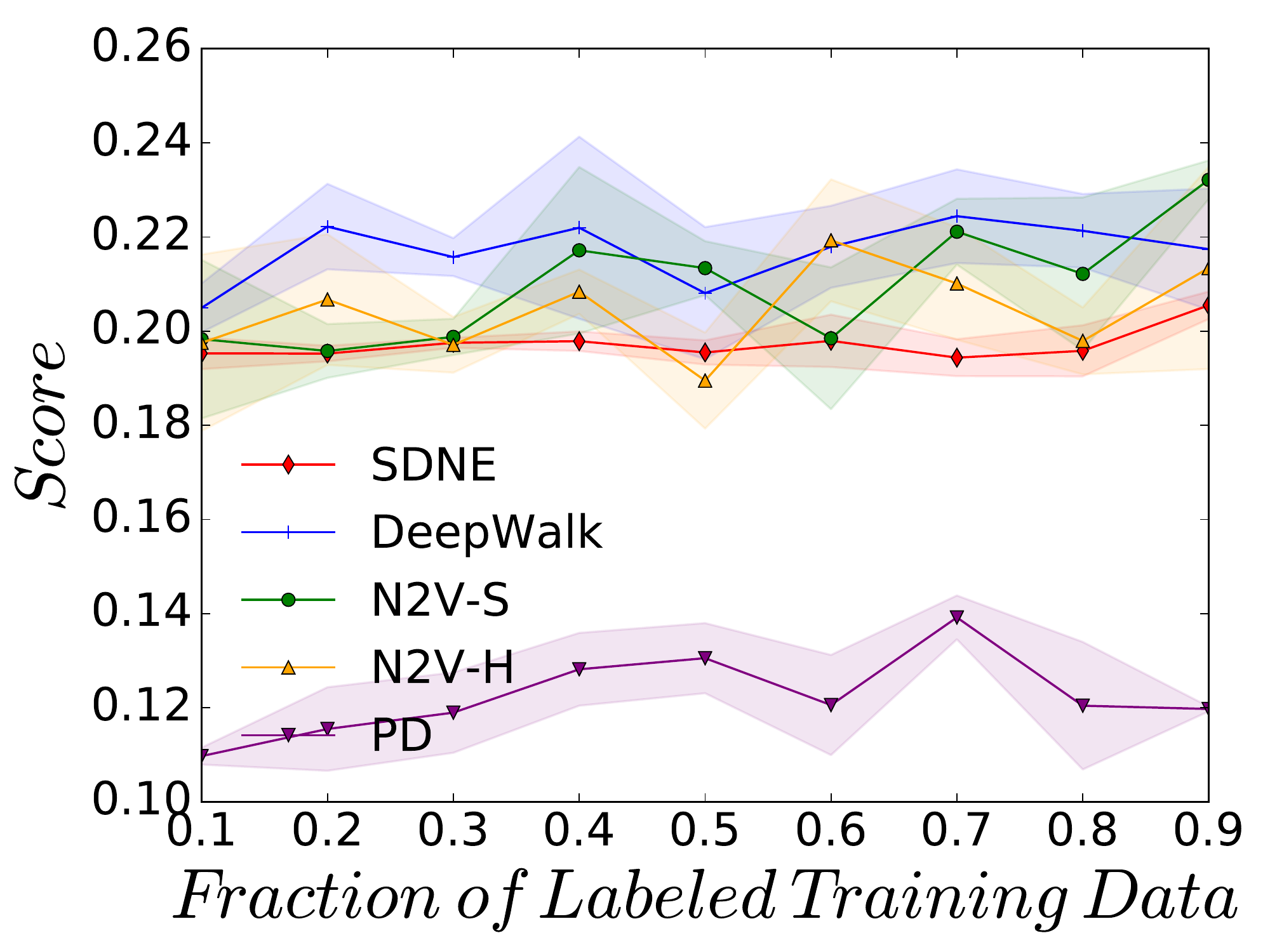}}
  \label{DCmiGN}\hfill
\subfloat[Micro HepTh]{%
    \includegraphics[width=0.25\linewidth]{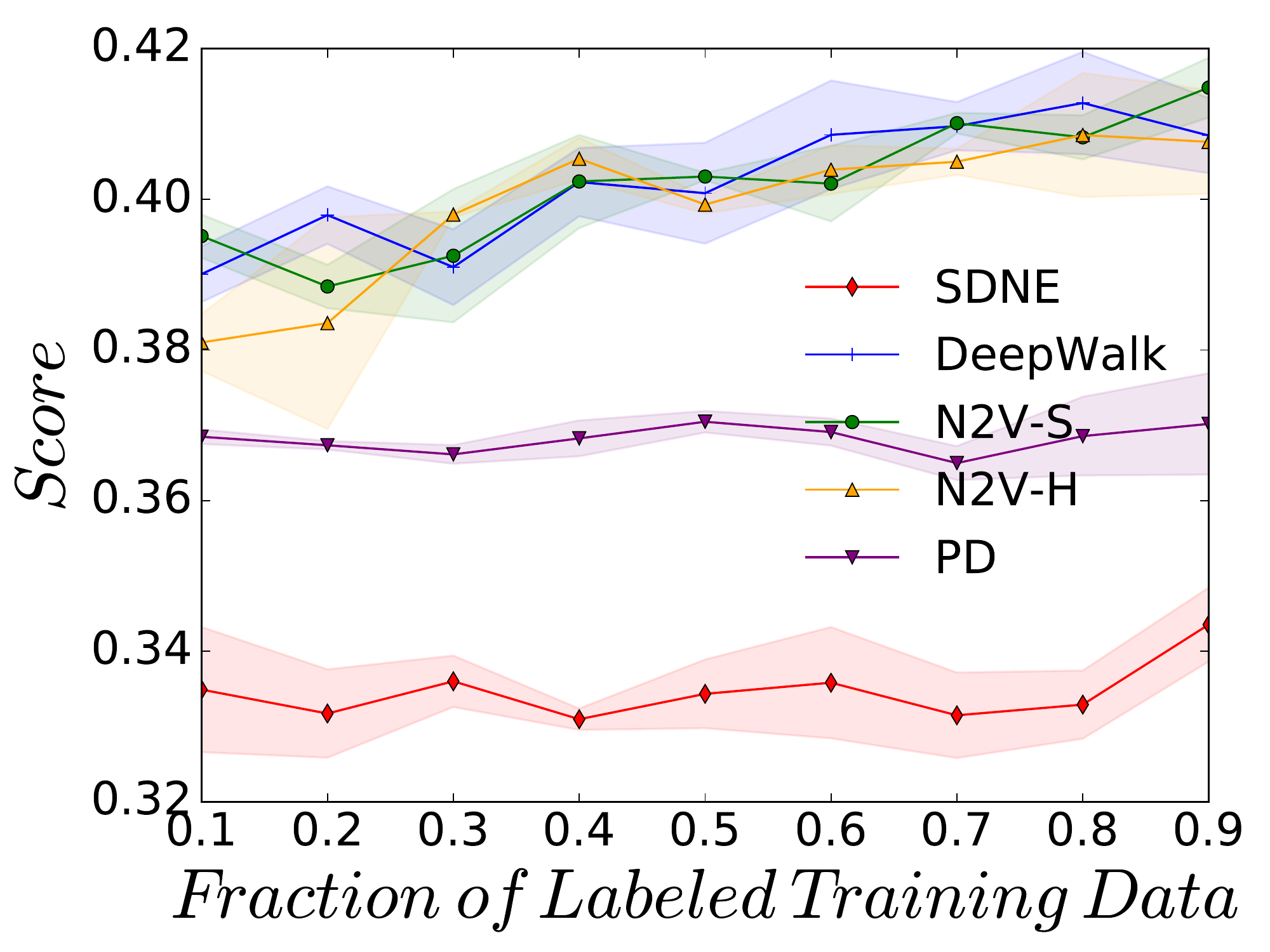}}
  \label{DCmiWI}\\

\subfloat[Macro Email-EU]{%
    \includegraphics[width=0.25\linewidth]{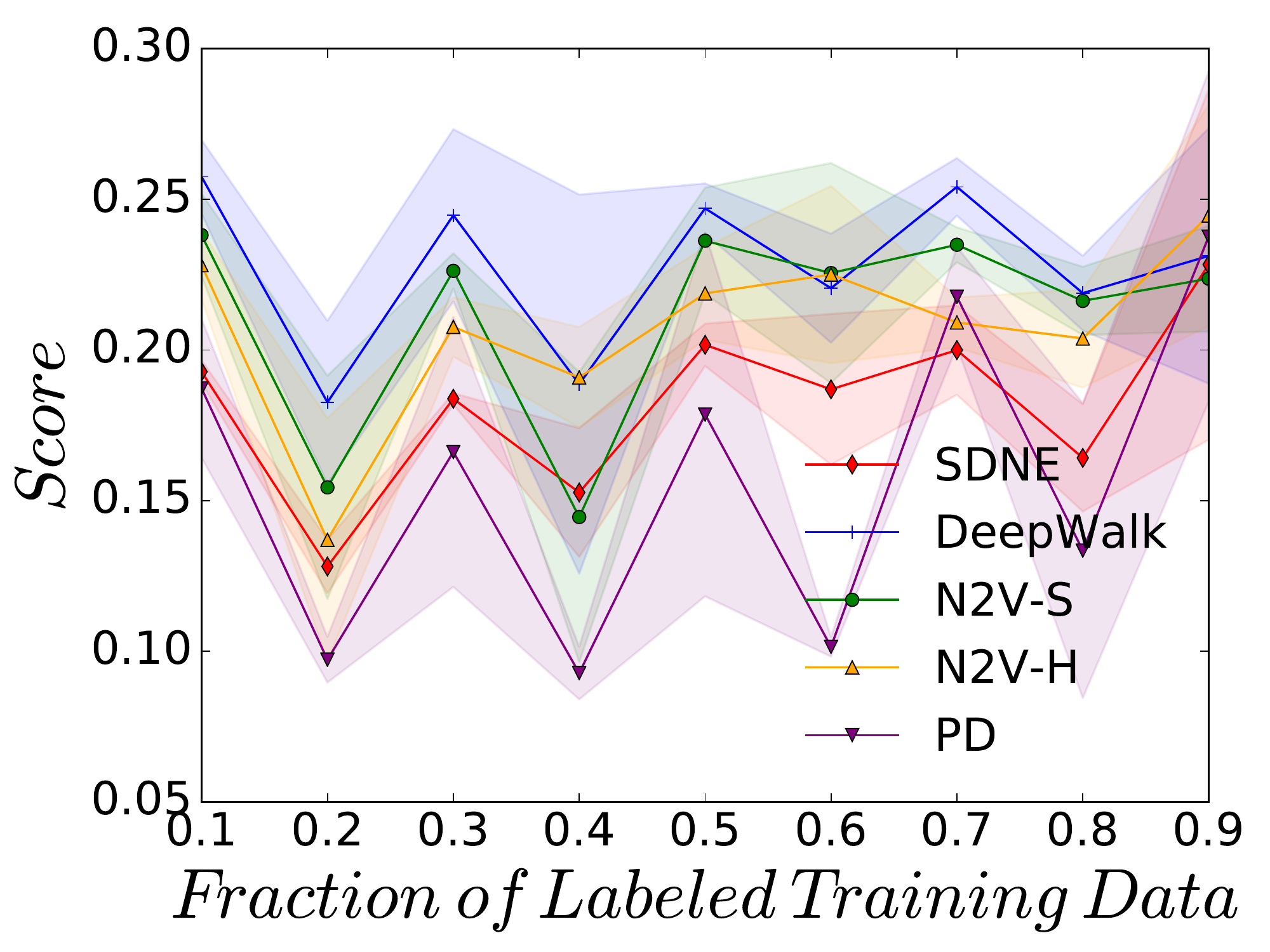}}
  \label{DCmaCA}\hfill
\subfloat[Micro Email-EU]{%
    \includegraphics[width=0.25\linewidth]{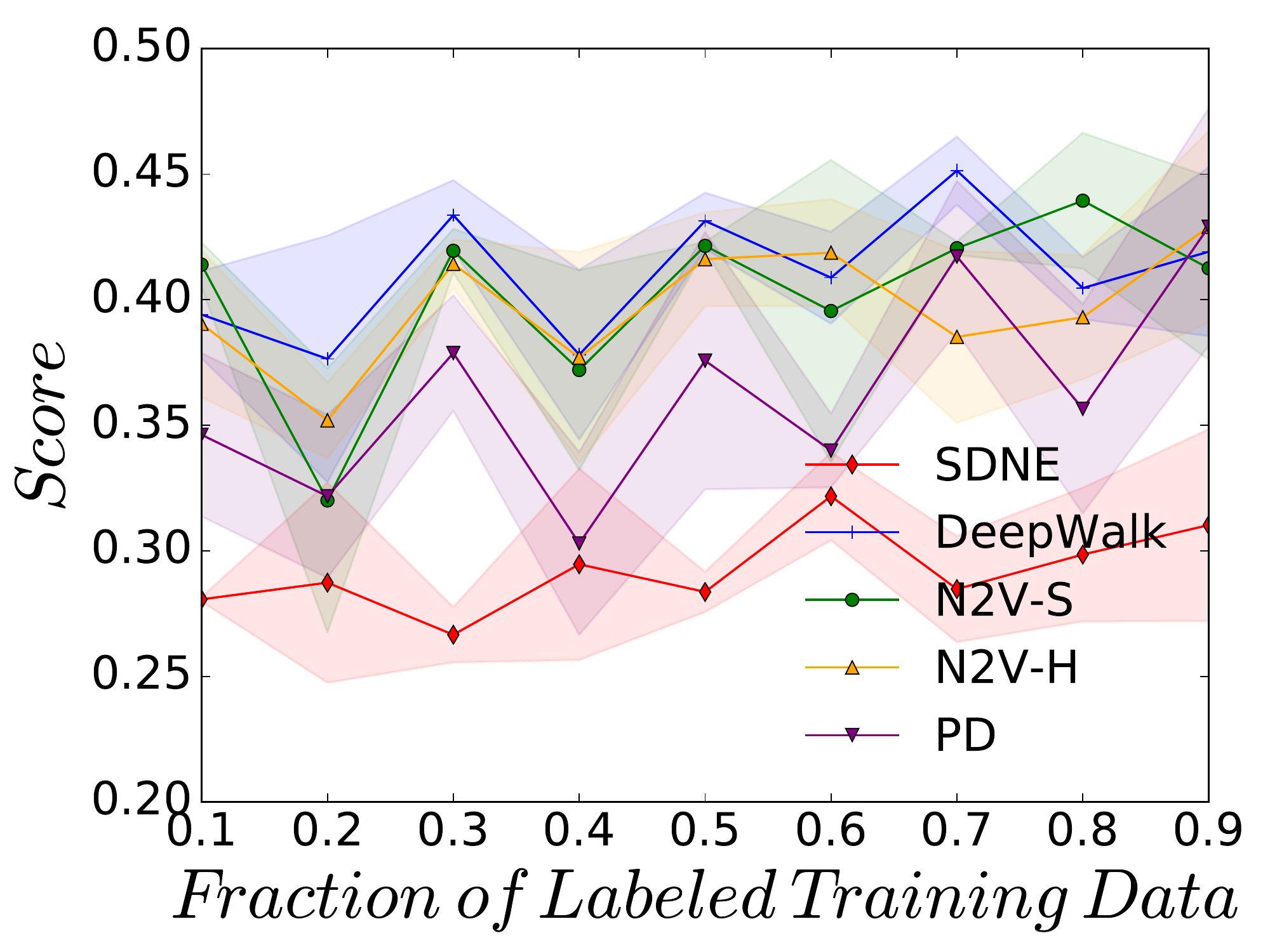}}
  \label{DCmaFB}\hfill
\subfloat[Macro Facebook]{%
    \includegraphics[width=0.25\linewidth]{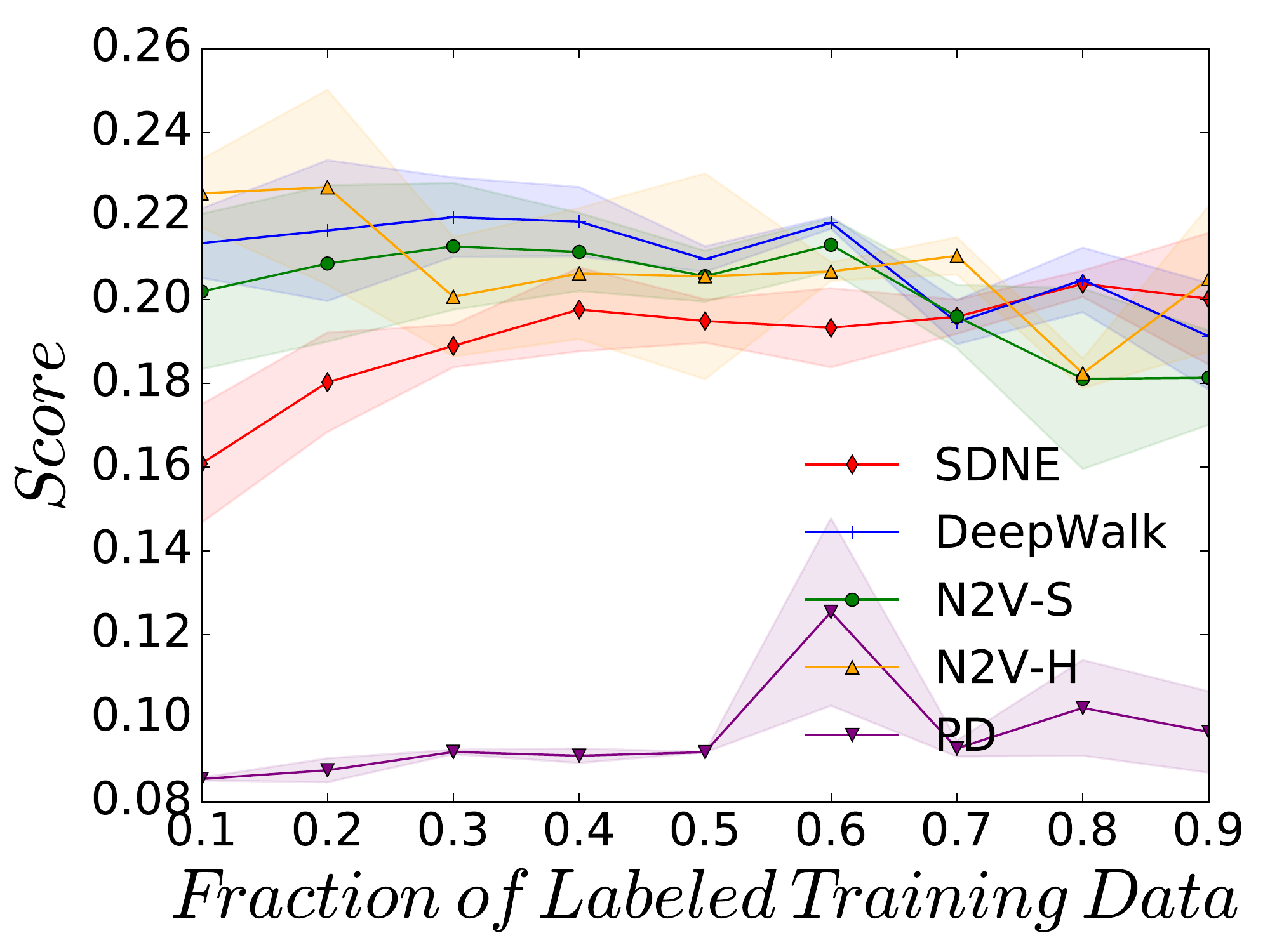}}
  \label{DCmaGN}\hfill
\subfloat[Micro Facebook]{%
    \includegraphics[width=0.25\linewidth]{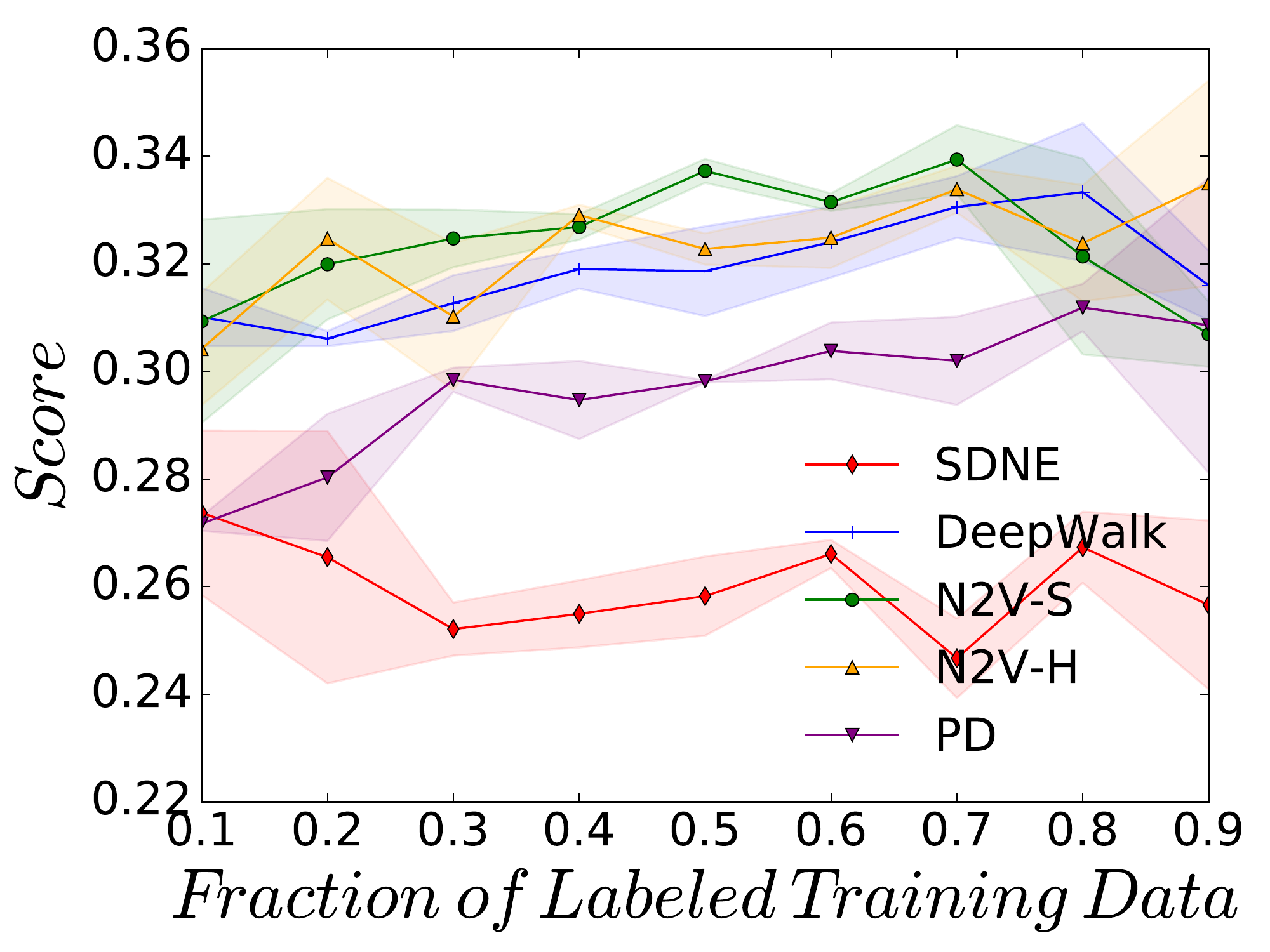}}
  \label{DCmaWI} \\ 

  \subfloat[Macro Openflights]{%
    \includegraphics[width=0.25\linewidth]{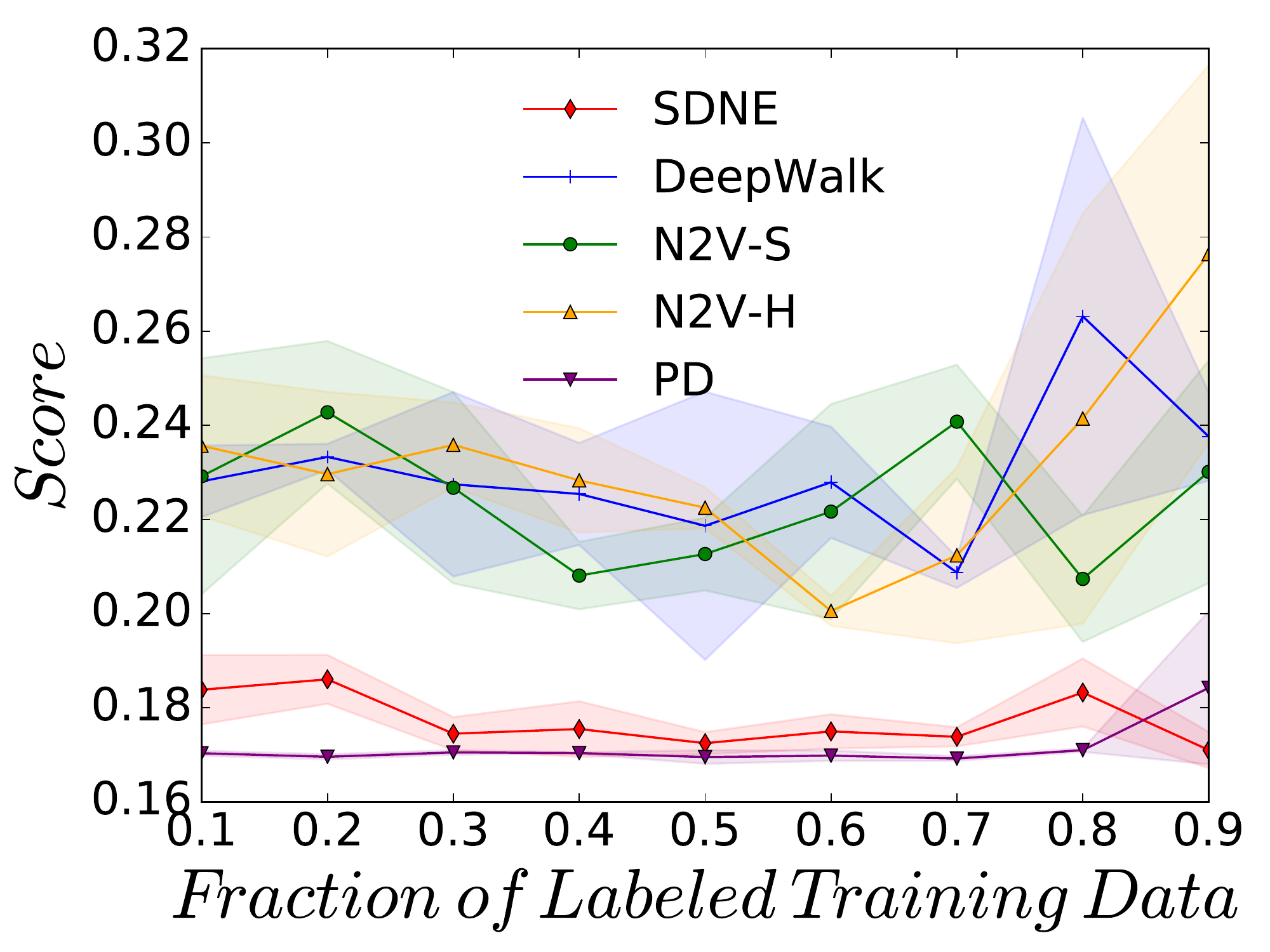}}
  \label{DCmaCA}\hfill
\subfloat[Micro Openflights]{%
    \includegraphics[width=0.25\linewidth]{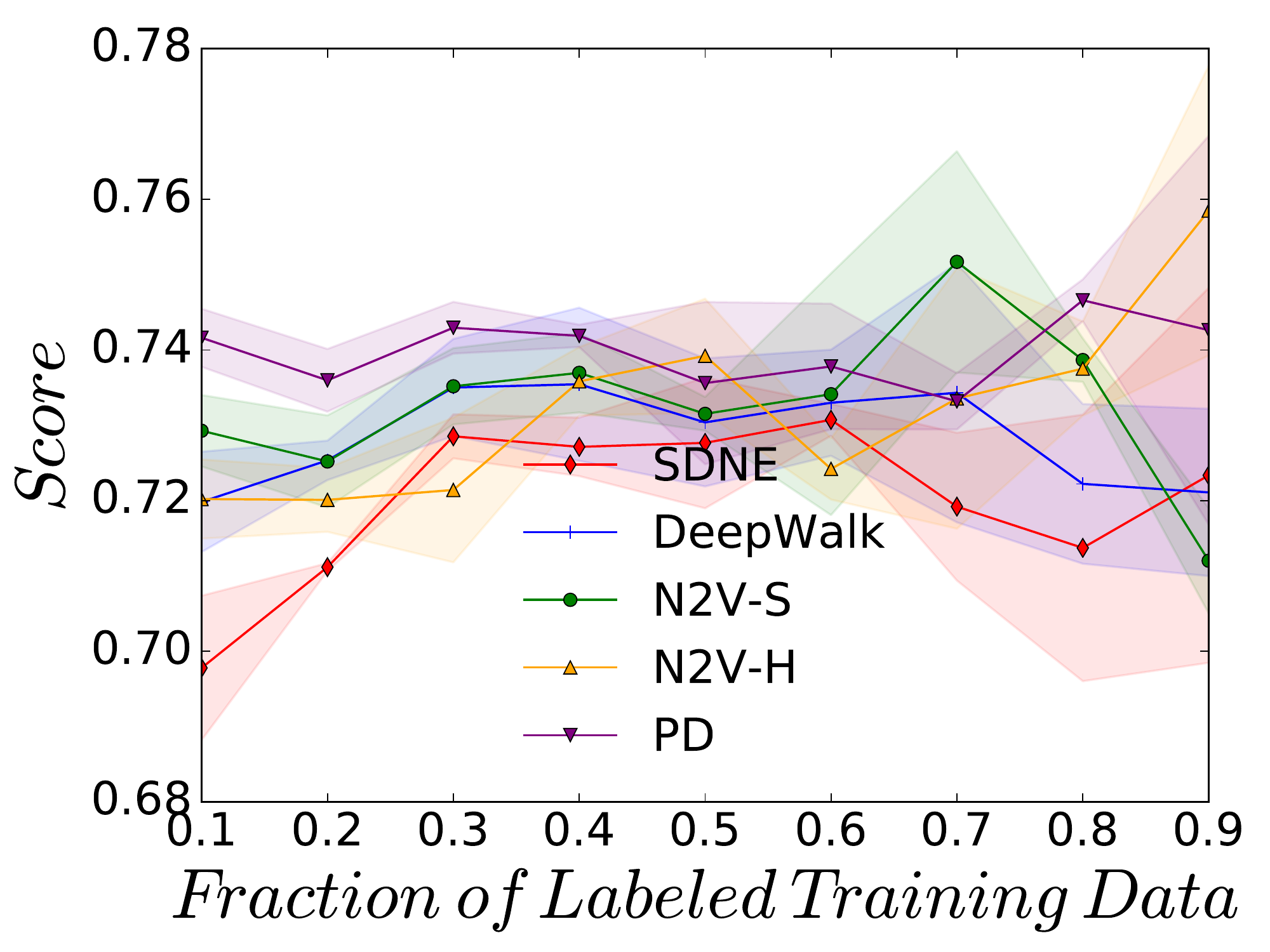}}
  \label{DCmaFB}\hfill
\subfloat[Macro Bitcoinotc]{%
    \includegraphics[width=0.25\linewidth]{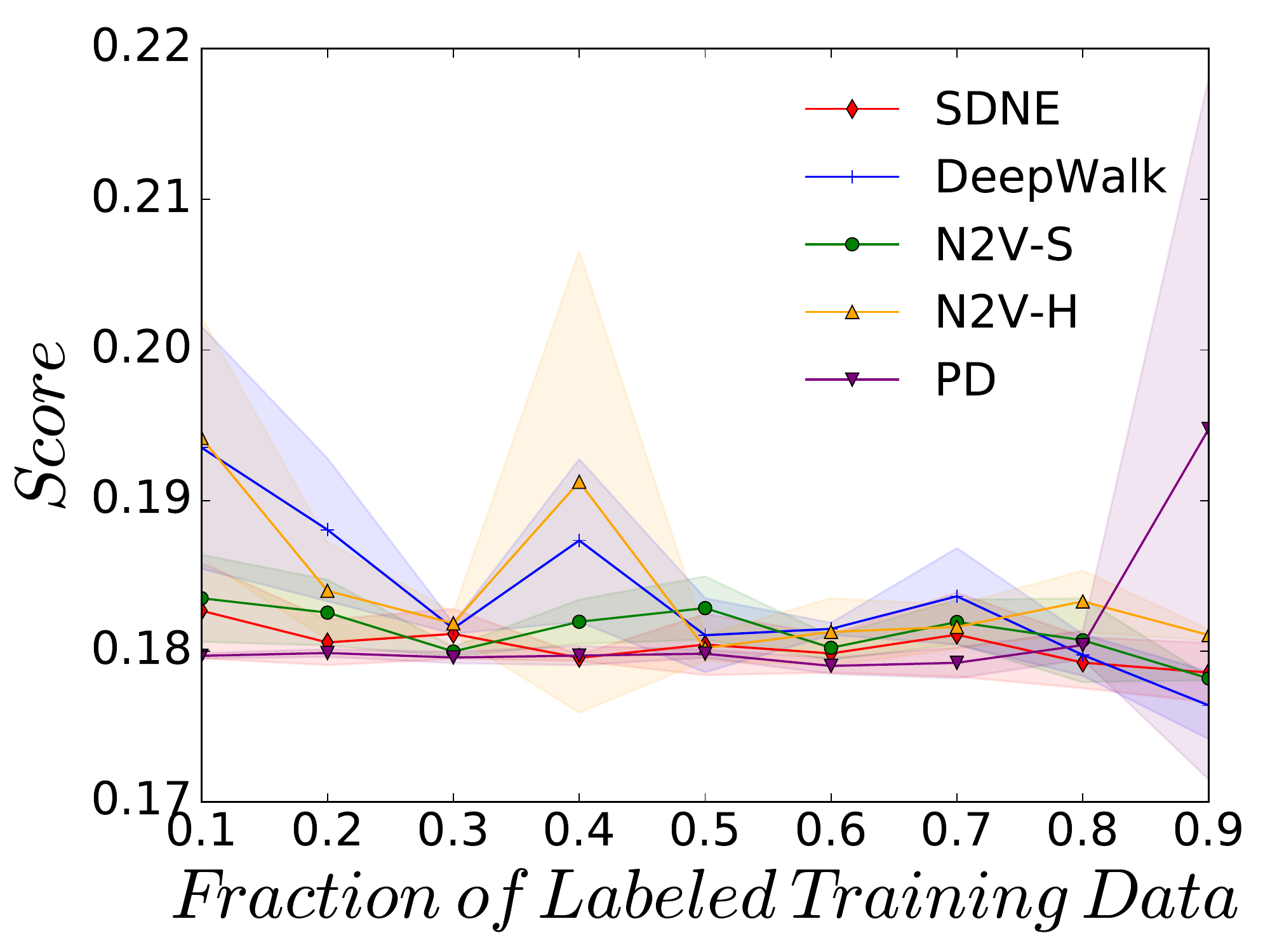}}
  \label{DCmaGN}\hfill
\subfloat[Micro Bitcoinotc]{%
    \includegraphics[width=0.25\linewidth]{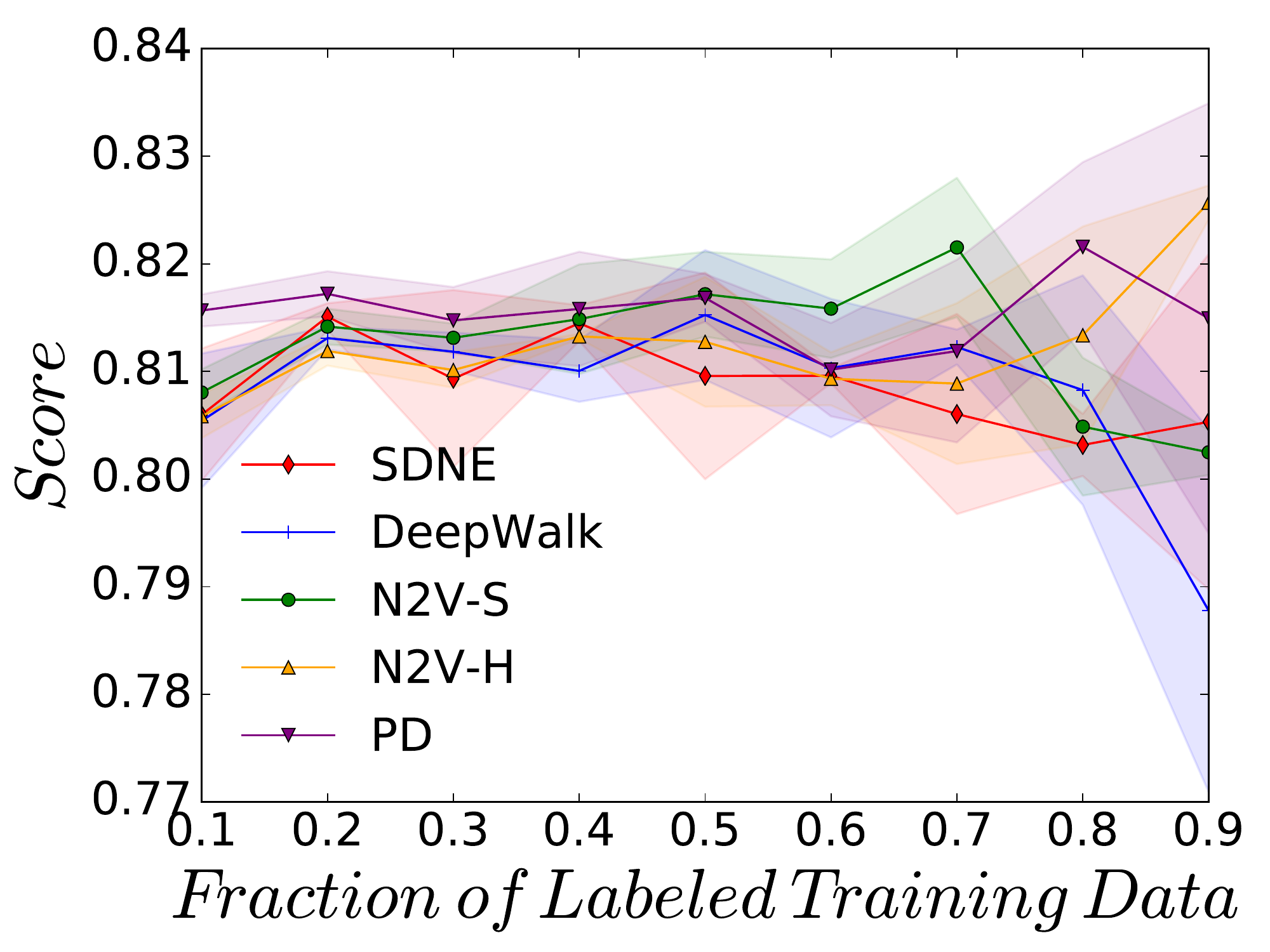}}
  \label{DCmaWI}
\caption{Micro and Macro F1 Scores, across a range of labelling fractions, for all approaches when predicting a vertex's degree (DG) value across all datasets.}
\label{fig:DG_FIG}
\end{figure*}

In this section we explore the effectiveness of five different models at performing the classification of the different embedding approaches - Logistic Regression (LR), Support Vector Machine (SVM) (Linear Kernel), SVM (RBF Kernel), a single hidden layer Neural Network and finally a second more complex Neural Network with two hidden layers and a larger number of hidden units. All the classifiers utilised in this section were taken from the Scikit-Learn Python package \cite{scikit-learn}. Additionally, given that our datasets do not have a equal distribution among the classes, we also explore the effectiveness of weighting the loss function used by the model inversely proportional to the frequency of the class \cite{karakoulas1999optimizing}. This use of a weighted loss function, although common in other areas of machine learning, has not been explored in regards to graph embeddings. 

For the results in this section, we present the mean Macro and Micro F1 scores, introduced in Section \ref{sec:precision}, after 5-fold cross validation. To assess the performance of the classifiers against the imbalance present in the datasets, we also display the percentage lift in mean test set accuracy over three rule-based prediction methods to act as baselines. These methods are Uniform Prediction (Where the classification of each item in the test is chosen uniformly at random from the possible classes), Stratified Prediction (where the classification follows the distribution of classes in the training set) and Frequent Class Prediction (Where the classification is determined by the most frequency class in the training set). A positive lift across all metrics strongly suggests that a mapping from the embedding space to the topological features is being learned, as the classification algorithm is over coming the biased distributions of classes in the dataset. 

We performed this experiment for all combination of datasets, embedding approaches and features, but due to the large quantity of results, we present only a subset here. Specifically we present the results for ego-Facebook dataset, using embeddings generated by DeepWalk and SDNE and classifying Degree, Triangle Count and Eigenvector Centrality. It should be noted that the patterns displayed here are representative of ones seen across all datasets.

Table \ref{tab:class-dw} highlights the performance of the potential classifiers, when using the DeepWalk embeddings taken from the ego-Facebook dataset. Results show that the choice of supervised classifier can have a large impact on the overall classification score. It can also be seen that the traditional choice of logistic regression does not produce the best results. Indeed the neural network and SVM classifier often gave the best scores but no classifier is best overall, suggesting that one needs to be chosen carefully for a given task.

Table \ref{tab:class-ae} highlights the results for the potential classifiers, when using the SDNE embeddings taken from the ego-Facebook dataset. Again, the variation in classification score across the set of tested classification metrics is quite substantial, with the linear SVM and neural network approaches having perhaps a small margin of improvement over the others. It is interesting to note that the logistic regression frequently used in the literature, never has the highest score in any metric. It can also be seen that, when compared with the DeepWalk results in Table \ref{tab:class-dw}, SDNE does worse at predicting all topological features which, although not the explicit purpose of this section, is interesting to note.

Using the results from this section, particularly the generally higher f1-macro scores mean a better results across all classes, all the classification results in Section \ref{sec:R} are presented using a single hidden layer neural network.

\section{Results}
\label{sec:R}

This section presents both the supervised and unsupervised results for predicting topological features from graph embeddings. 

\subsection{Topological Feature Prediction}

\begin{figure*}
  \centering
\subfloat[Macro Drosophila]{%
    \includegraphics[width=0.25\linewidth]{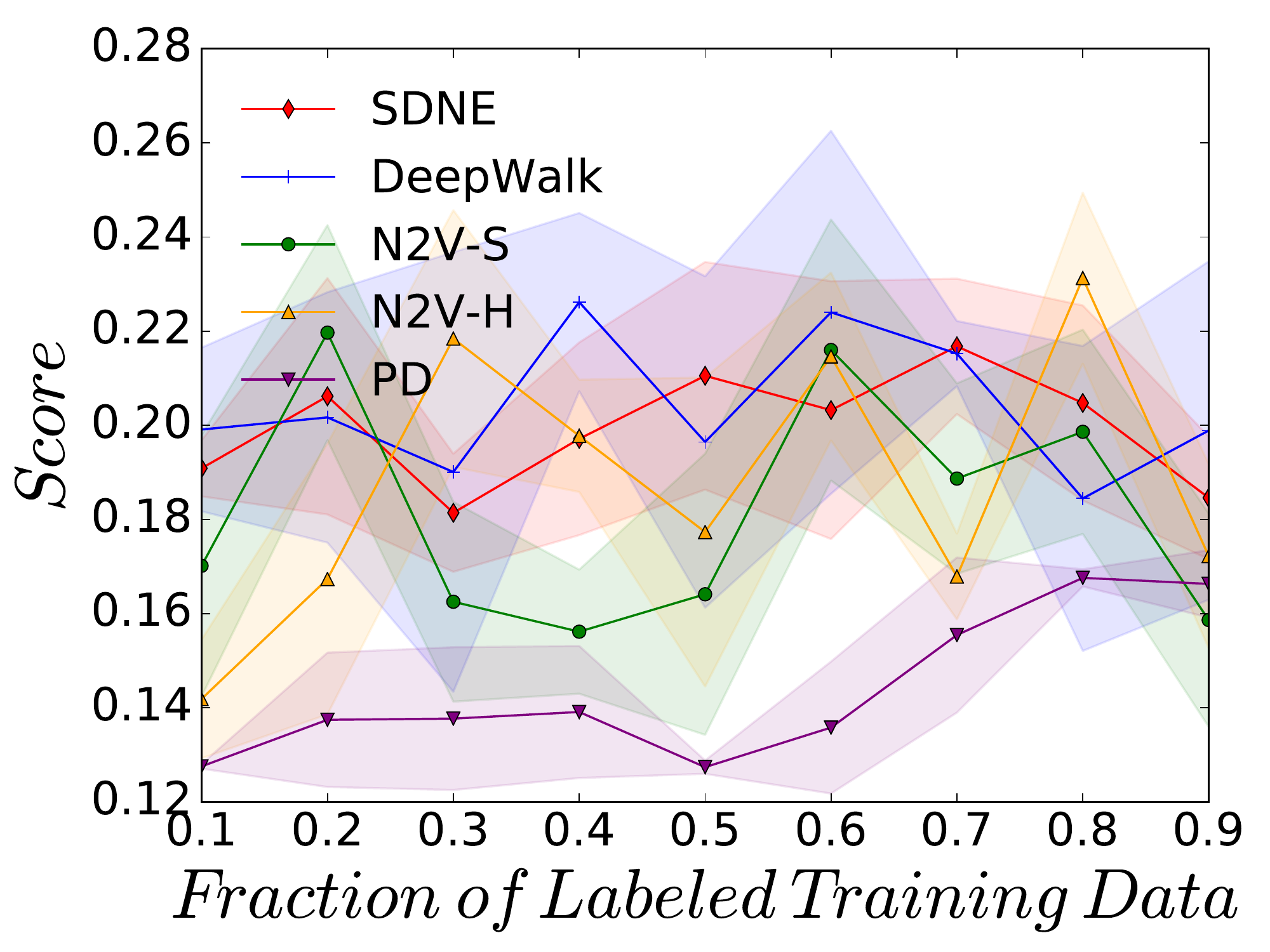}}
  \label{DCmiCA}\hfill
\subfloat[Micro Drosophila]{%
    \includegraphics[width=0.25\linewidth]{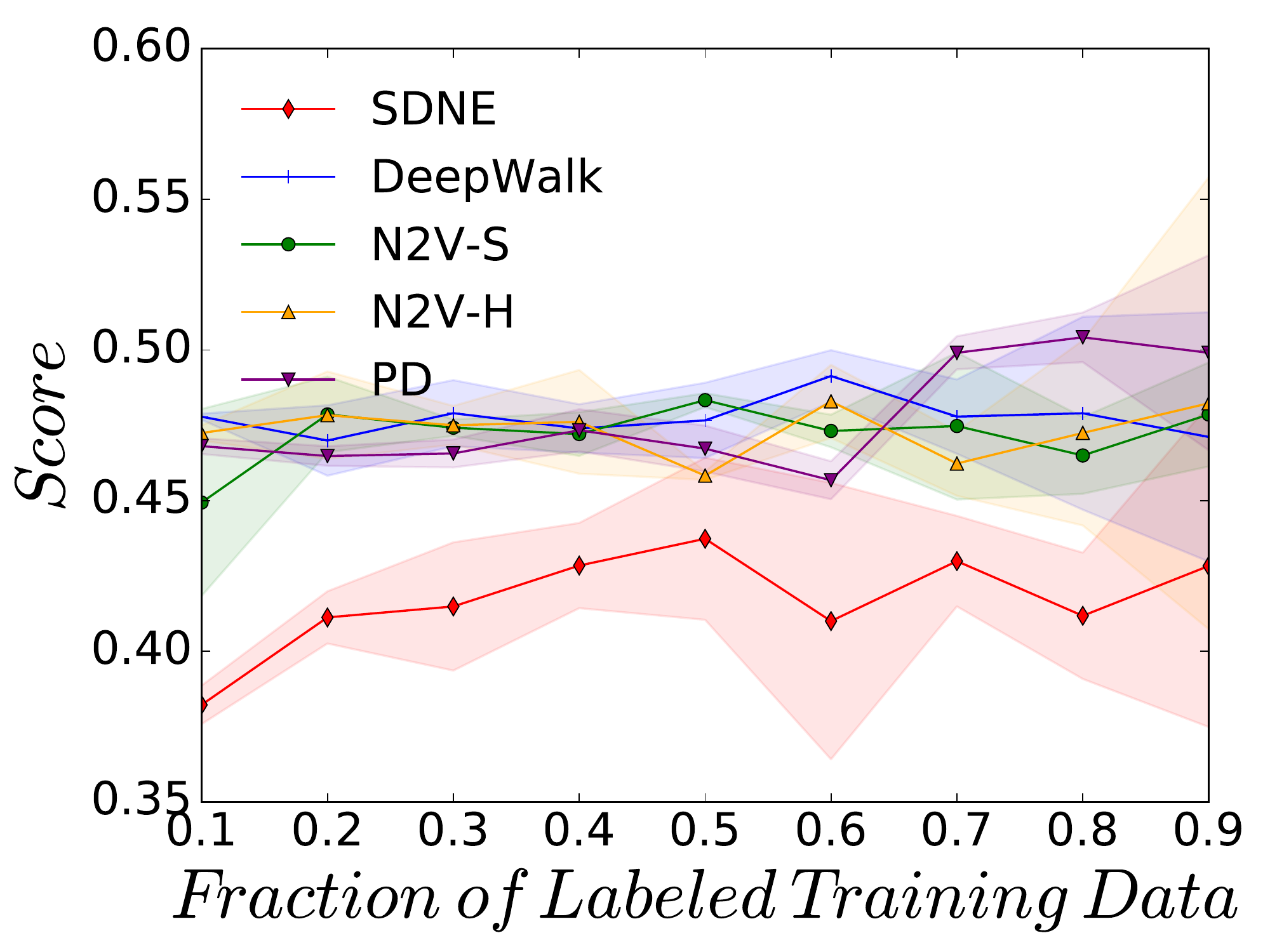}}
  \label{DCmiFB}\hfill
\subfloat[Macro HepTh]{%
    \includegraphics[width=0.25\linewidth]{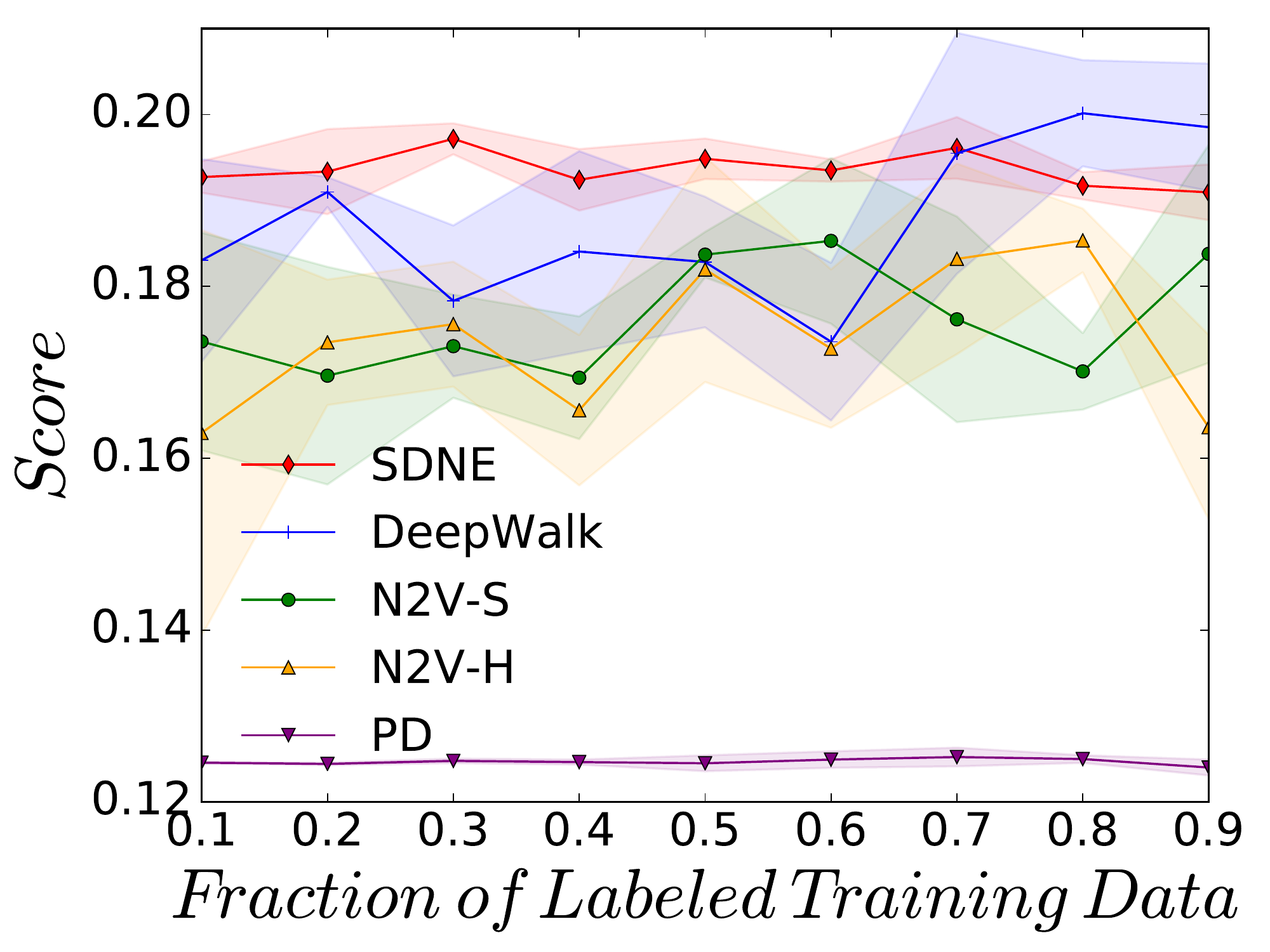}}
  \label{DCmiGN}\hfill
\subfloat[Micro HepTh]{%
    \includegraphics[width=0.25\linewidth]{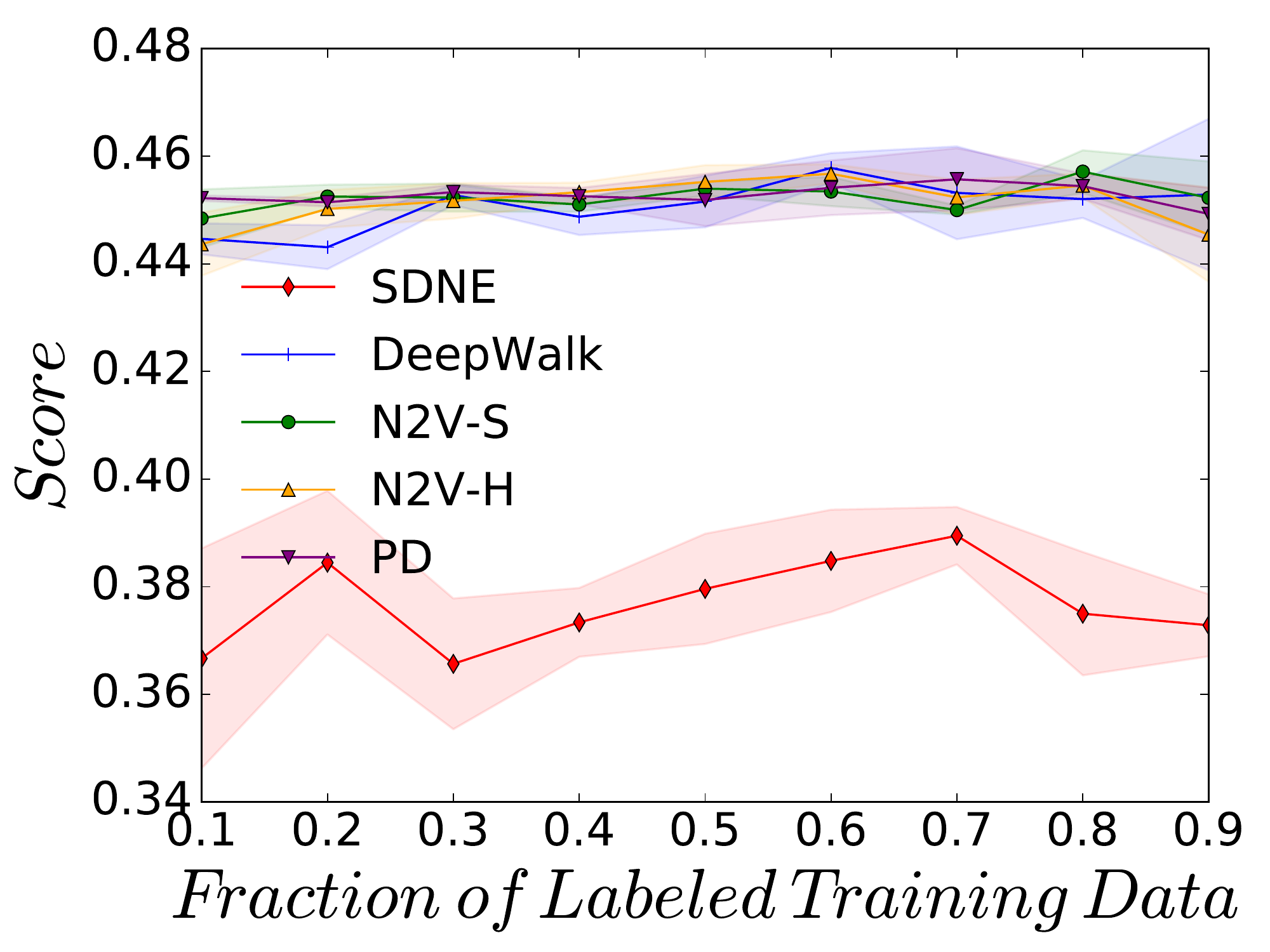}}
  \label{DCmiWI}\\

\subfloat[Macro Email-EU]{%
    \includegraphics[width=0.25\linewidth]{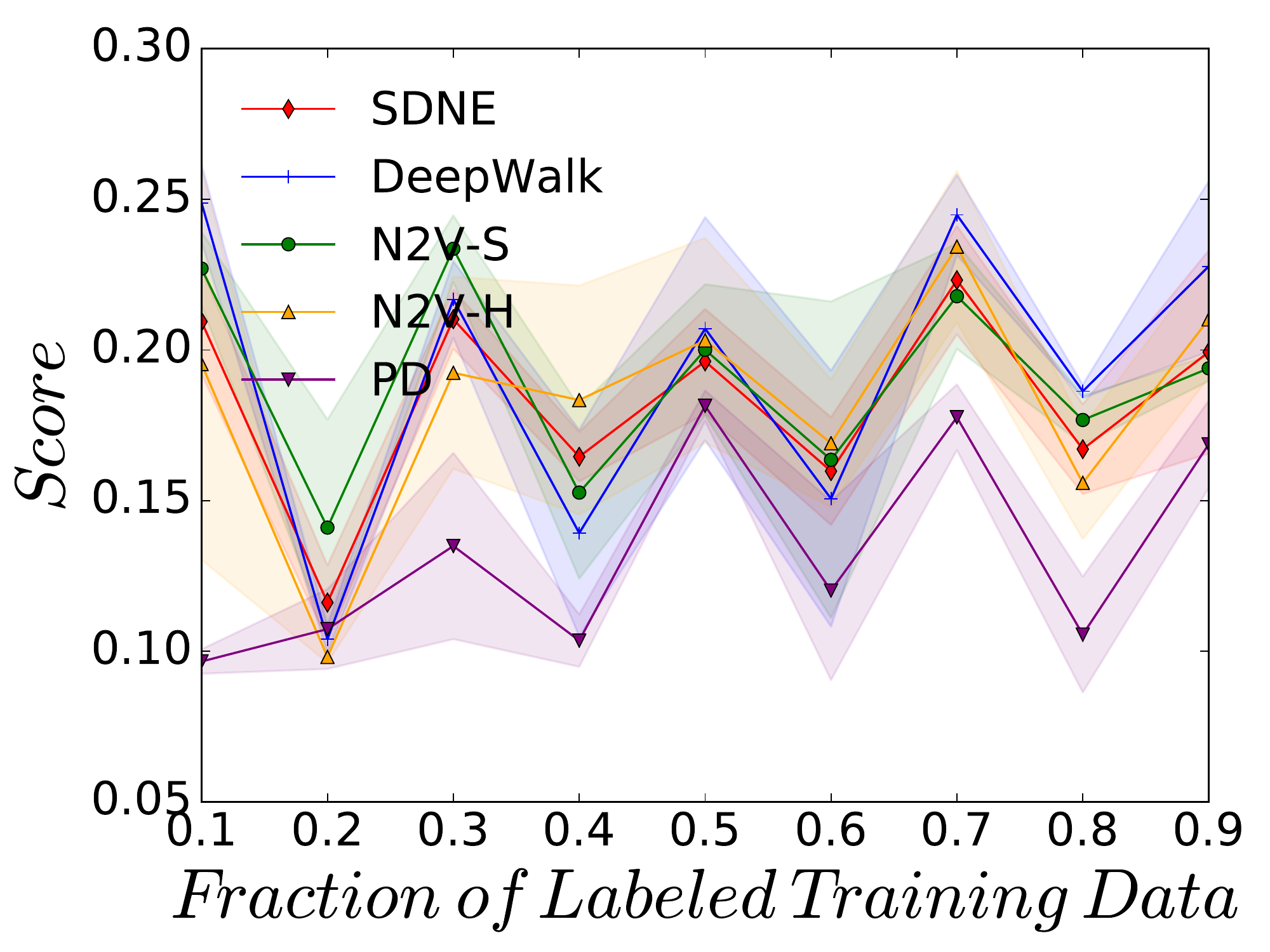}}
  \label{DCmaCA}\hfill
\subfloat[Micro Email-EU]{%
    \includegraphics[width=0.25\linewidth]{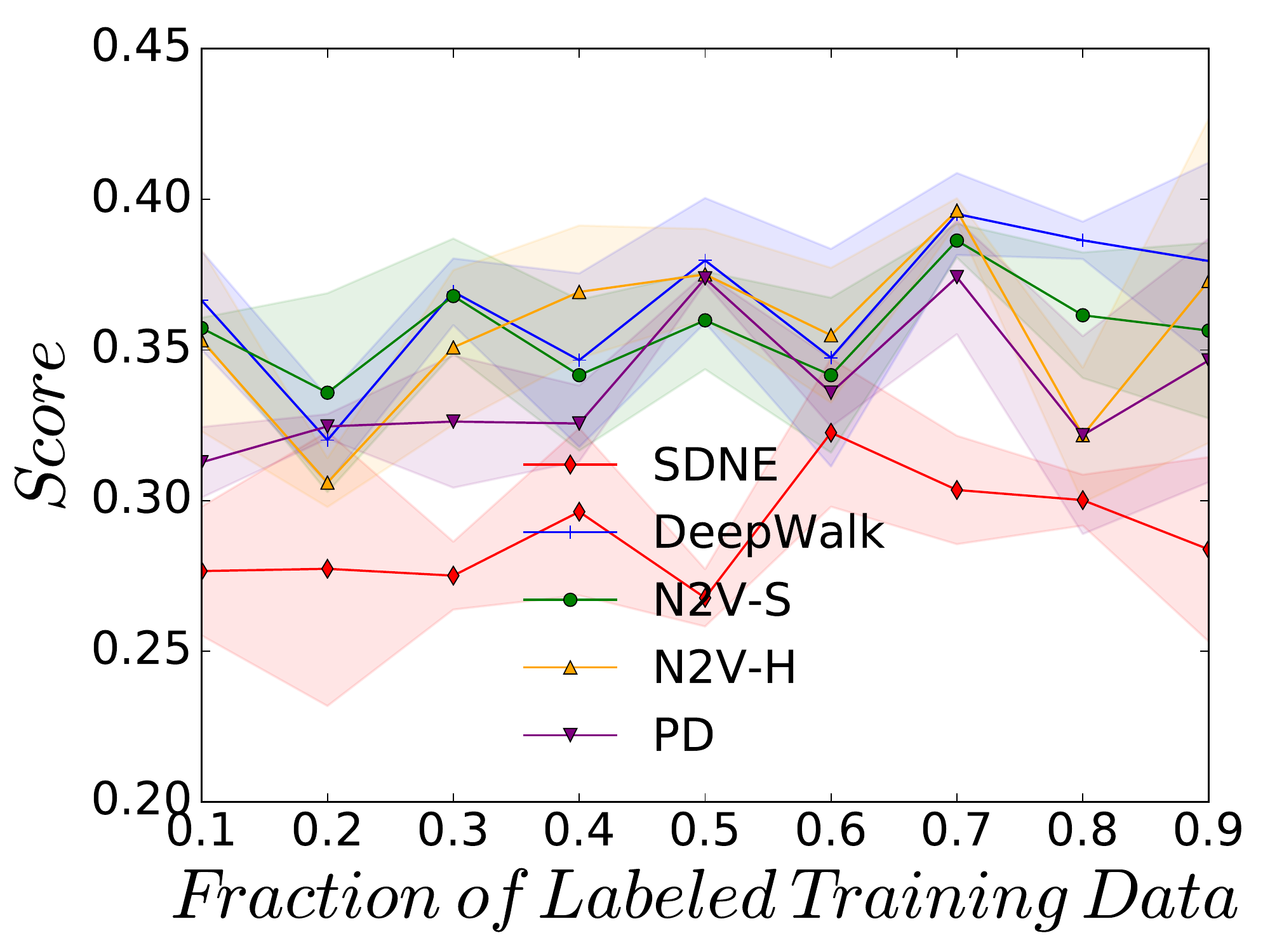}}
  \label{DCmaFB}\hfill
\subfloat[Macro Facebook]{%
    \includegraphics[width=0.25\linewidth]{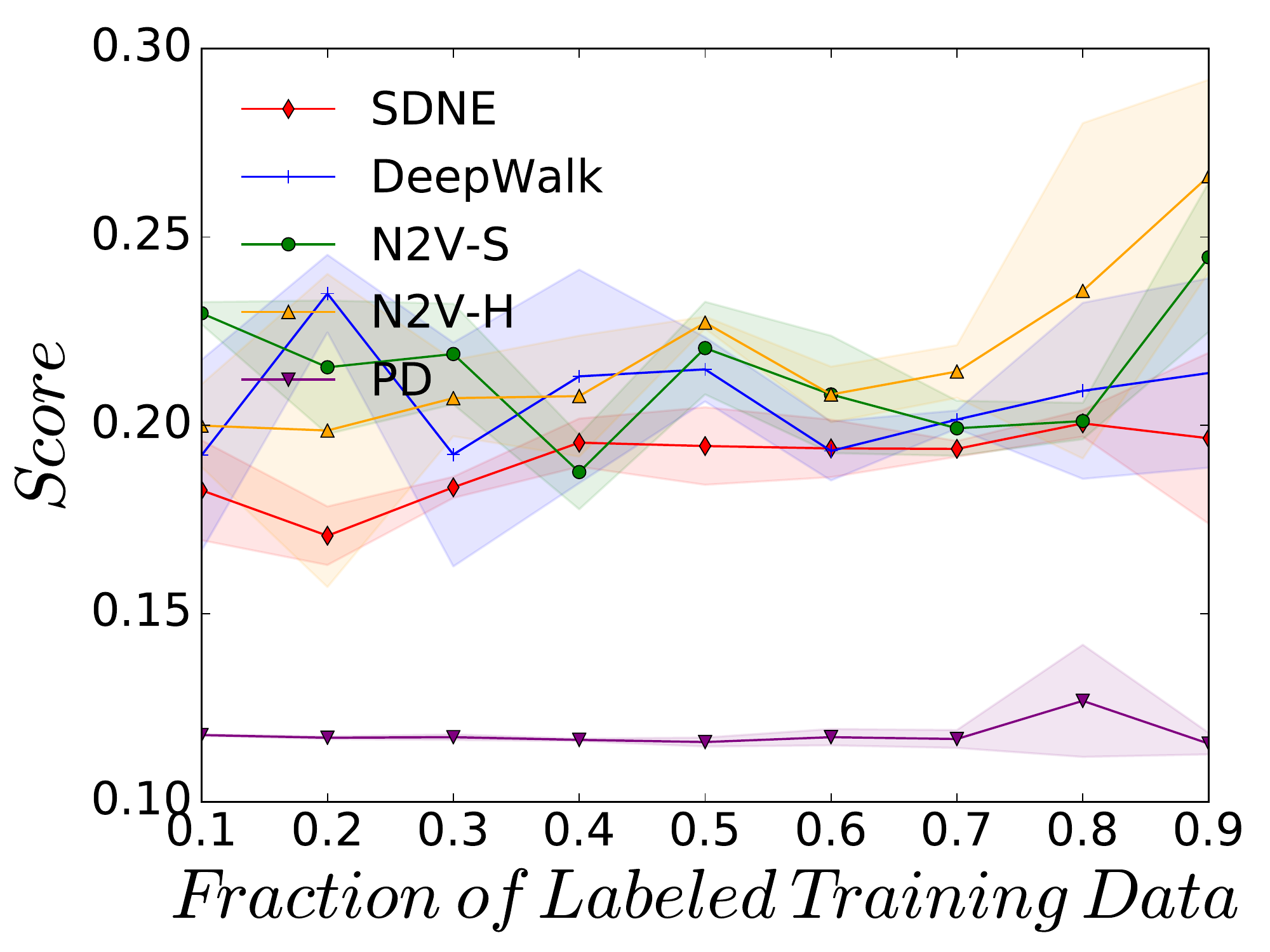}}
  \label{DCmaGN}\hfill
\subfloat[Micro Facebook]{%
    \includegraphics[width=0.25\linewidth]{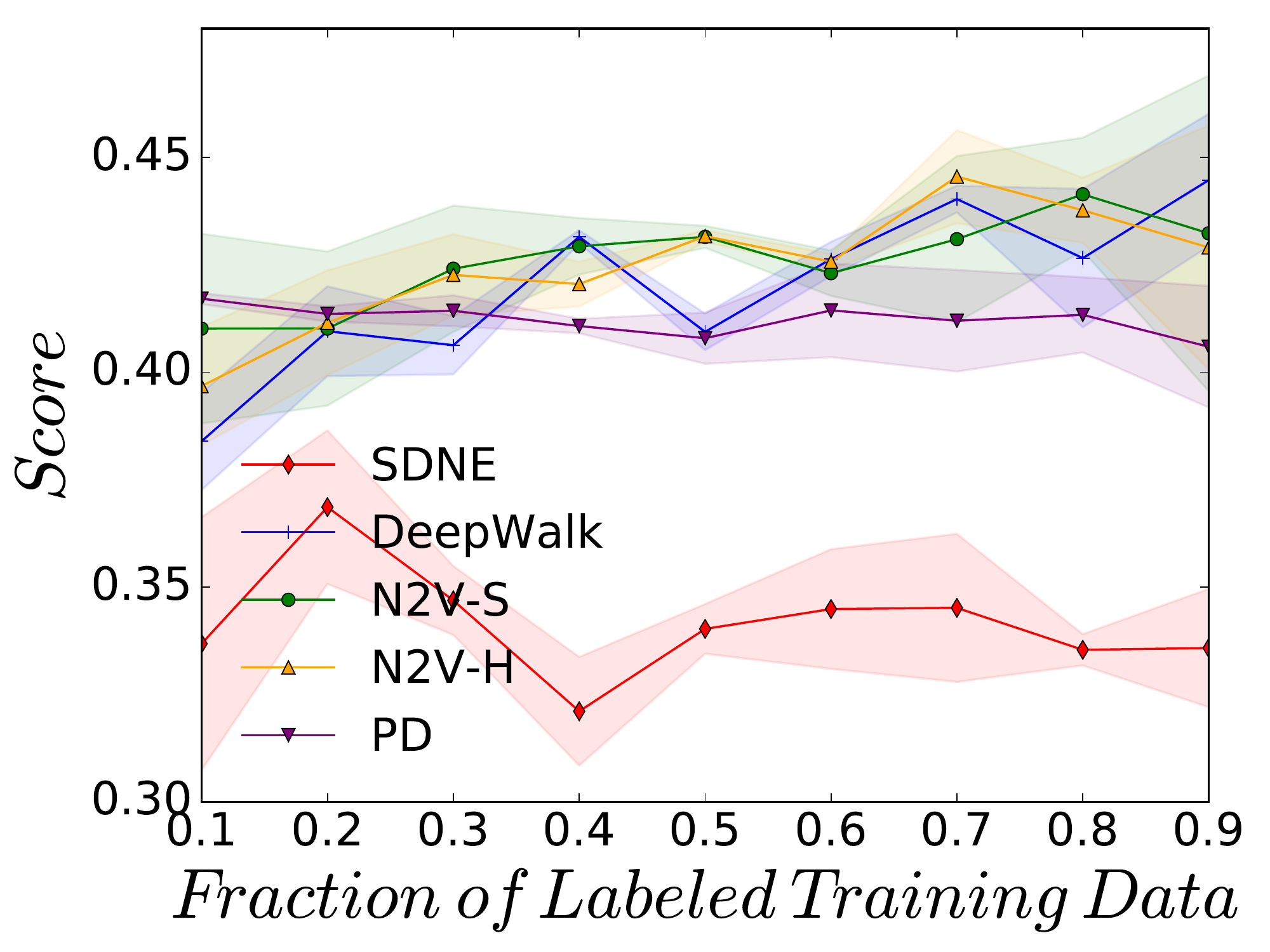}}
  \label{DCmaWI} \\ 

  \subfloat[Macro Openflights]{%
    \includegraphics[width=0.25\linewidth]{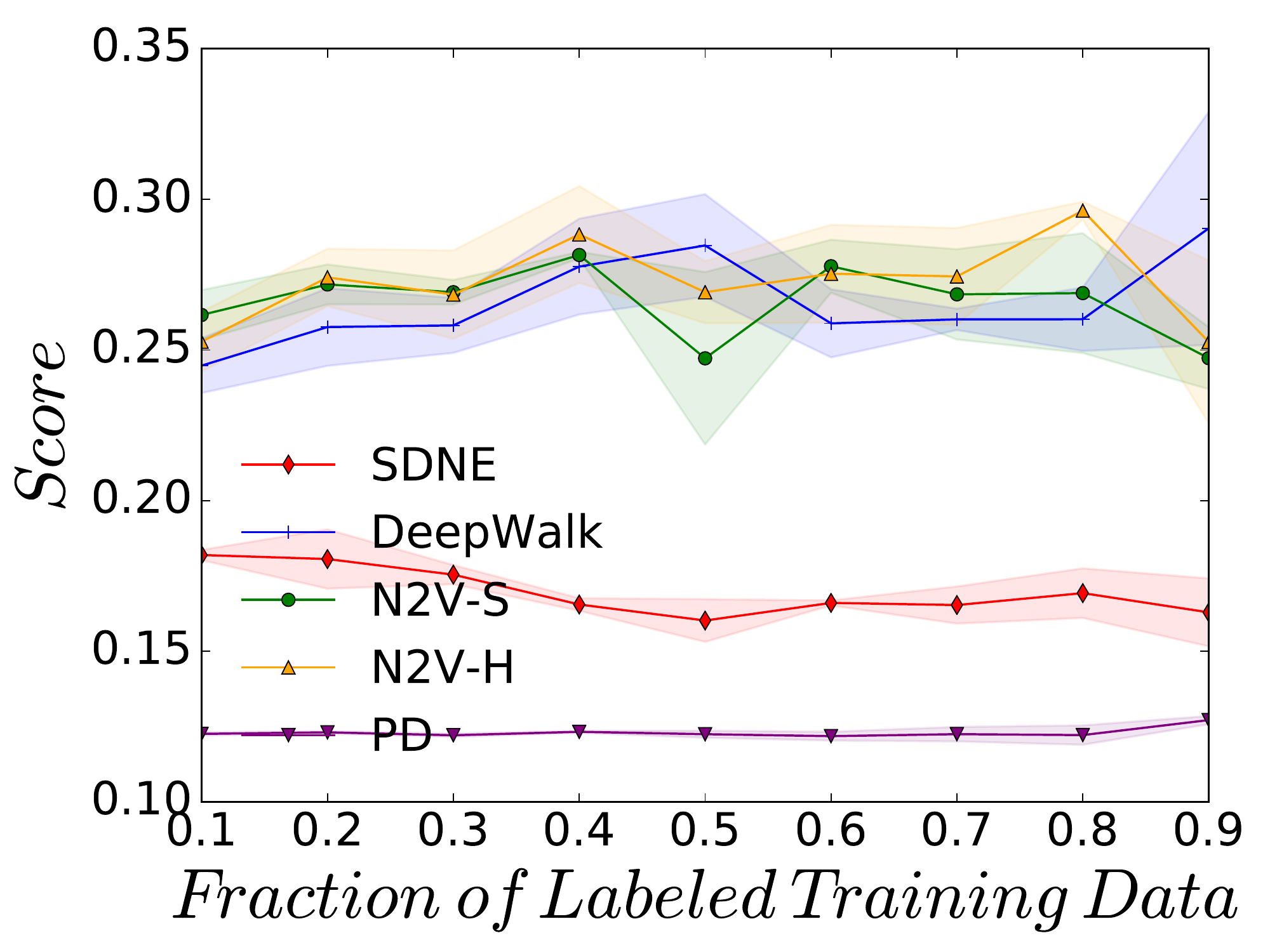}}
  \label{DCmaCA}\hfill
\subfloat[Micro Openflights]{%
    \includegraphics[width=0.25\linewidth]{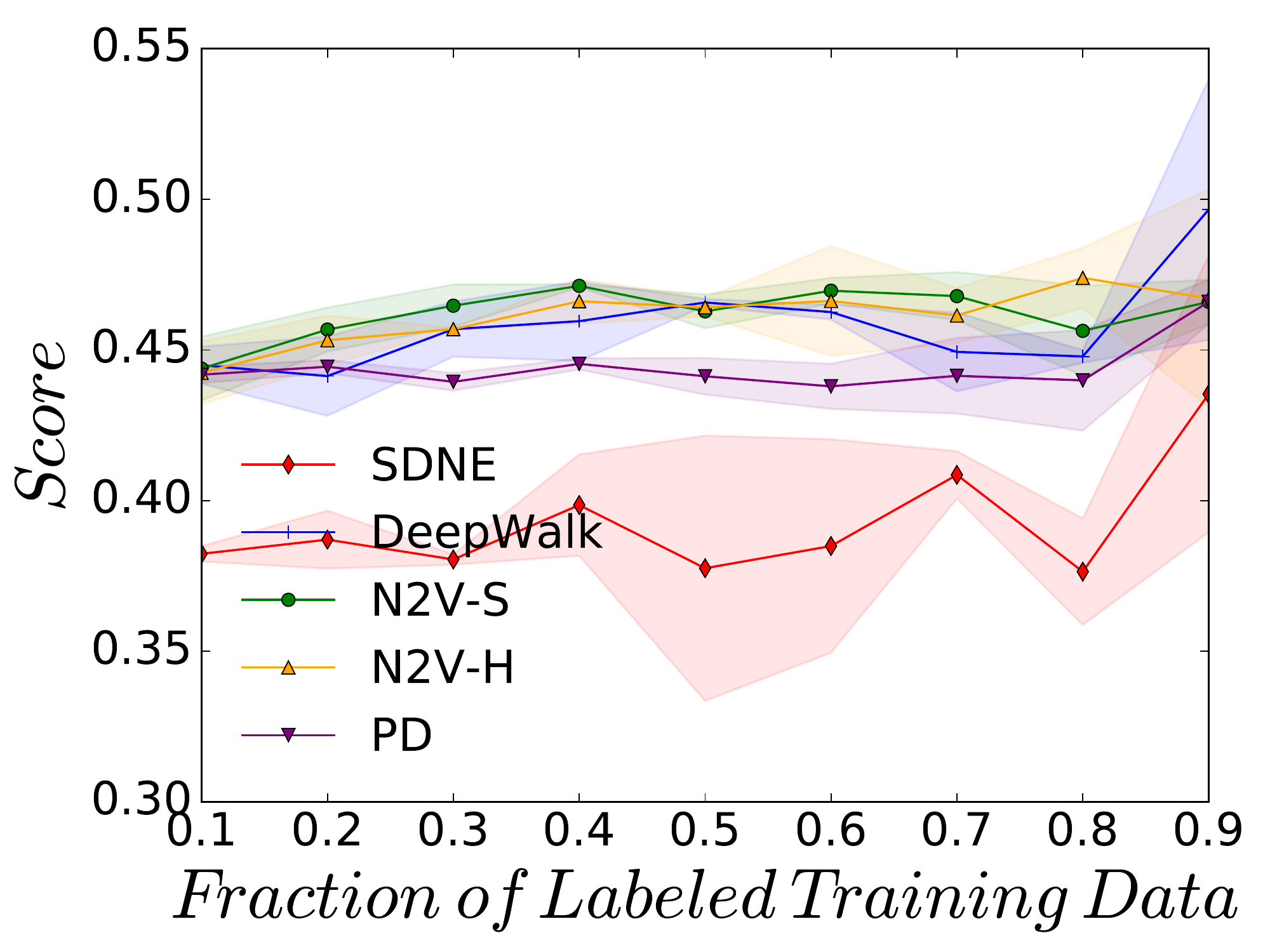}}
  \label{DCmaFB}\hfill
\subfloat[Macro Bitcoinotc]{%
    \includegraphics[width=0.25\linewidth]{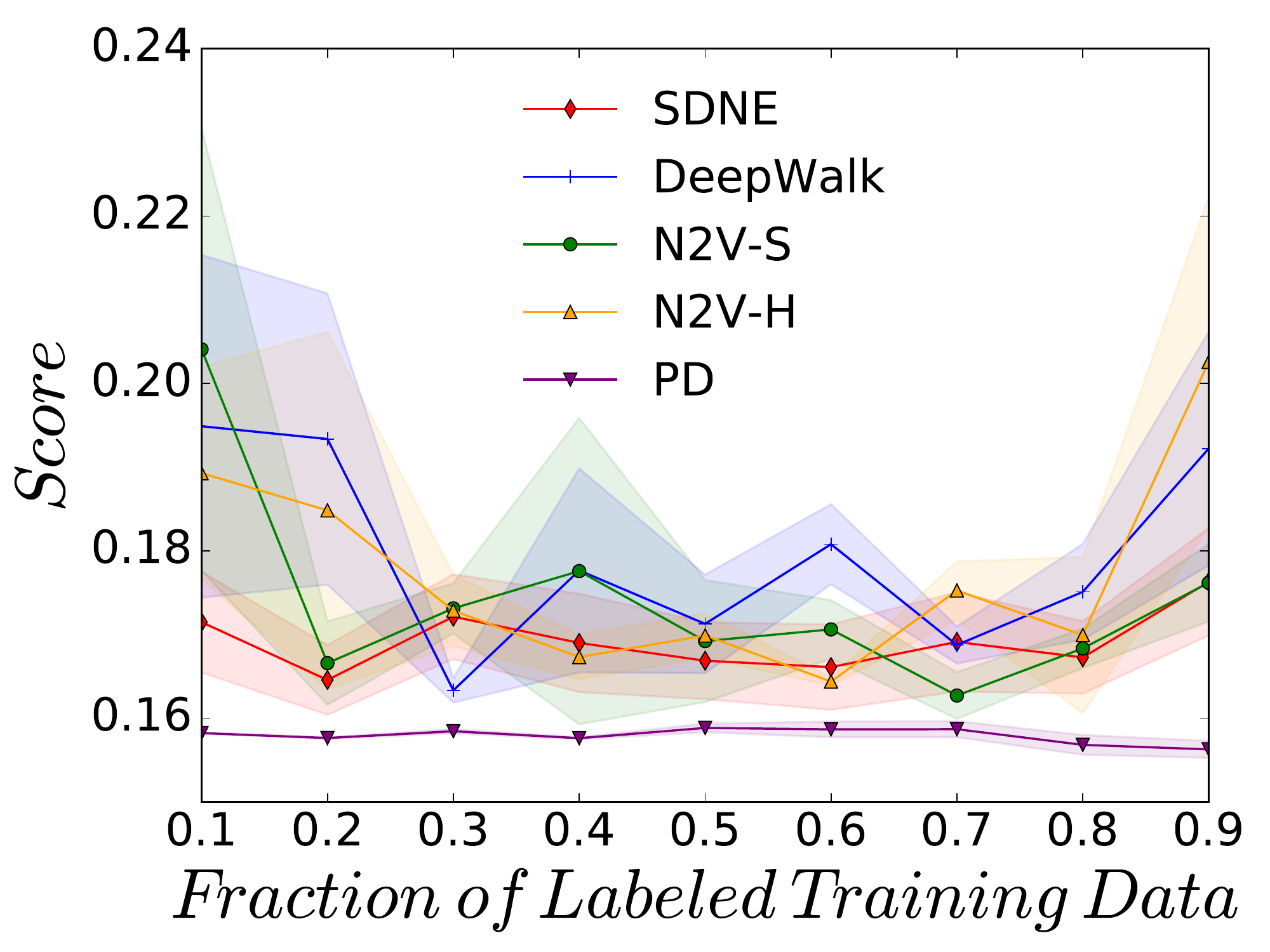}}
  \label{DCmaGN}\hfill
\subfloat[Micro Bitcoinotc]{%
    \includegraphics[width=0.25\linewidth]{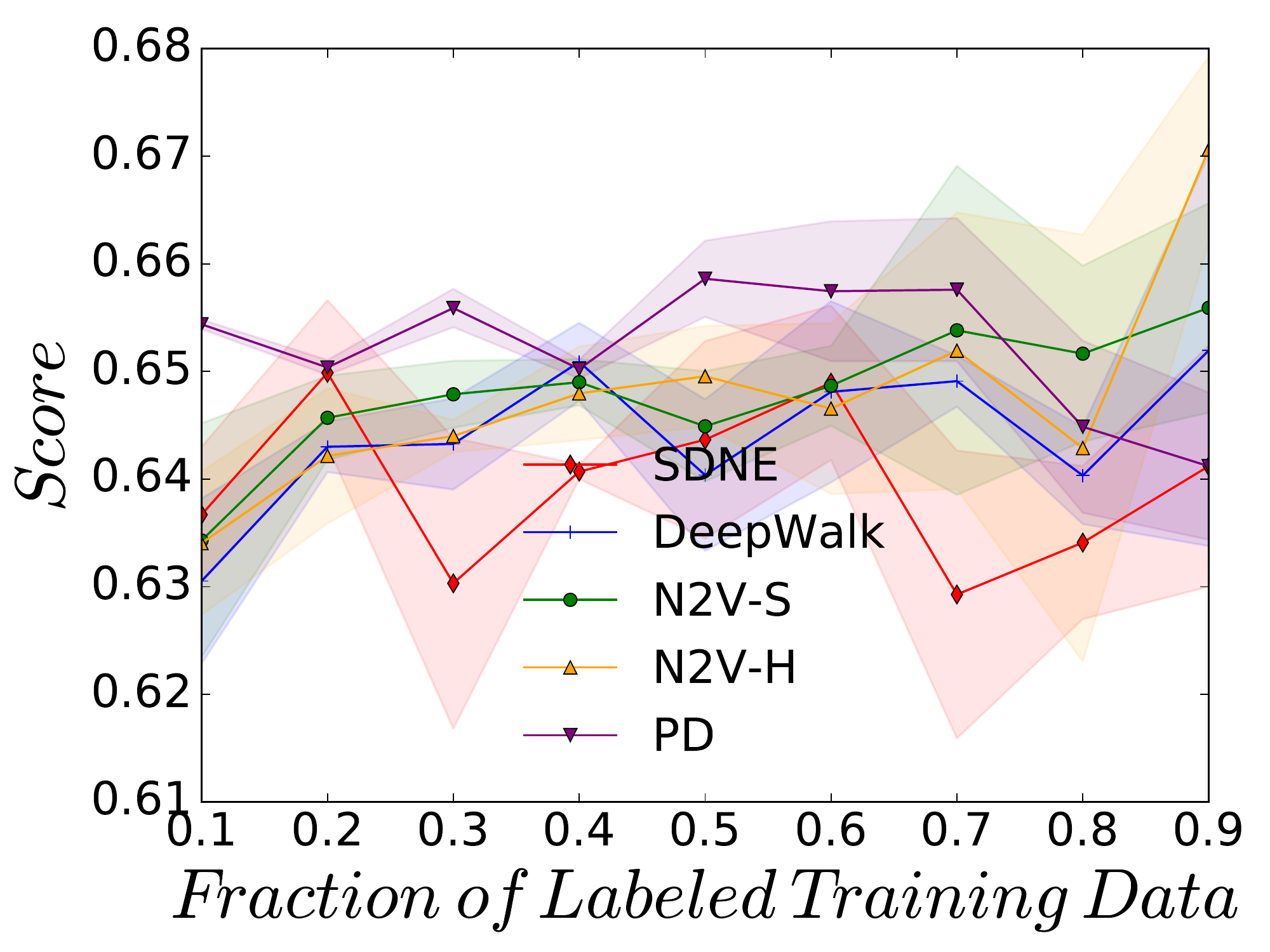}}
  \label{DCmaWI}
\caption{Micro and Macro F1 Scores, across a range of labelling fractions, for all approaches when predicting a vertex's Degree Centrality (DG) Value across all datasets.}
\label{fig:DC_FIG}
\end{figure*}

\begin{figure*}
  \centering
\subfloat[Macro Drosophila]{%
    \includegraphics[width=0.25\linewidth]{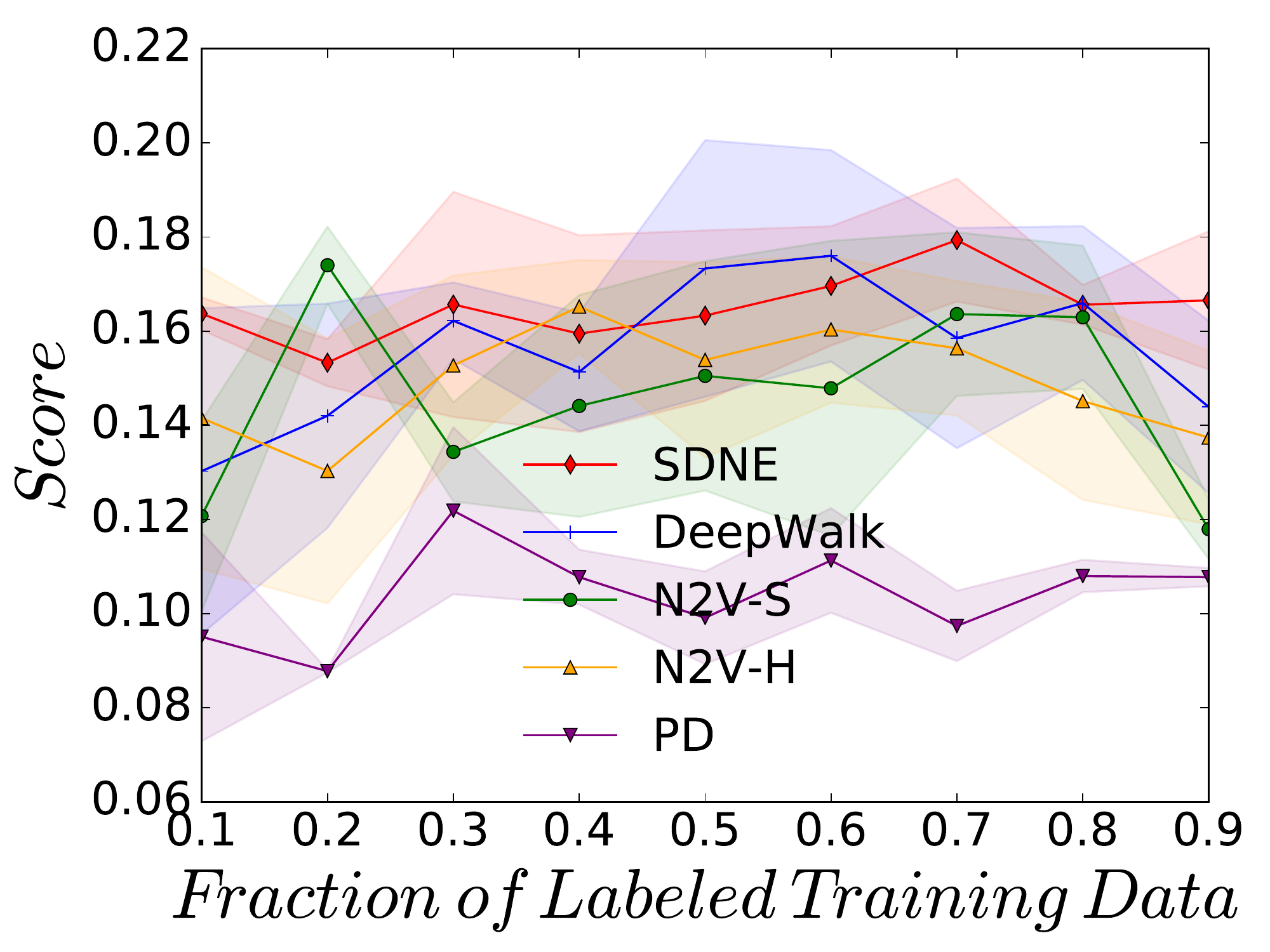}}
  \label{DCmiCA}\hfill
\subfloat[Micro Drosophila]{%
    \includegraphics[width=0.25\linewidth]{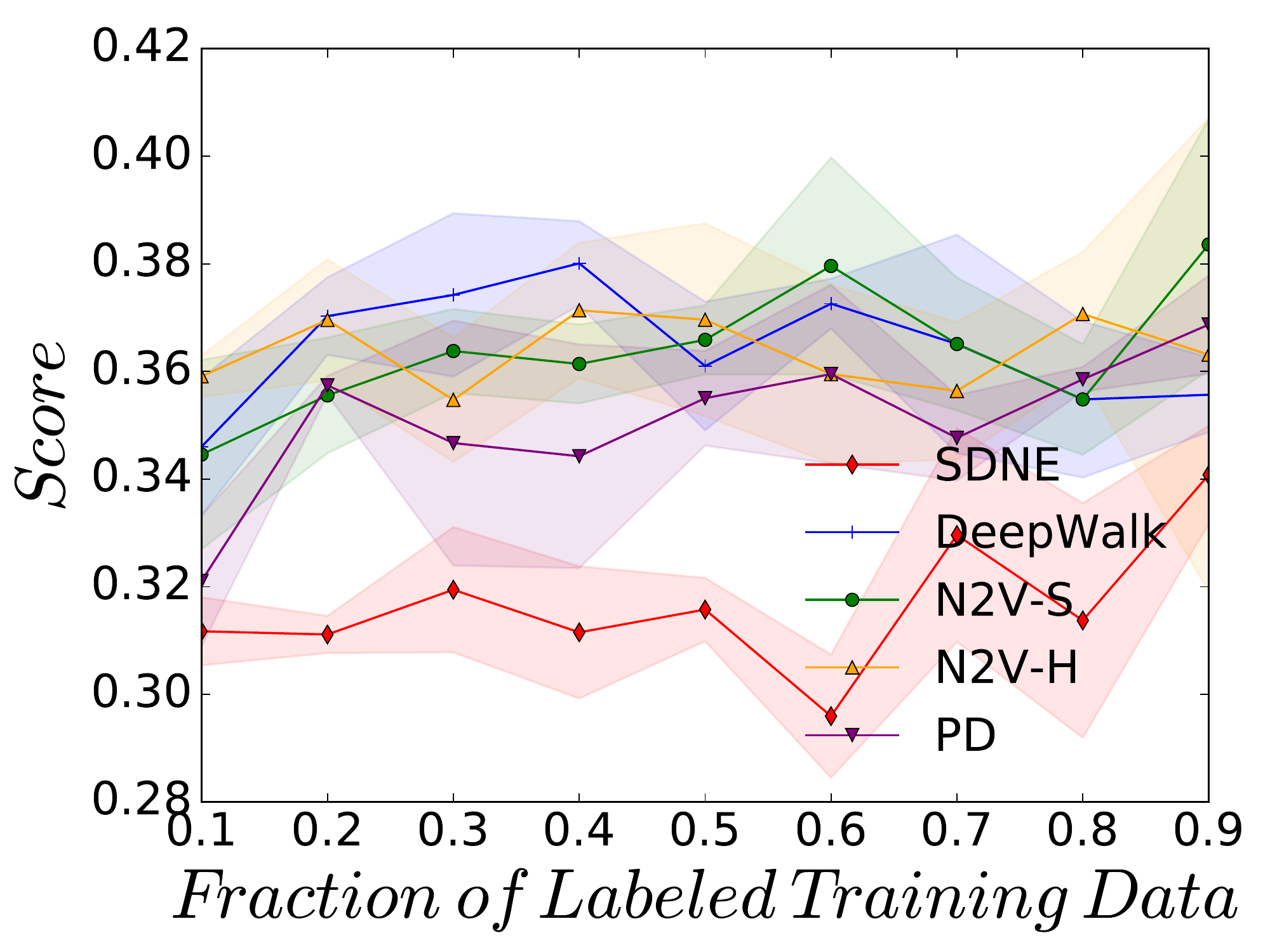}}
  \label{DCmiFB}\hfill
\subfloat[Macro HepTh]{%
    \includegraphics[width=0.25\linewidth]{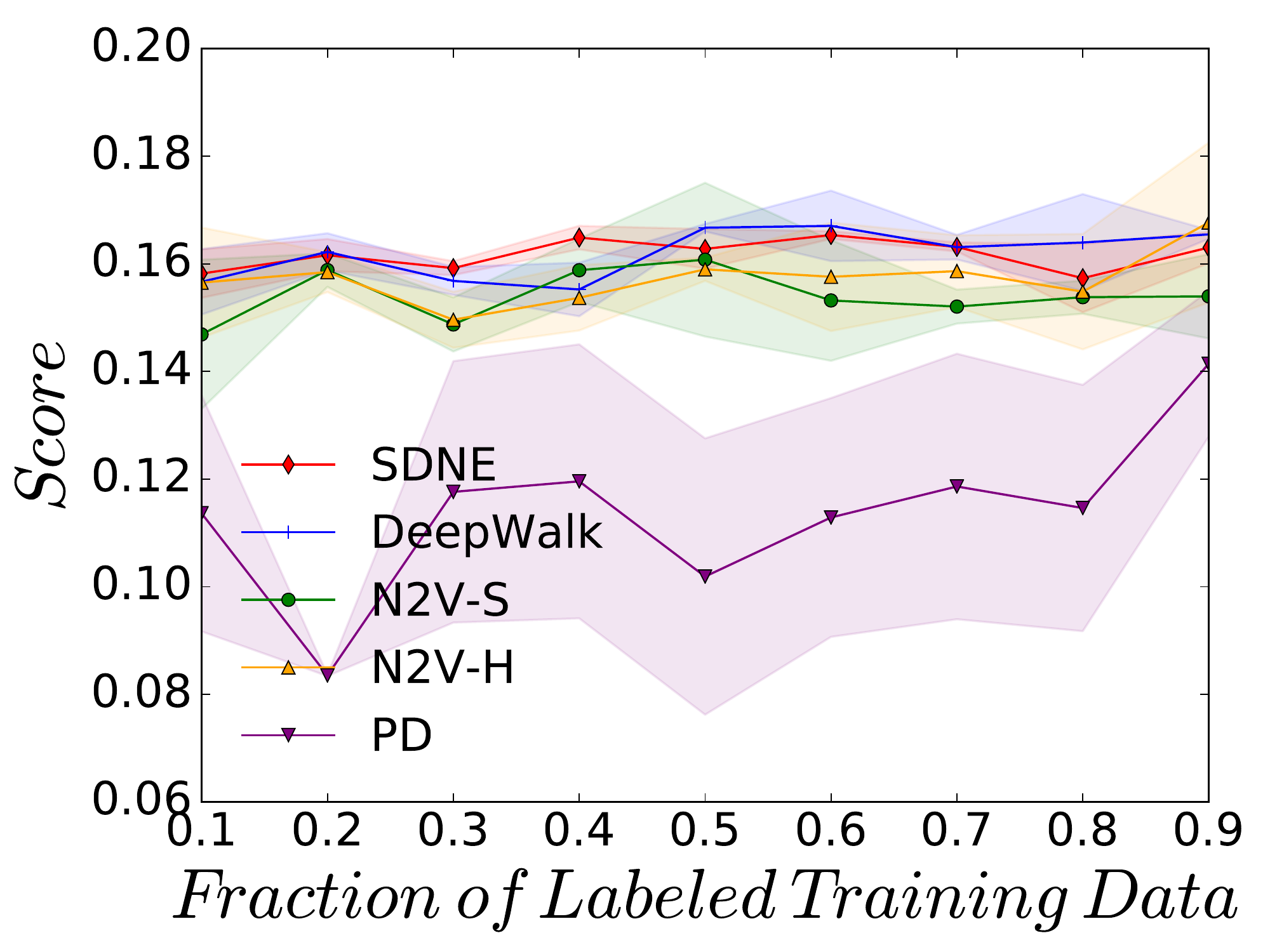}}
  \label{DCmiGN}\hfill
\subfloat[Micro HepTh]{%
    \includegraphics[width=0.25\linewidth]{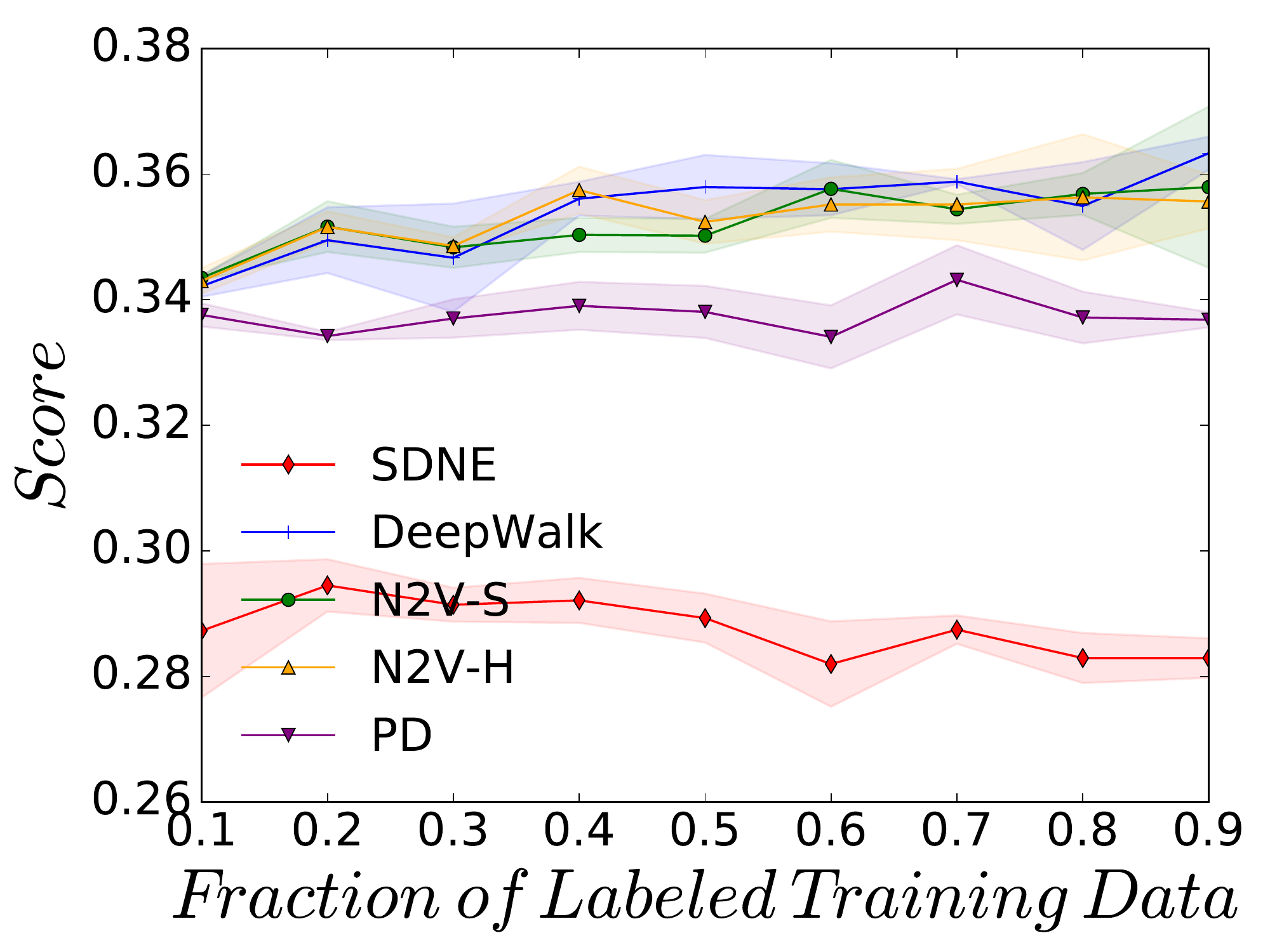}}
  \label{DCmiWI}\\

\subfloat[Macro Email-EU]{%
    \includegraphics[width=0.25\linewidth]{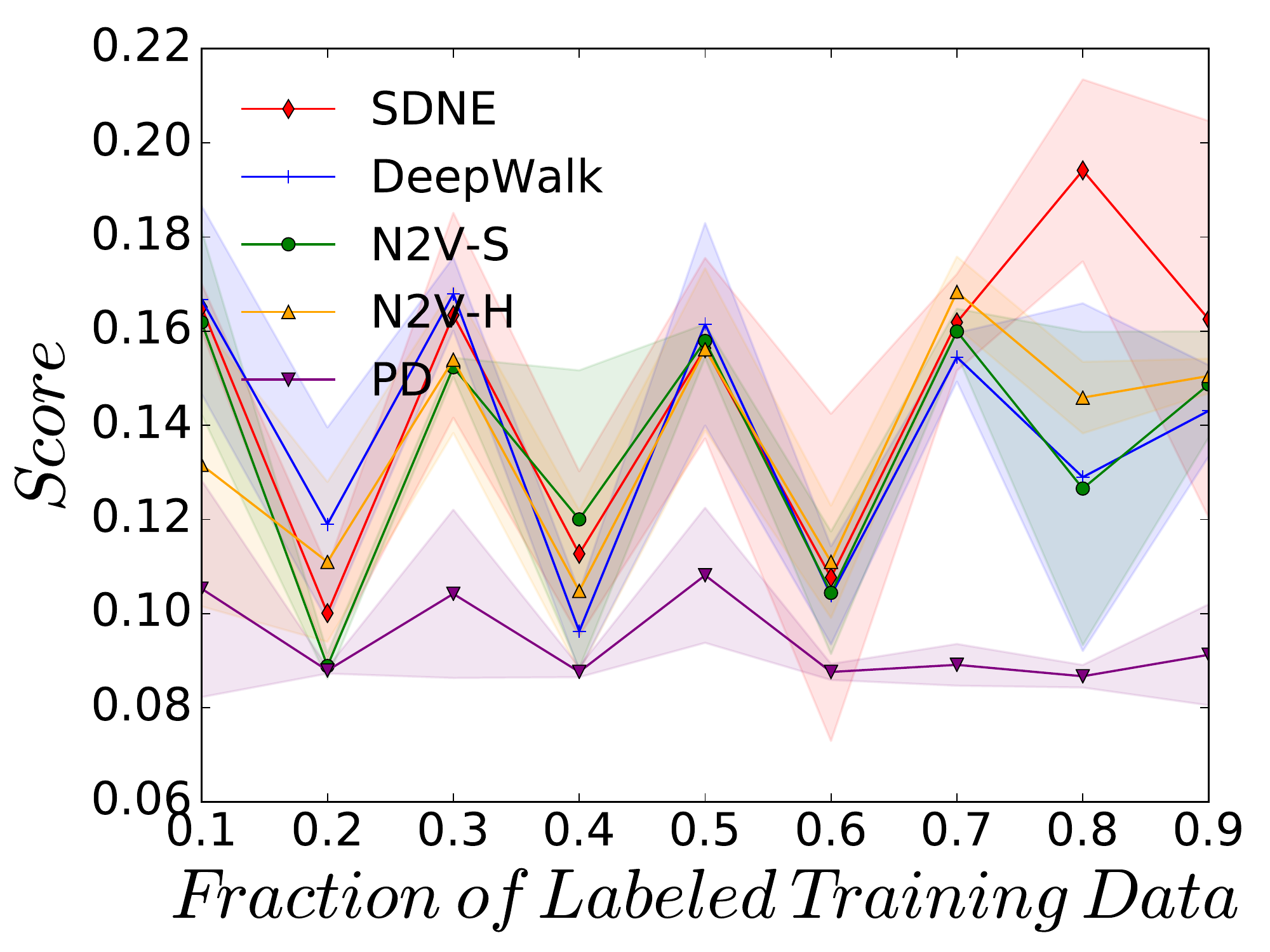}}
  \label{DCmaCA}\hfill
\subfloat[Micro Email-EU]{%
    \includegraphics[width=0.25\linewidth]{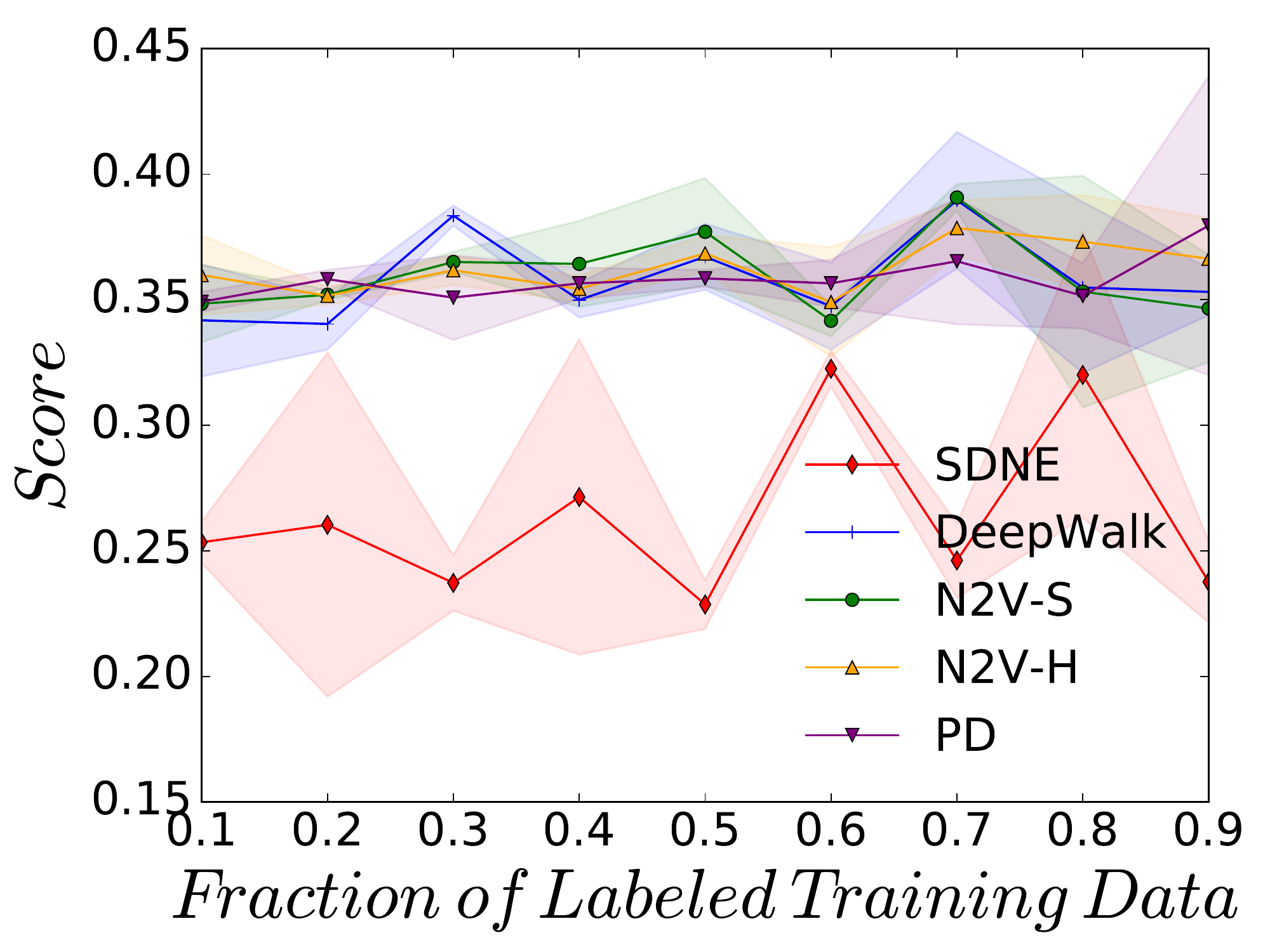}}
  \label{DCmaFB}\hfill
\subfloat[Macro Facebook]{%
    \includegraphics[width=0.25\linewidth]{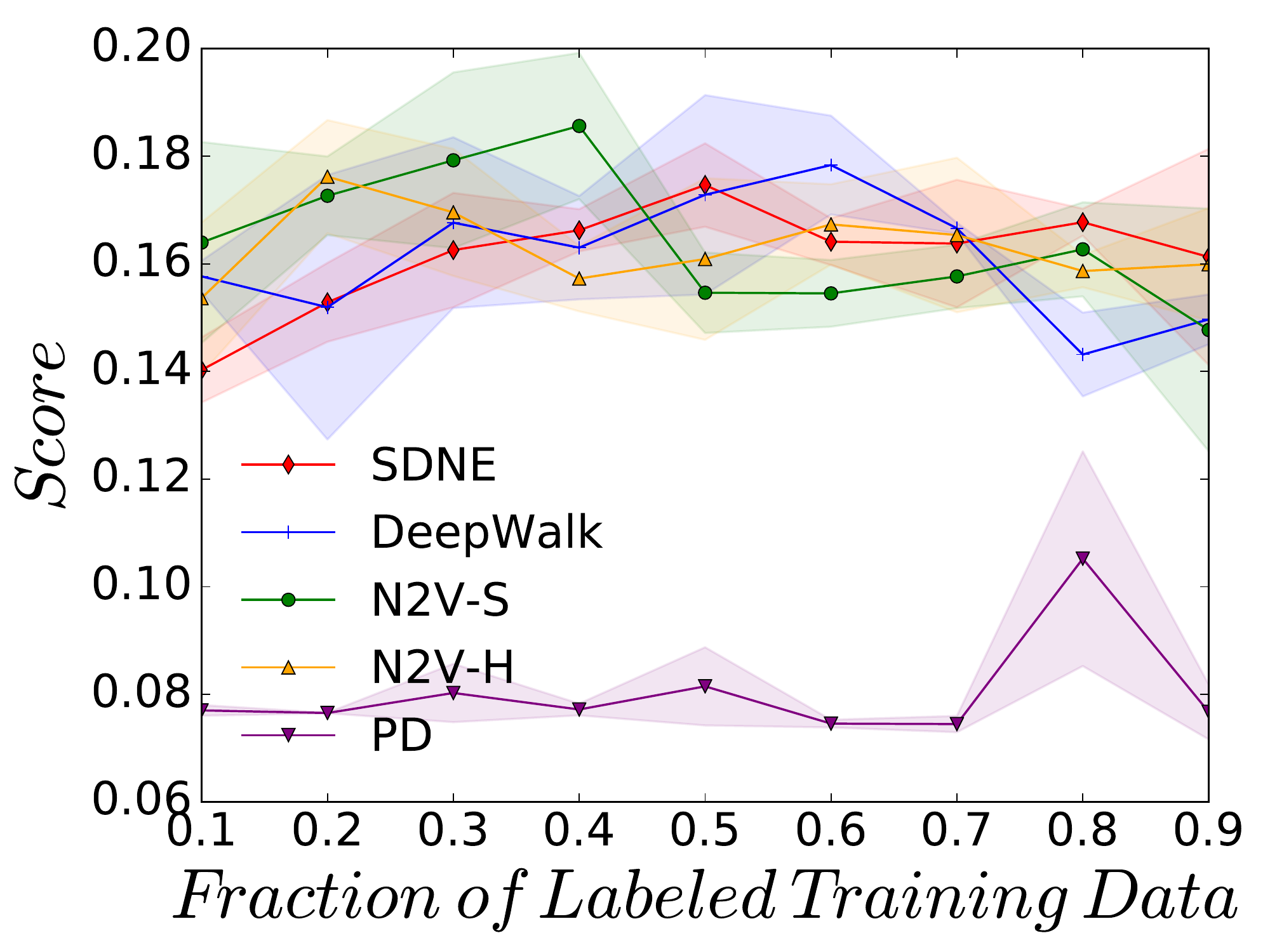}}
  \label{DCmaGN}\hfill
\subfloat[Micro Facebook]{%
    \includegraphics[width=0.25\linewidth]{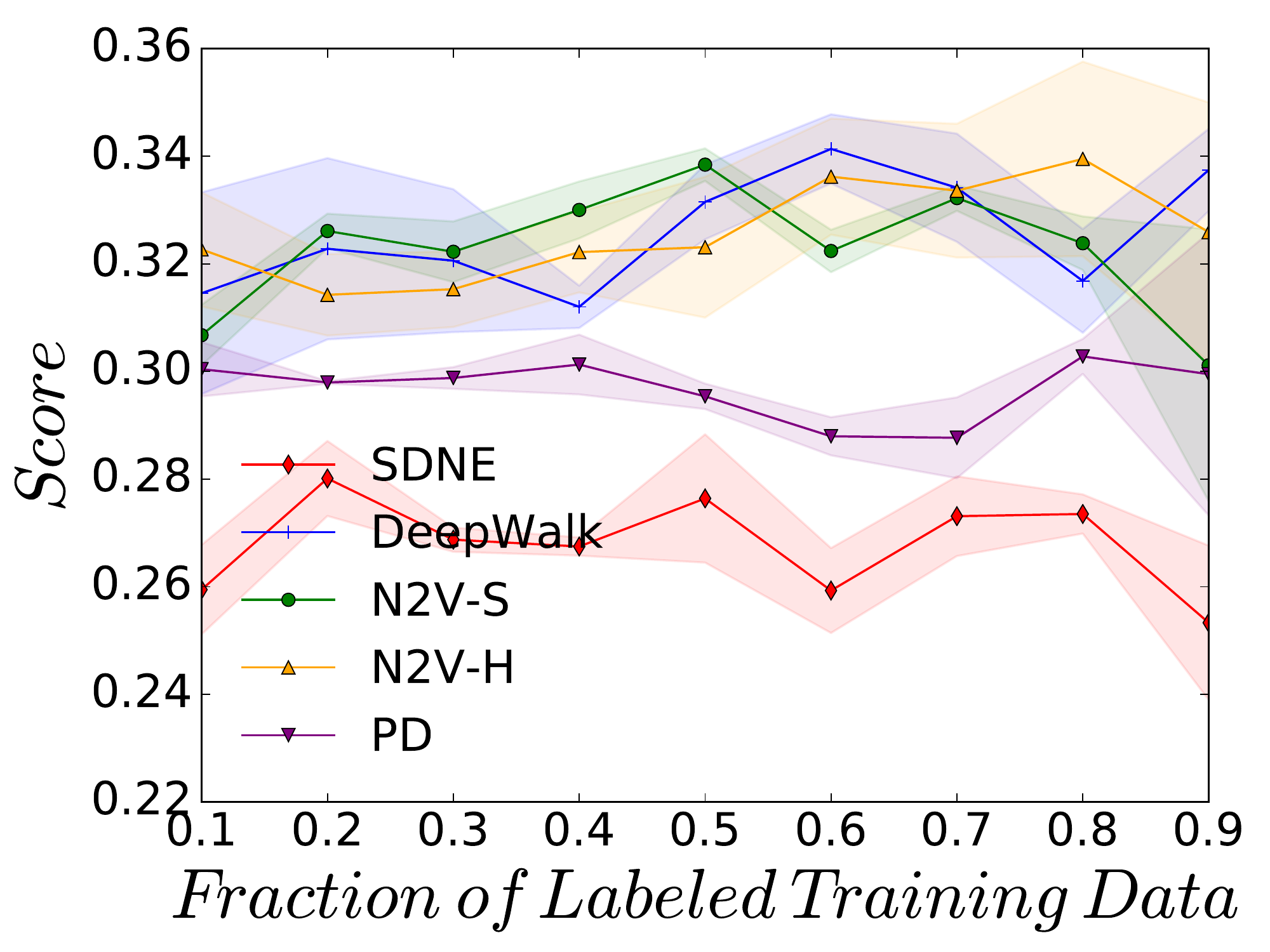}}
  \label{DCmaWI} \\ 

  \subfloat[Macro Openflights]{%
    \includegraphics[width=0.25\linewidth]{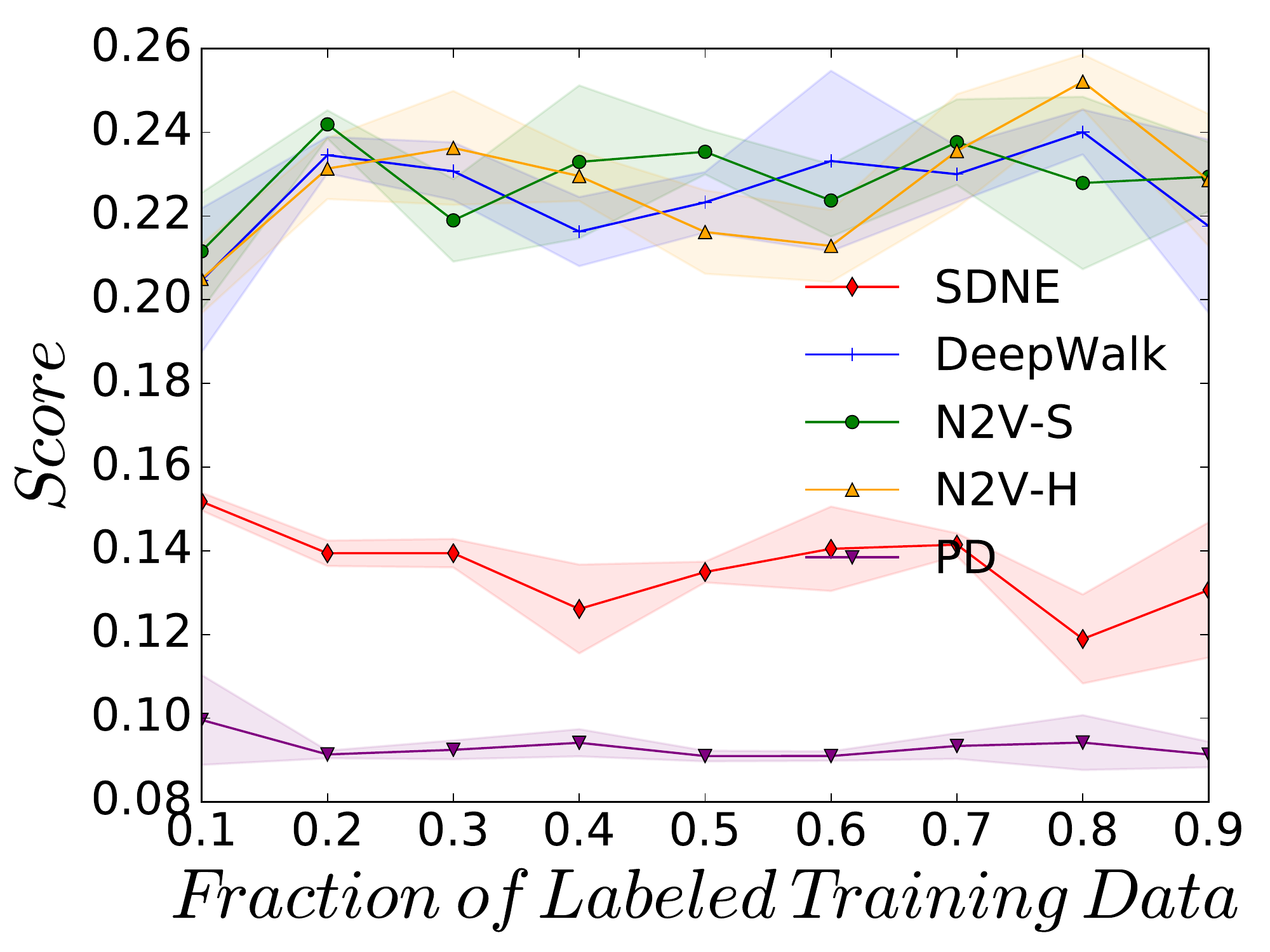}}
  \label{DCmaCA}\hfill
\subfloat[Micro Openflights]{%
    \includegraphics[width=0.25\linewidth]{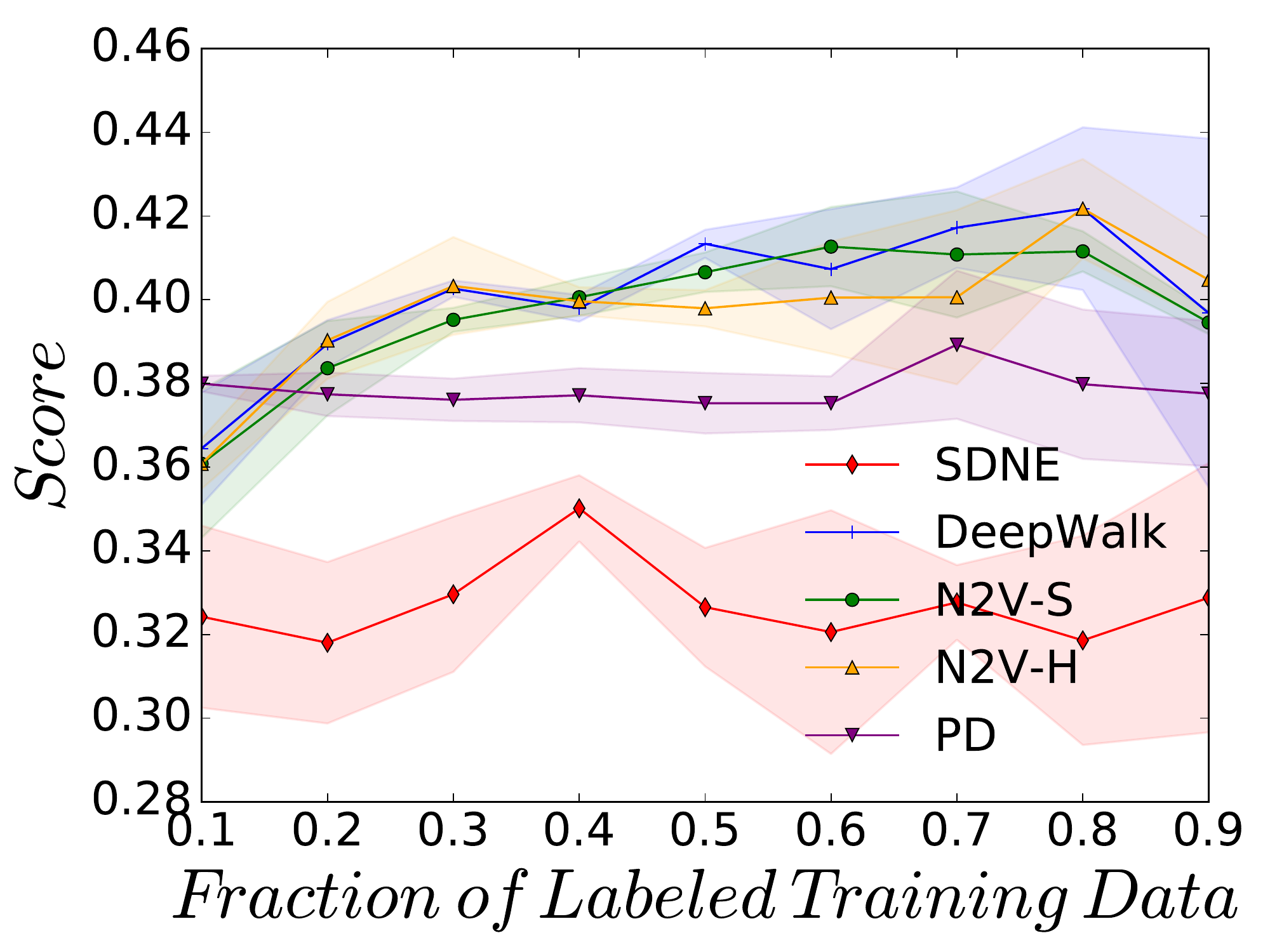}}
  \label{DCmaFB}\hfill
\subfloat[Macro Bitcoinotc]{%
    \includegraphics[width=0.25\linewidth]{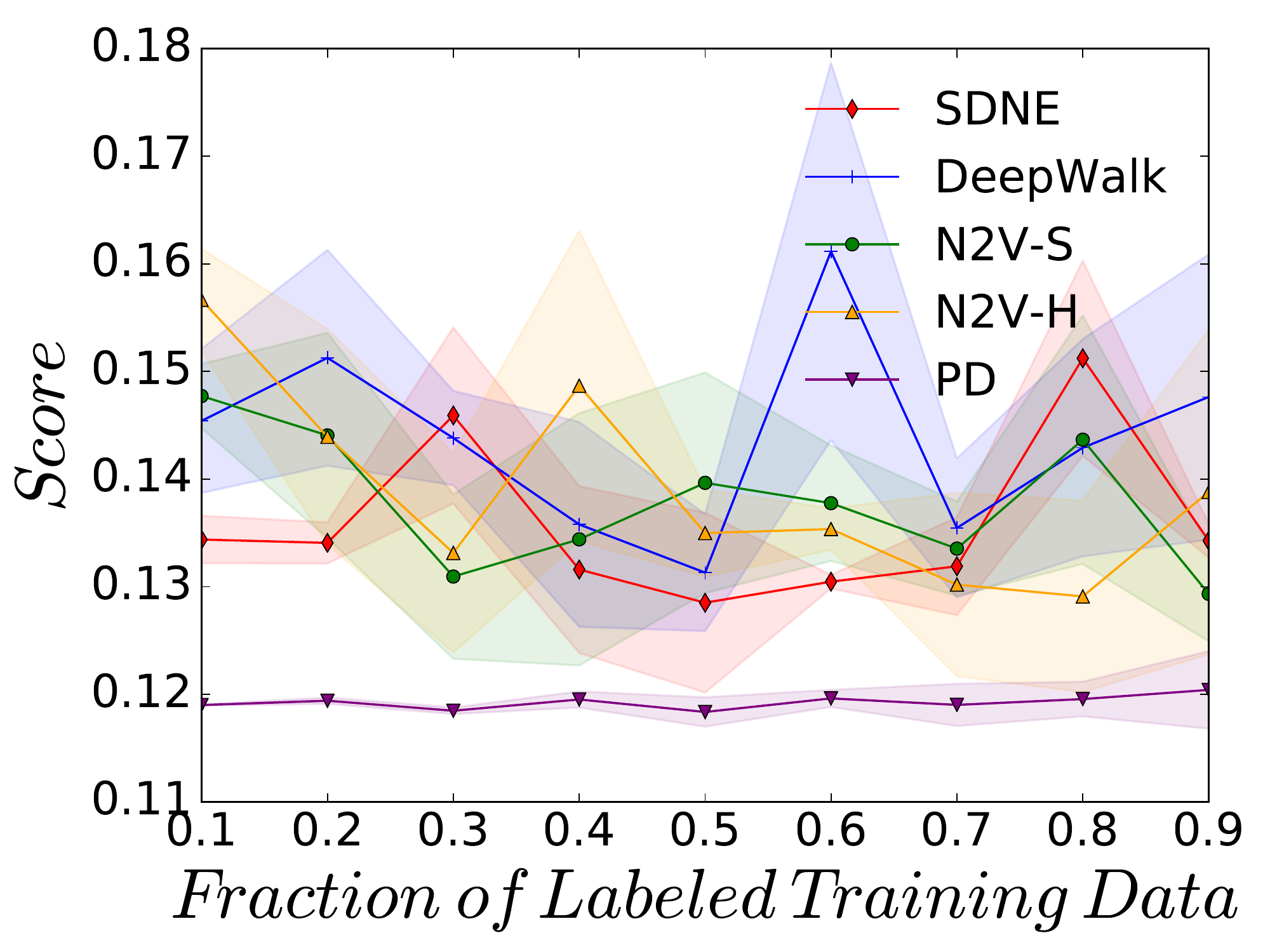}}
  \label{DCmaGN}\hfill
\subfloat[Micro Bitcoinotc]{%
    \includegraphics[width=0.25\linewidth]{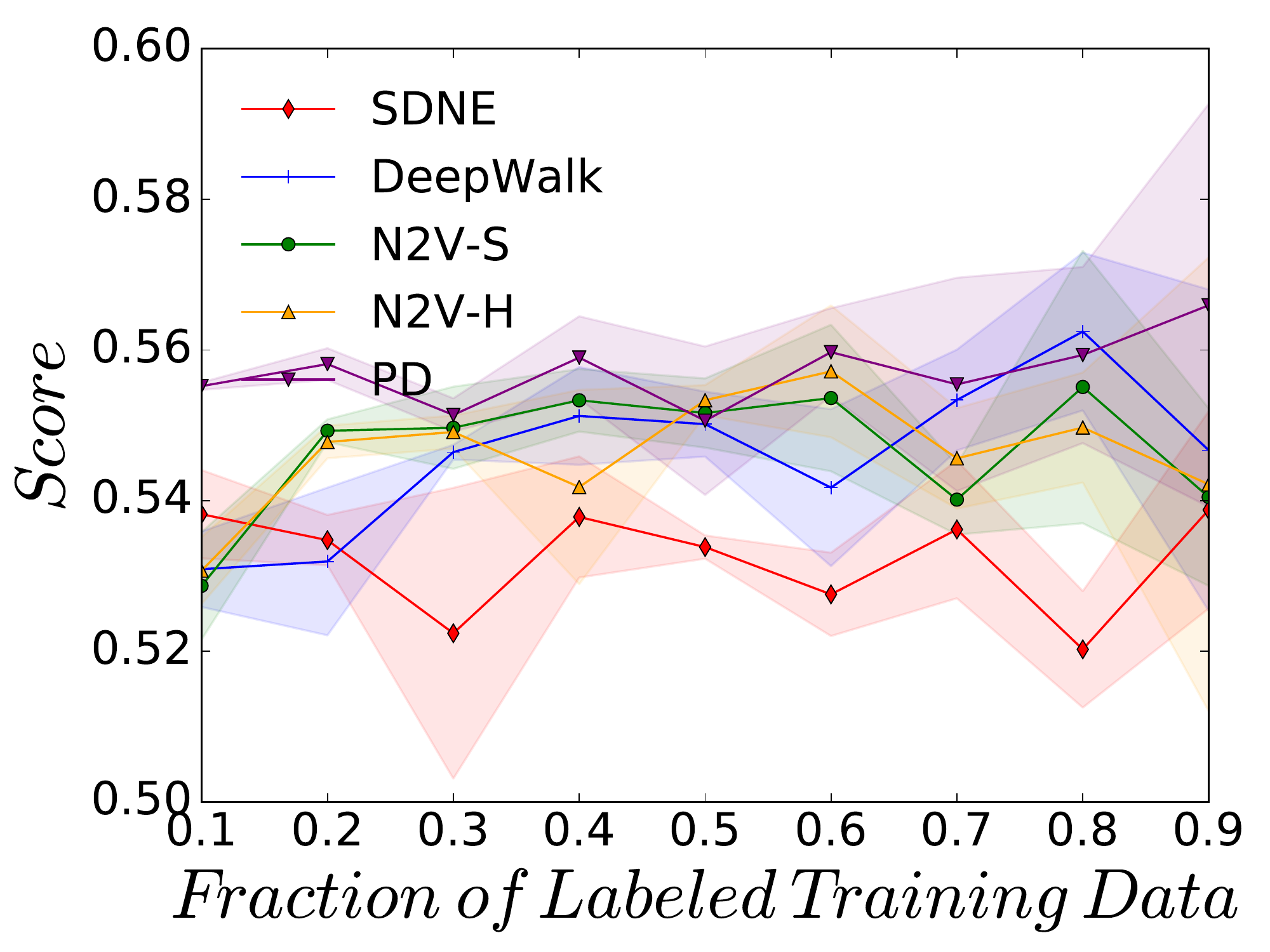}}
  \label{DCmaWI}
\caption{Micro and Macro F1 Scores, across a range of labelling fractions, for all approaches when predicting a vertex's Triangle Count (TR) Value across all datasets.}
\label{fig:TR_FIG}
\end{figure*}

\begin{figure*}
  \centering
\subfloat[Macro Drosophila]{%
    \includegraphics[width=0.25\linewidth]{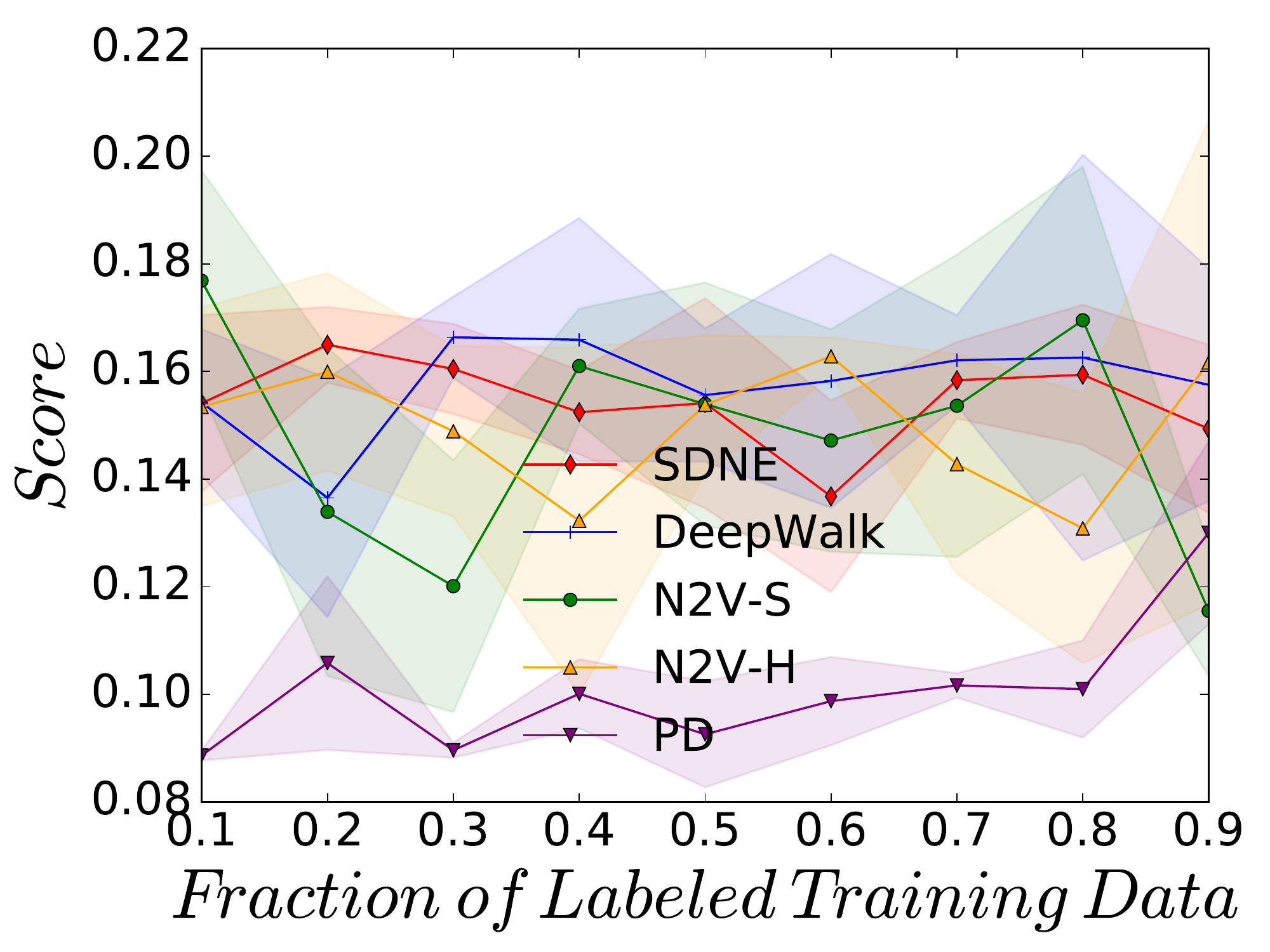}}
  \label{DCmiCA}\hfill
\subfloat[Micro Drosophila]{%
    \includegraphics[width=0.25\linewidth]{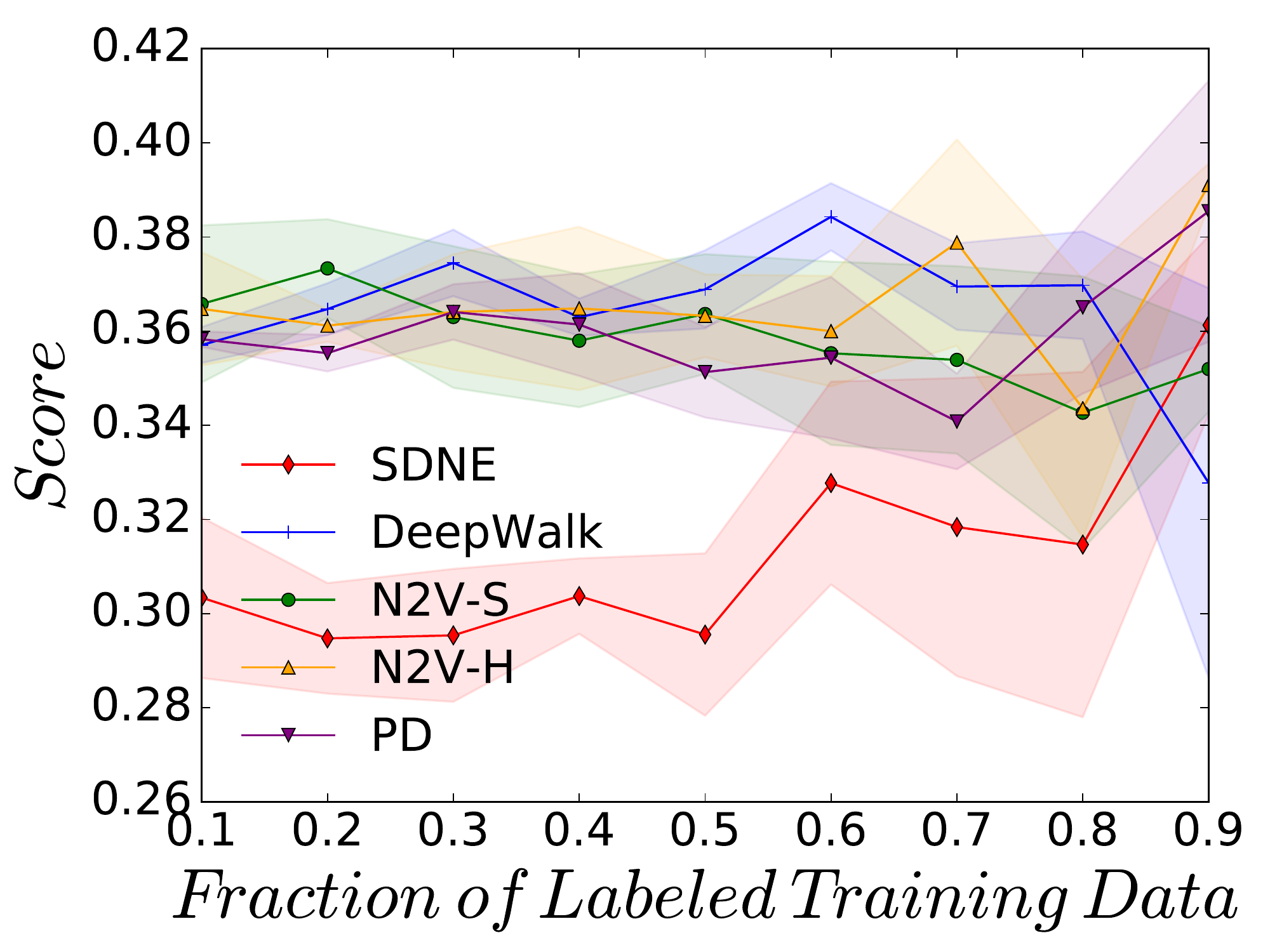}}
  \label{DCmiFB}\hfill
\subfloat[Macro HepTh]{%
    \includegraphics[width=0.25\linewidth]{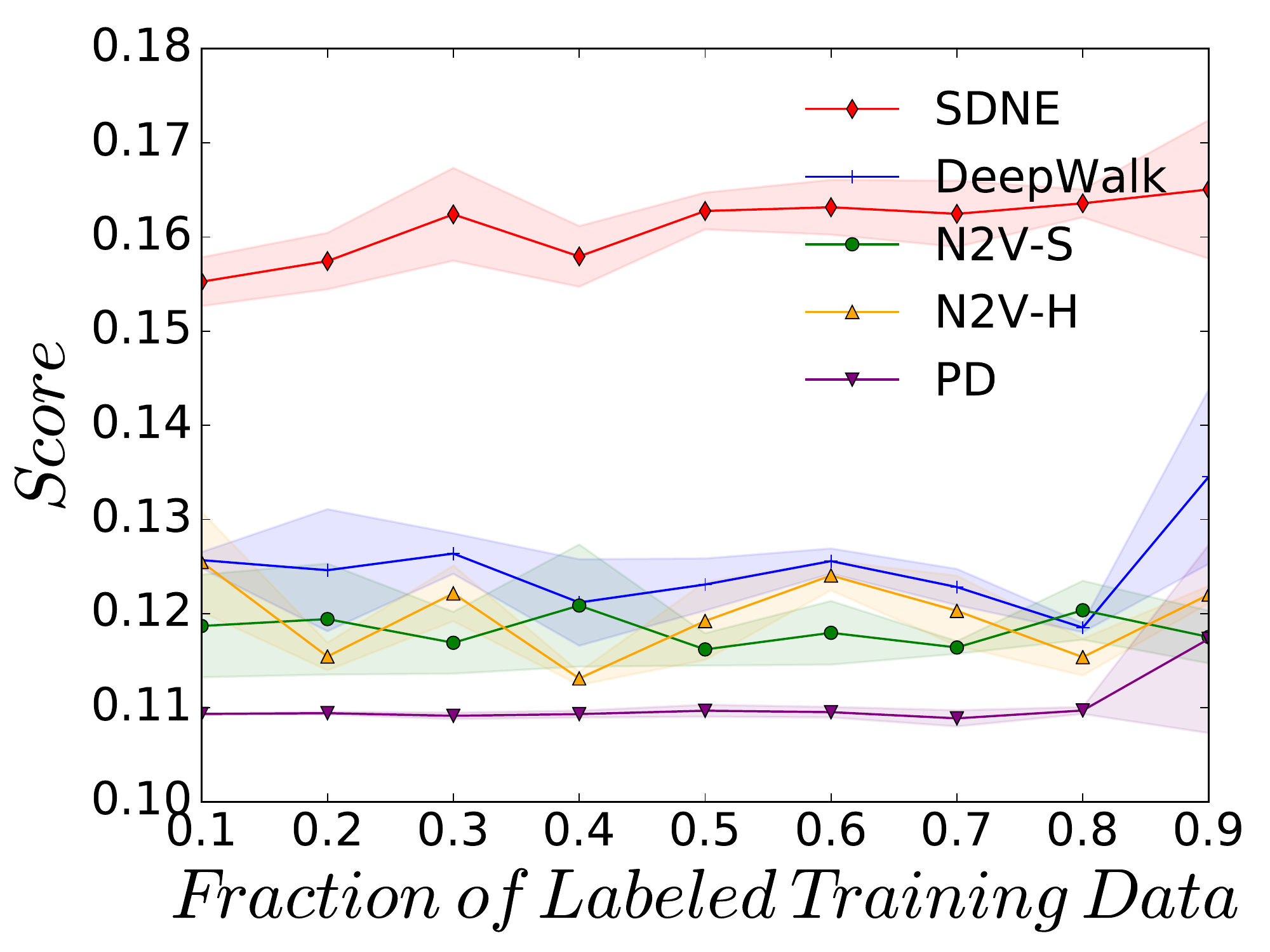}}
  \label{DCmiGN}\hfill
\subfloat[Micro HepTh]{%
    \includegraphics[width=0.25\linewidth]{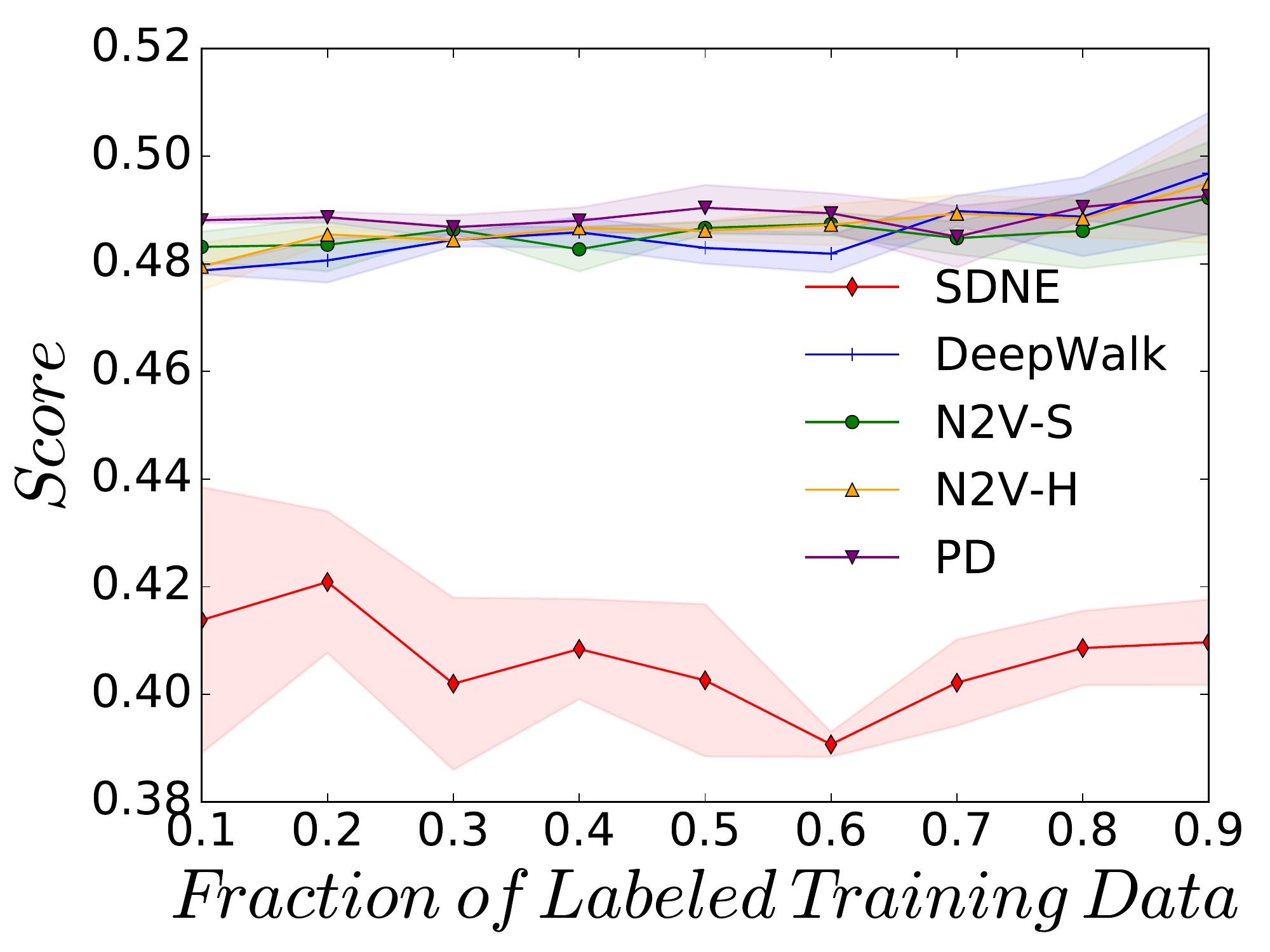}}
  \label{DCmiWI}\\

\subfloat[Macro Email-EU]{%
    \includegraphics[width=0.25\linewidth]{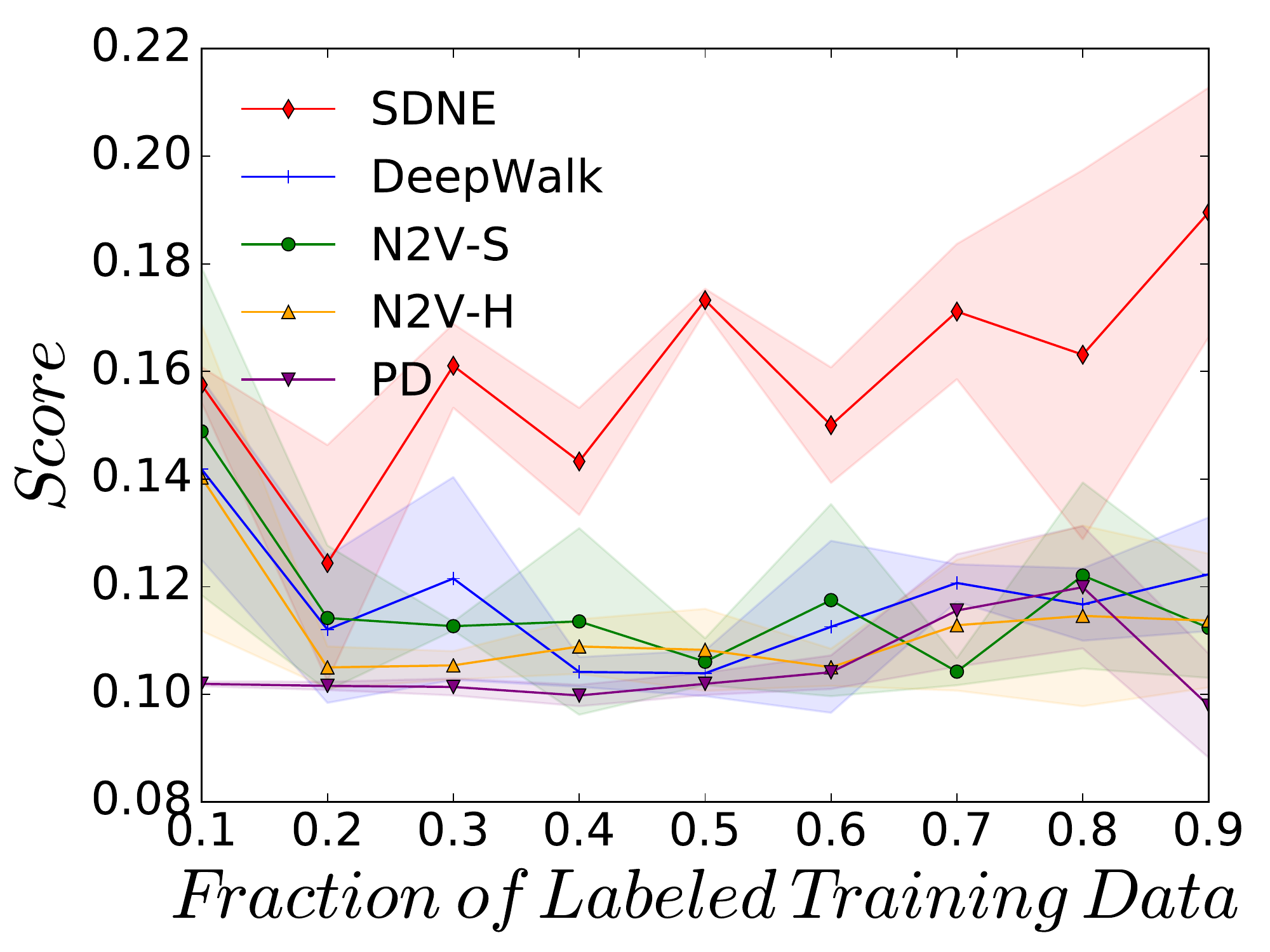}}
  \label{DCmaCA}\hfill
\subfloat[Micro Email-EU]{%
    \includegraphics[width=0.25\linewidth]{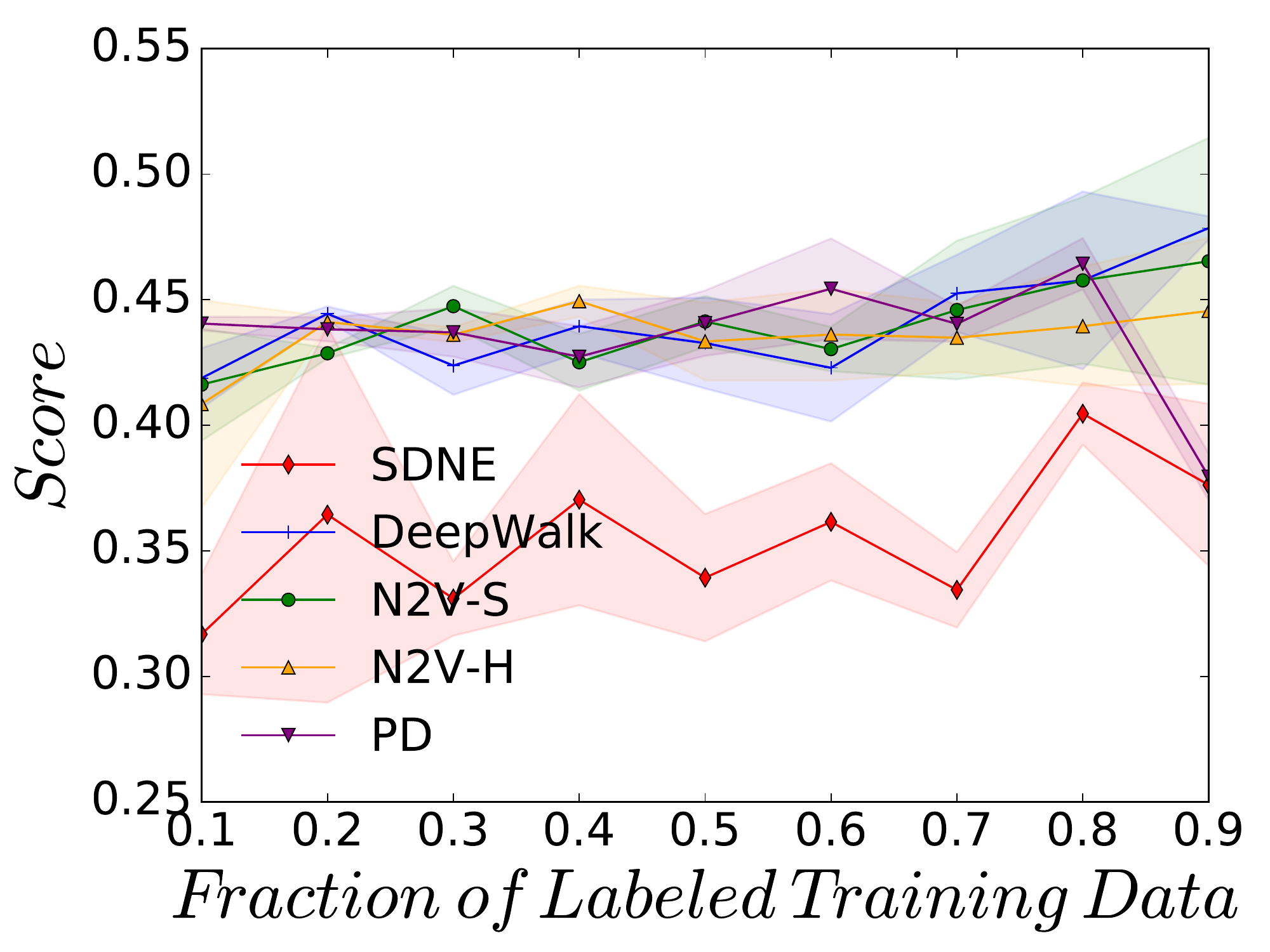}}
  \label{DCmaFB}\hfill
\subfloat[Macro Facebook]{%
    \includegraphics[width=0.25\linewidth]{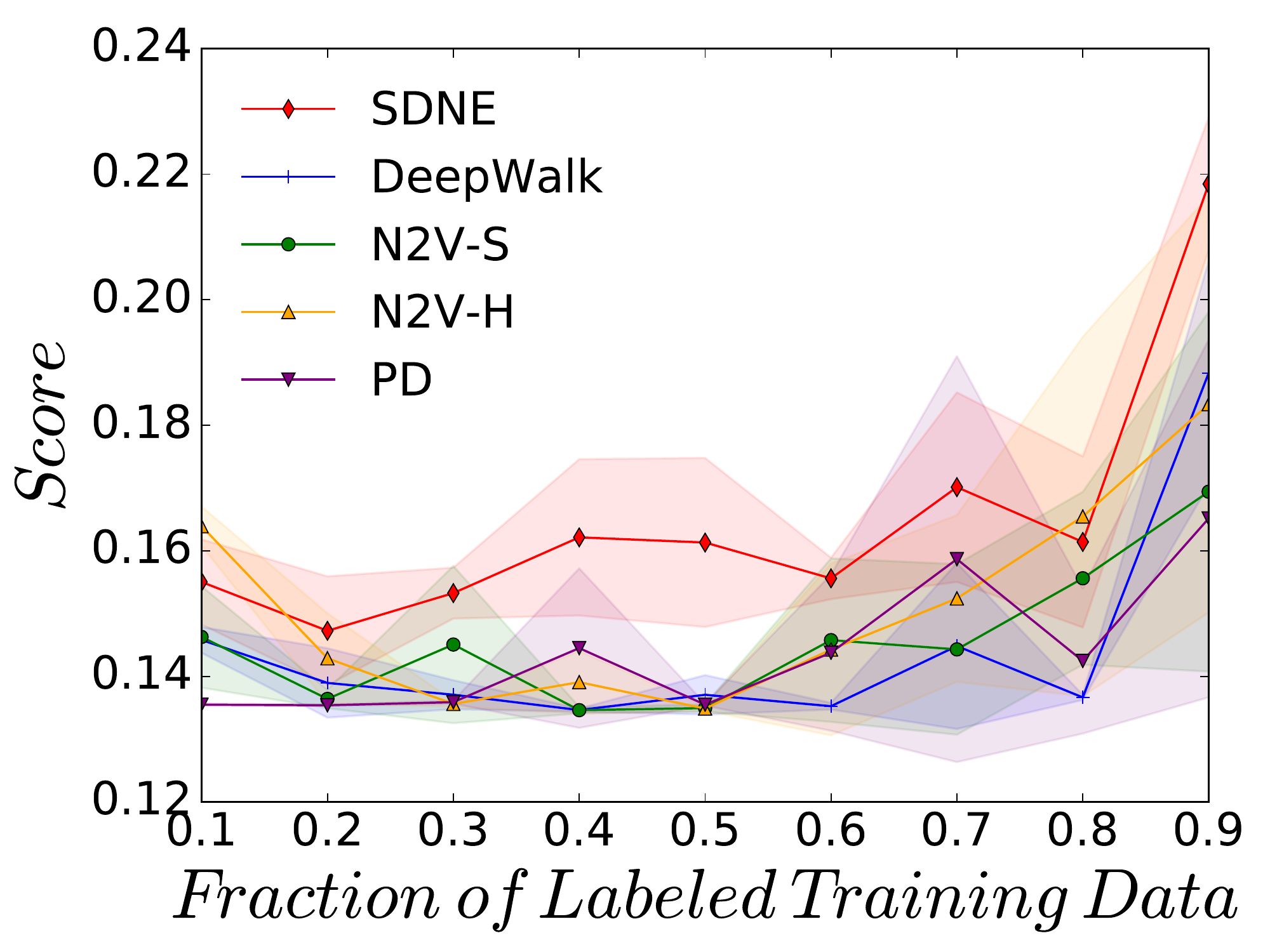}}
  \label{DCmaGN}\hfill
\subfloat[Micro Facebook]{%
    \includegraphics[width=0.25\linewidth]{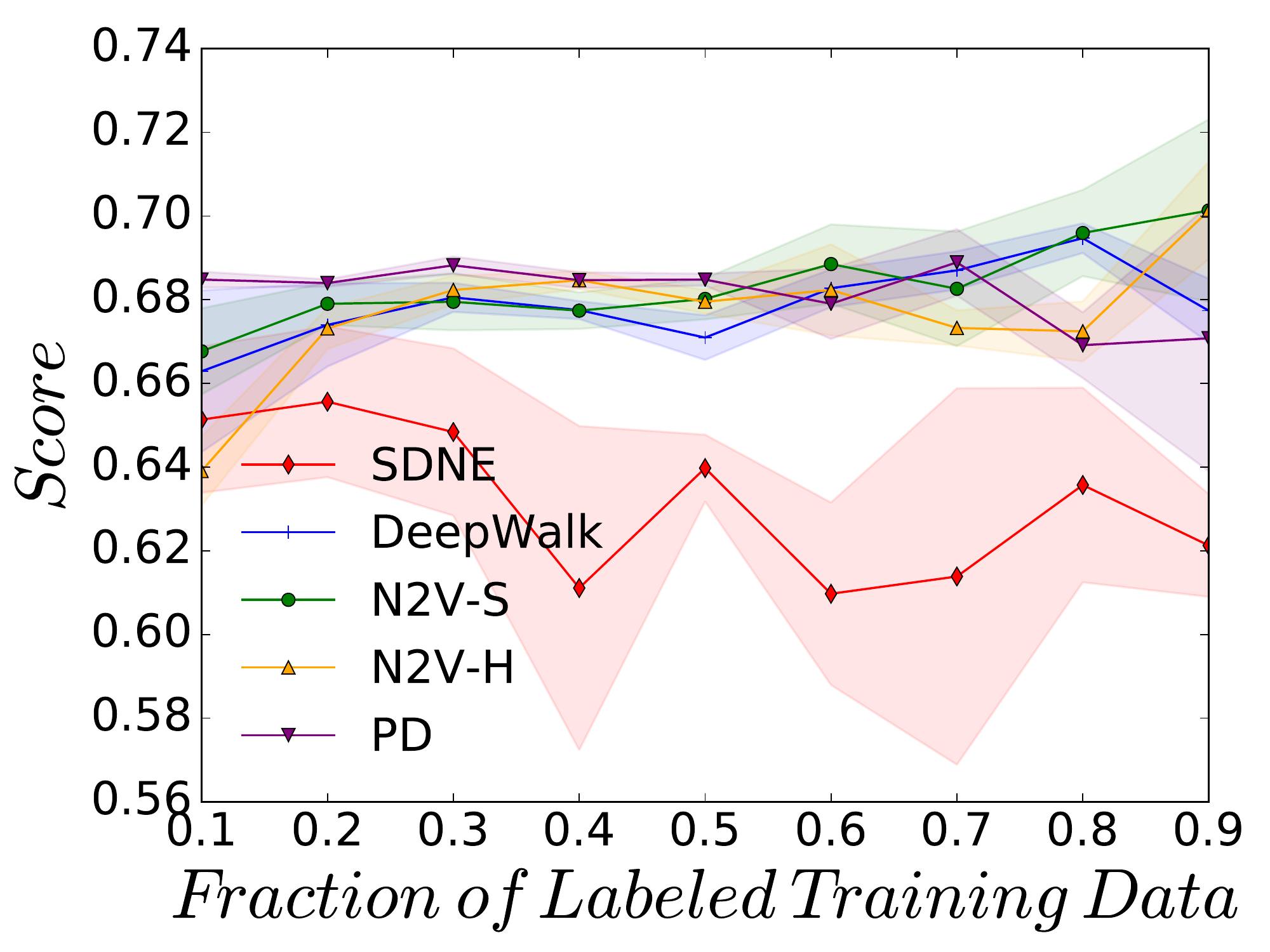}}
  \label{DCmaWI} \\ 

  \subfloat[Macro Openflights]{%
    \includegraphics[width=0.25\linewidth]{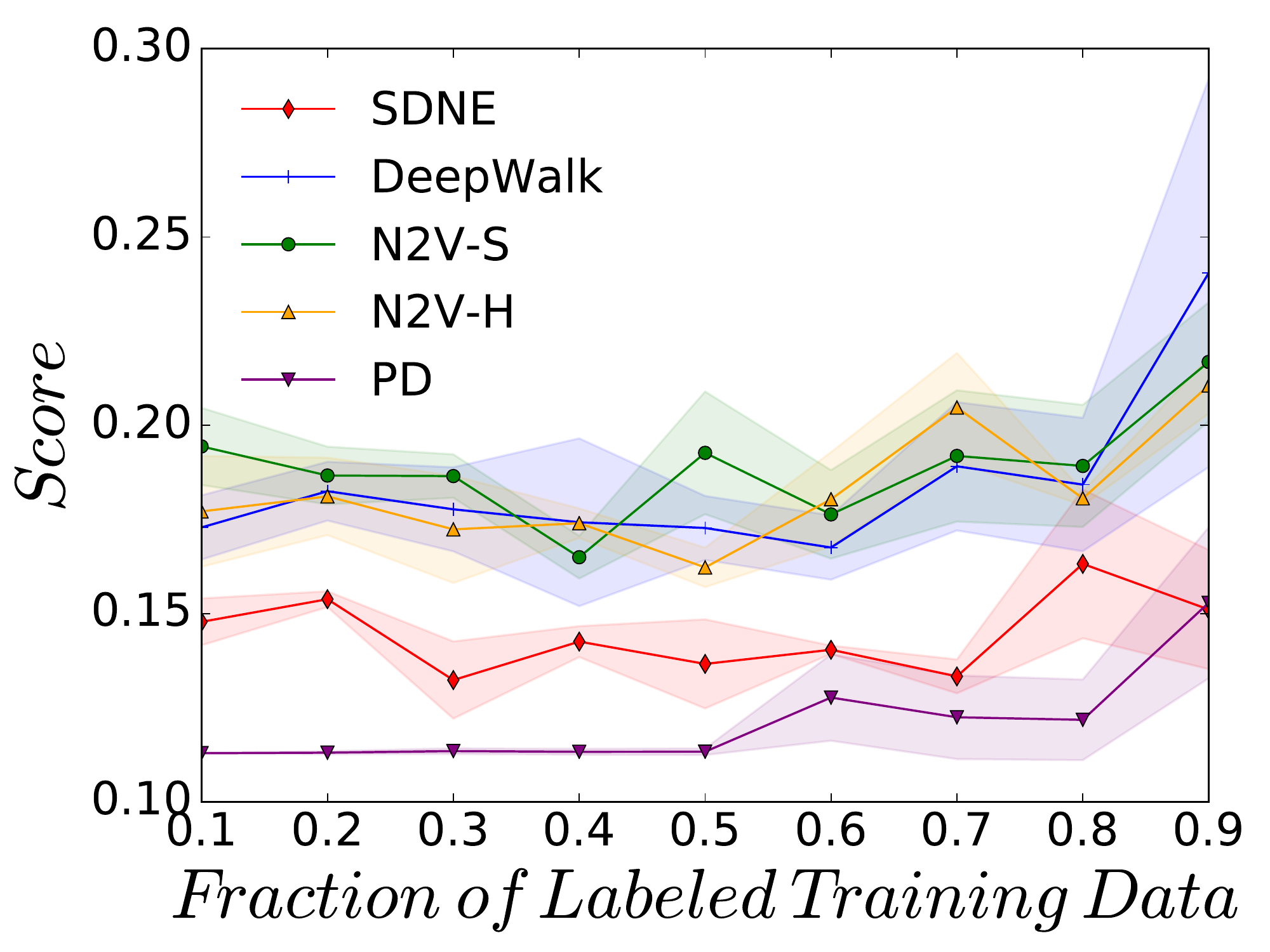}}
  \label{DCmaCA}\hfill
\subfloat[Micro Openflights]{%
    \includegraphics[width=0.25\linewidth]{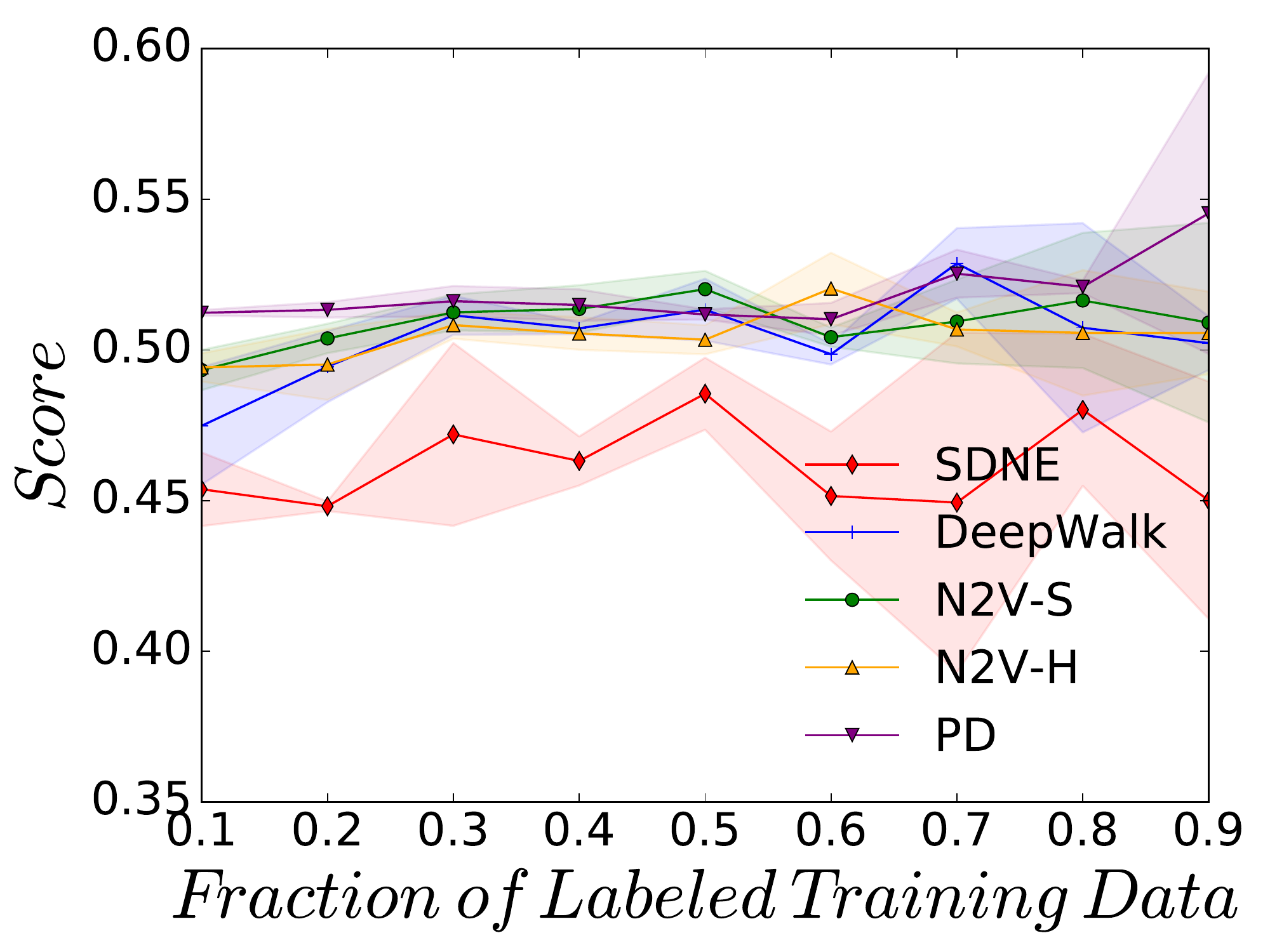}}
  \label{DCmaFB}\hfill
\subfloat[Macro Bitcoinotc]{%
    \includegraphics[width=0.25\linewidth]{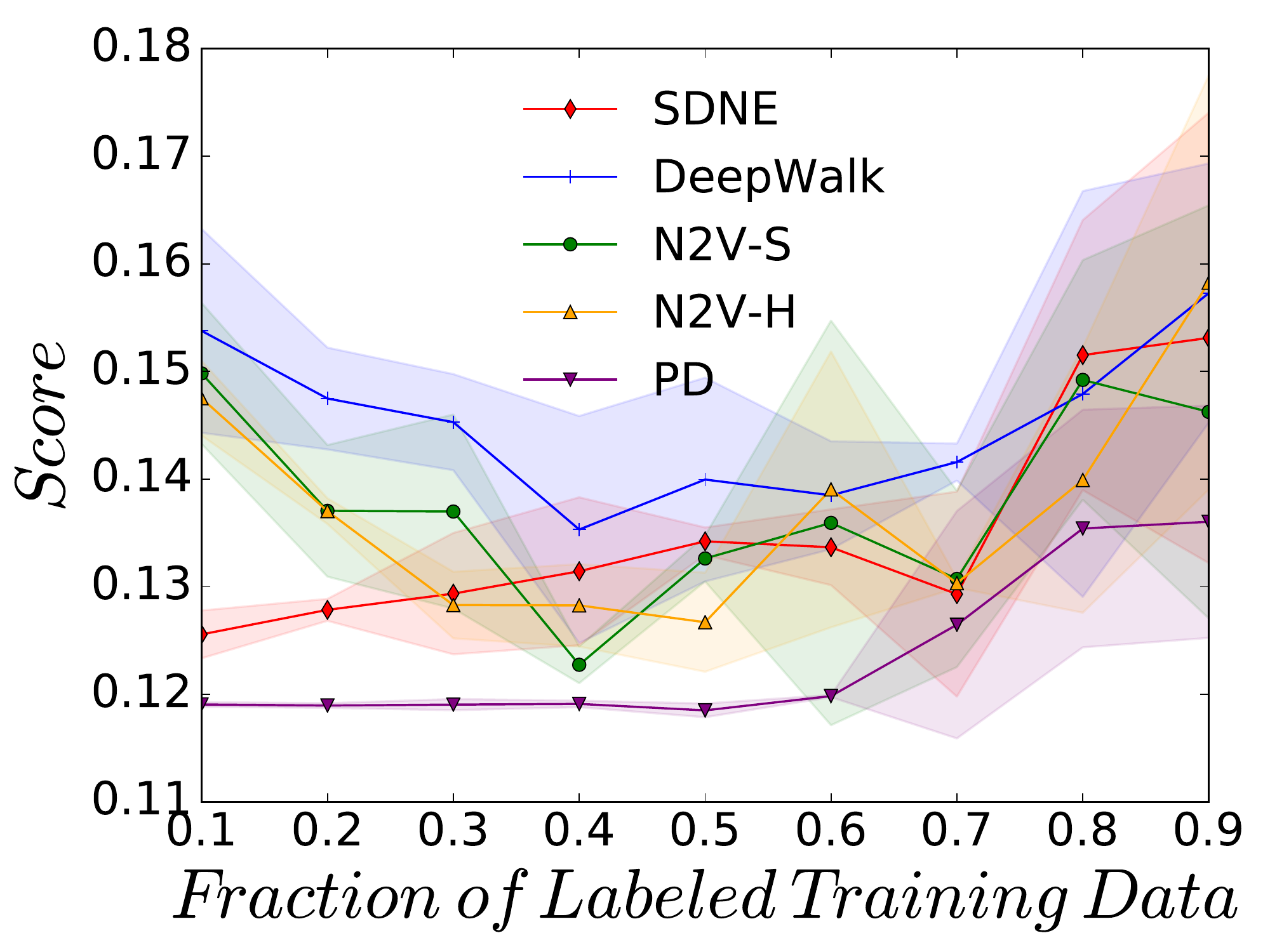}}
  \label{DCmaGN}\hfill
\subfloat[Micro Bitcoinotc]{%
    \includegraphics[width=0.25\linewidth]{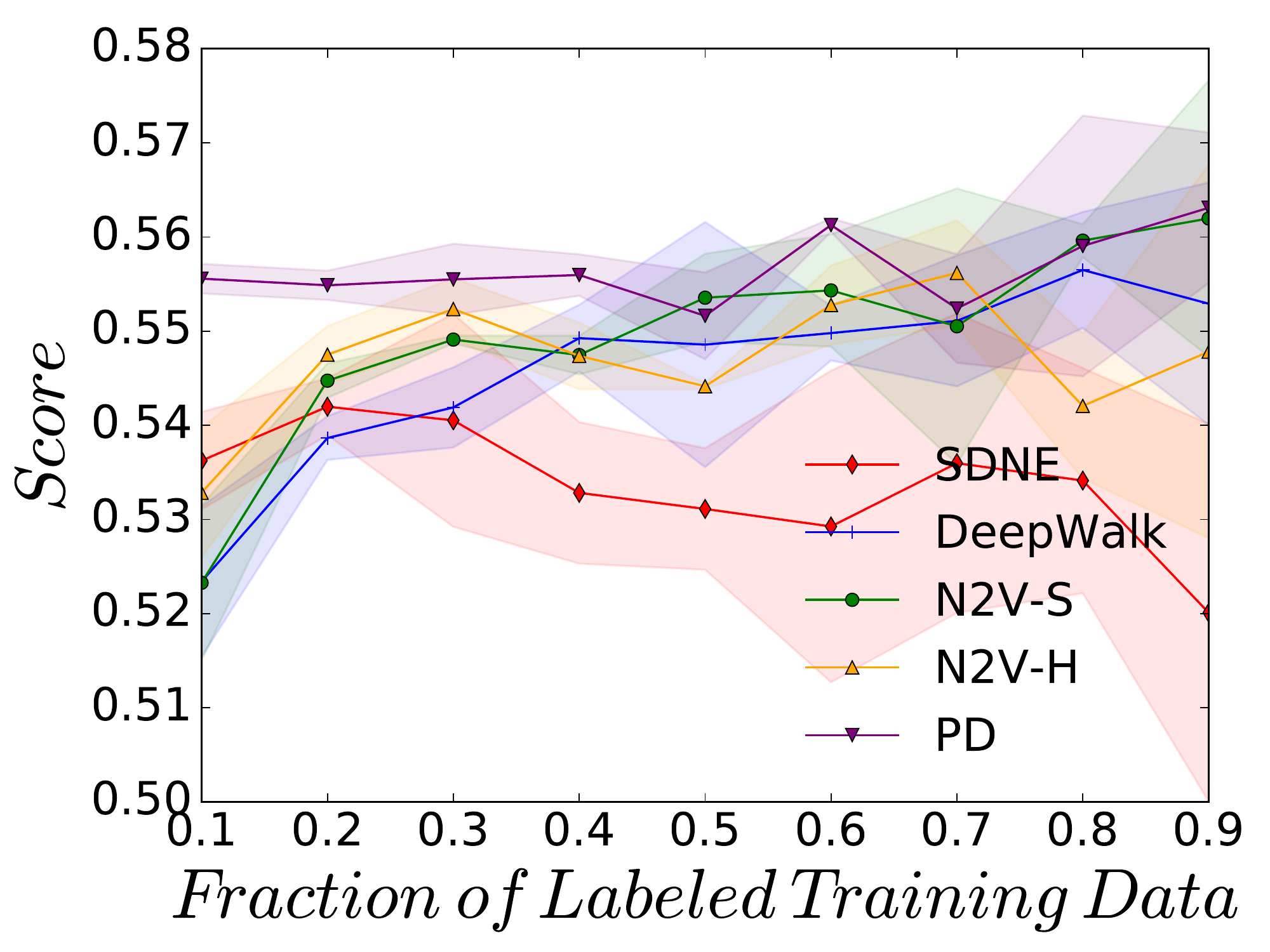}}
  \label{DCmaWI}
\caption{Micro-f1 and Macro-f1 Scores, across a range of labelling fractions, for all approaches when predicting a vertex's Local Clustering Coefficient value across all datasets.}
\label{fig:CLU_FIG}
\end{figure*}

\begin{figure*}
  \centering
\subfloat[Macro Drosophila]{%
    \includegraphics[width=0.25\linewidth]{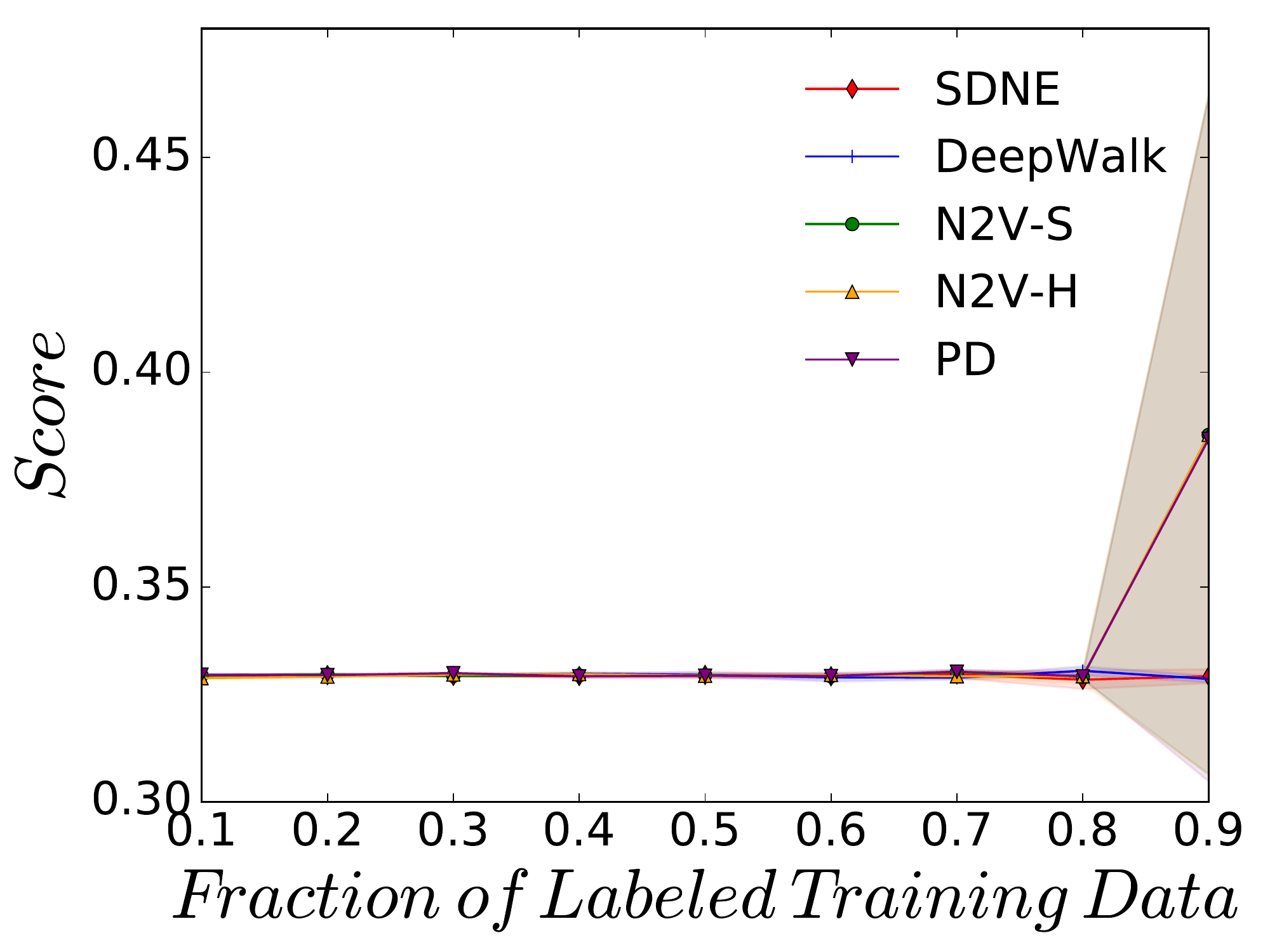}}
  \label{DCmiCA}\hfill
\subfloat[Micro Drosophila]{%
    \includegraphics[width=0.25\linewidth]{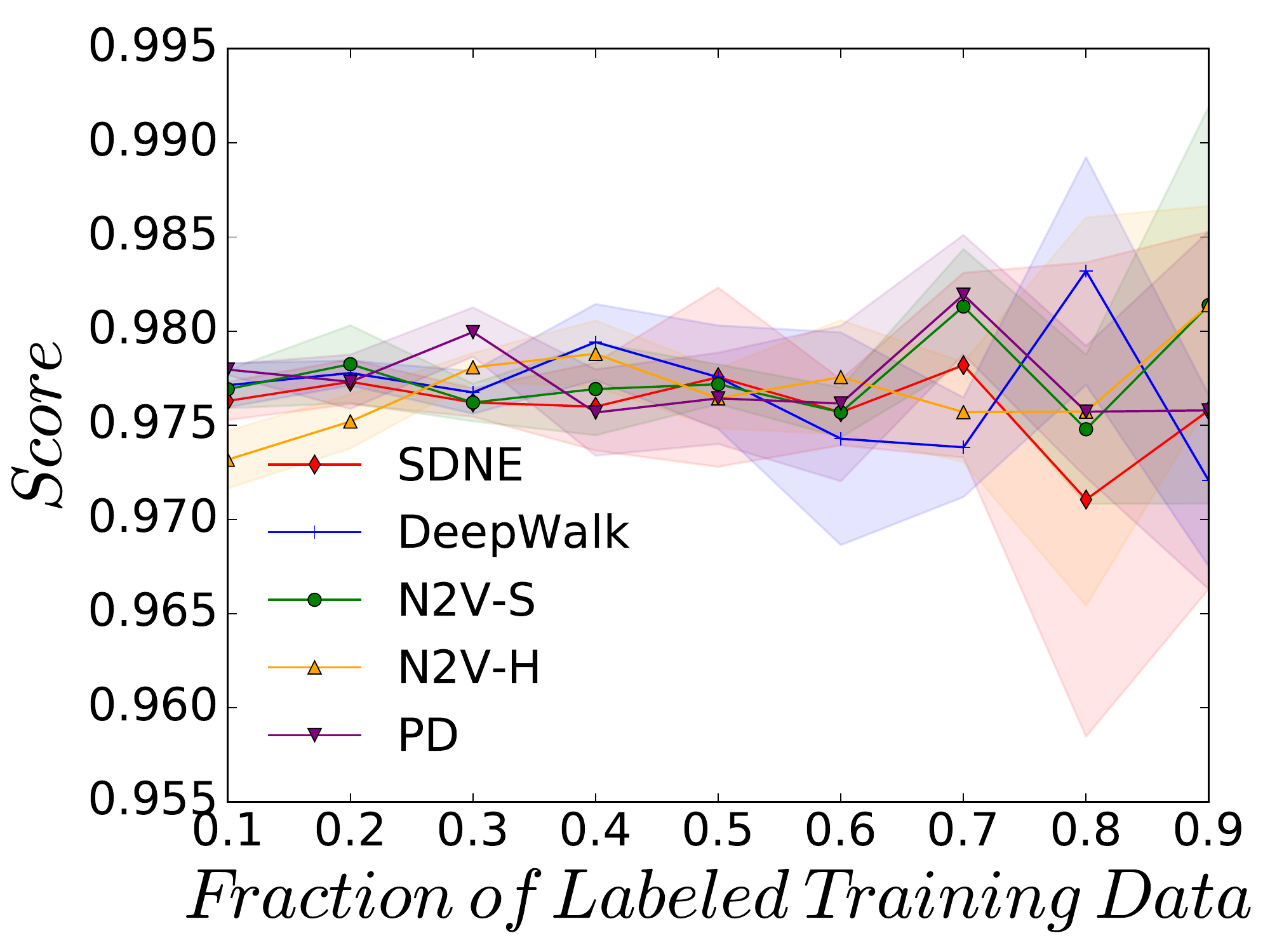}}
  \label{DCmiFB}\hfill
\subfloat[Macro HepTh]{%
    \includegraphics[width=0.25\linewidth]{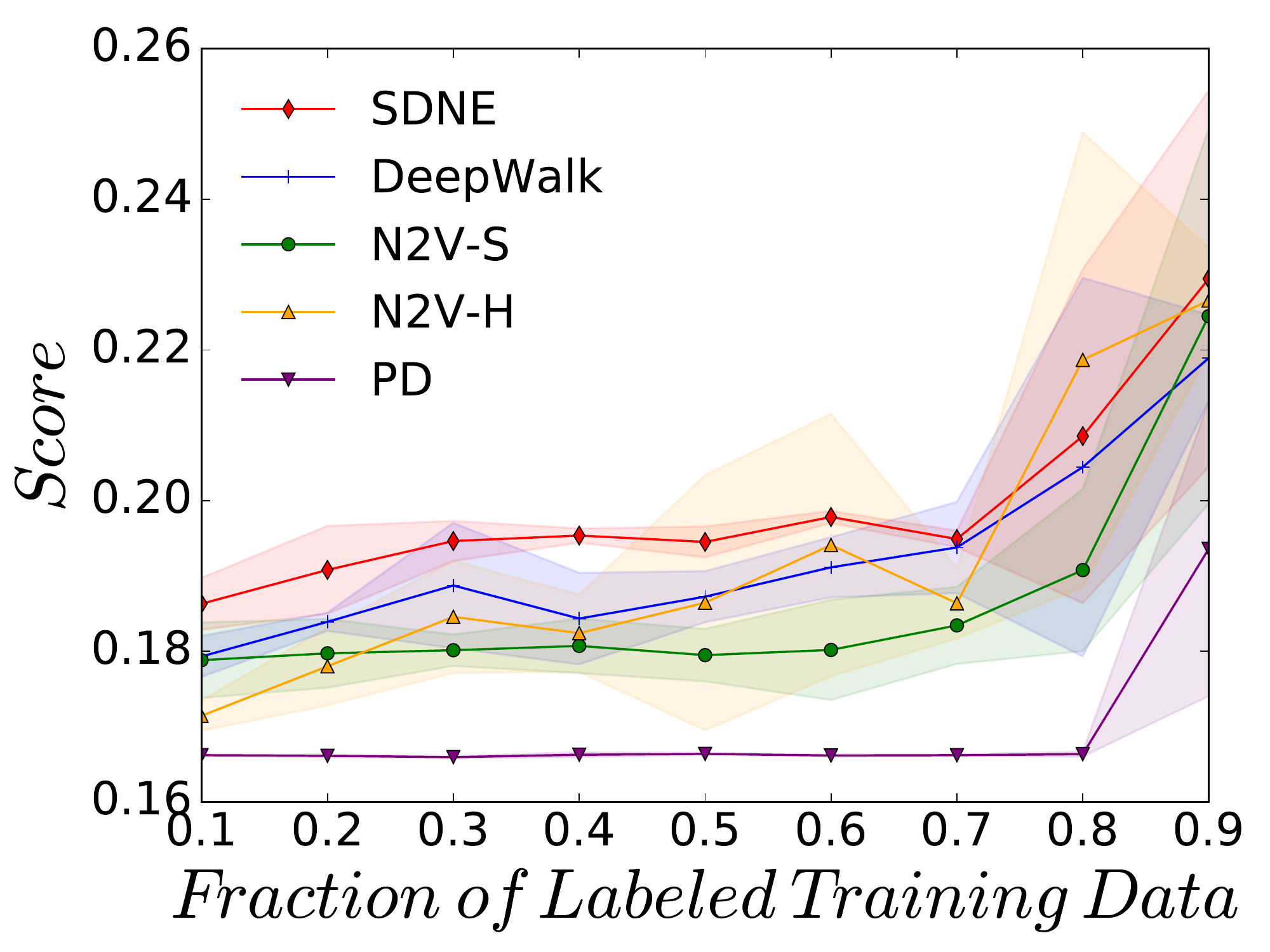}}
  \label{DCmiGN}\hfill
\subfloat[Micro HepTh]{%
    \includegraphics[width=0.25\linewidth]{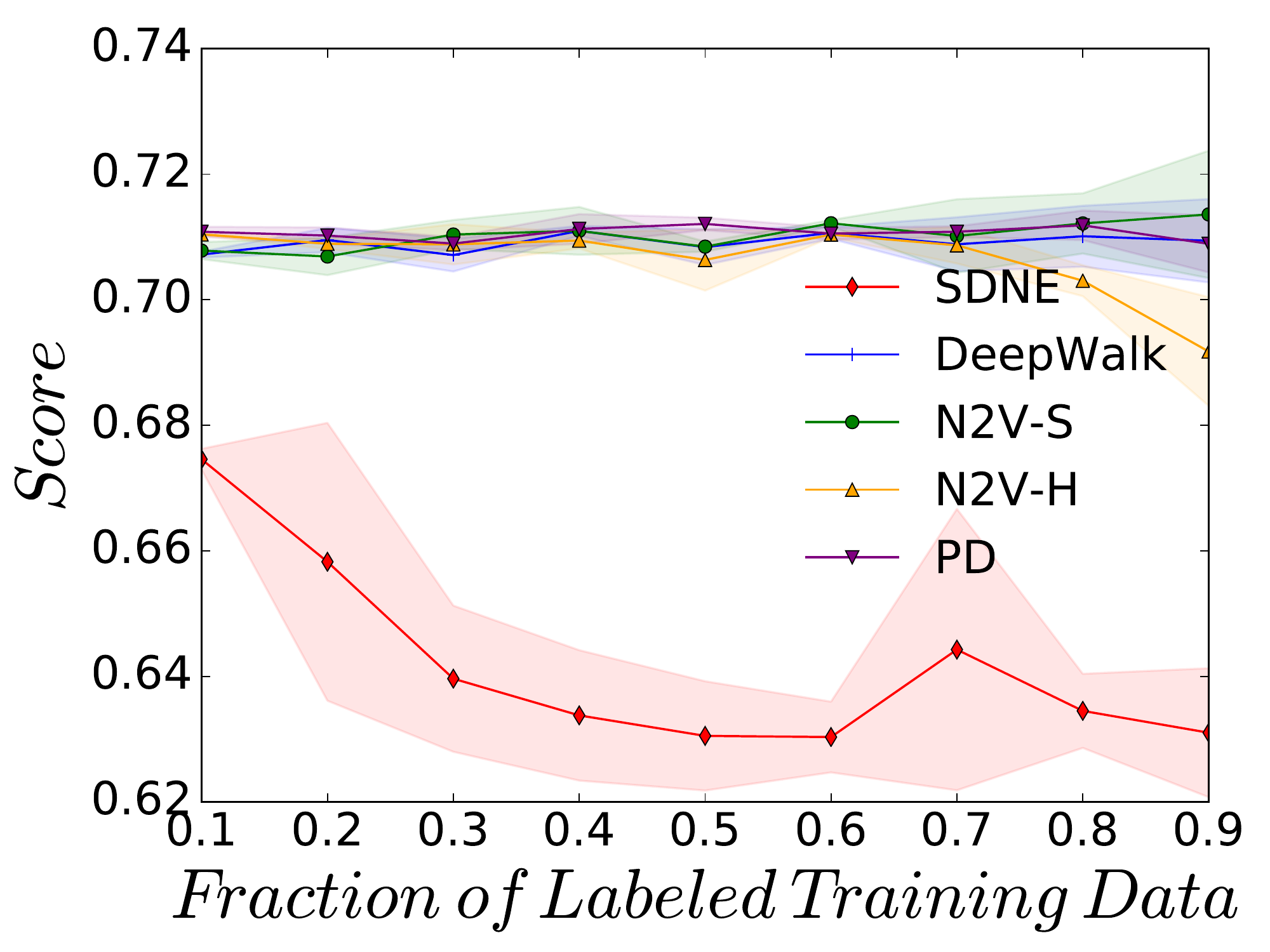}}
  \label{DCmiWI}\\

\subfloat[Macro Email-EU]{%
    \includegraphics[width=0.25\linewidth]{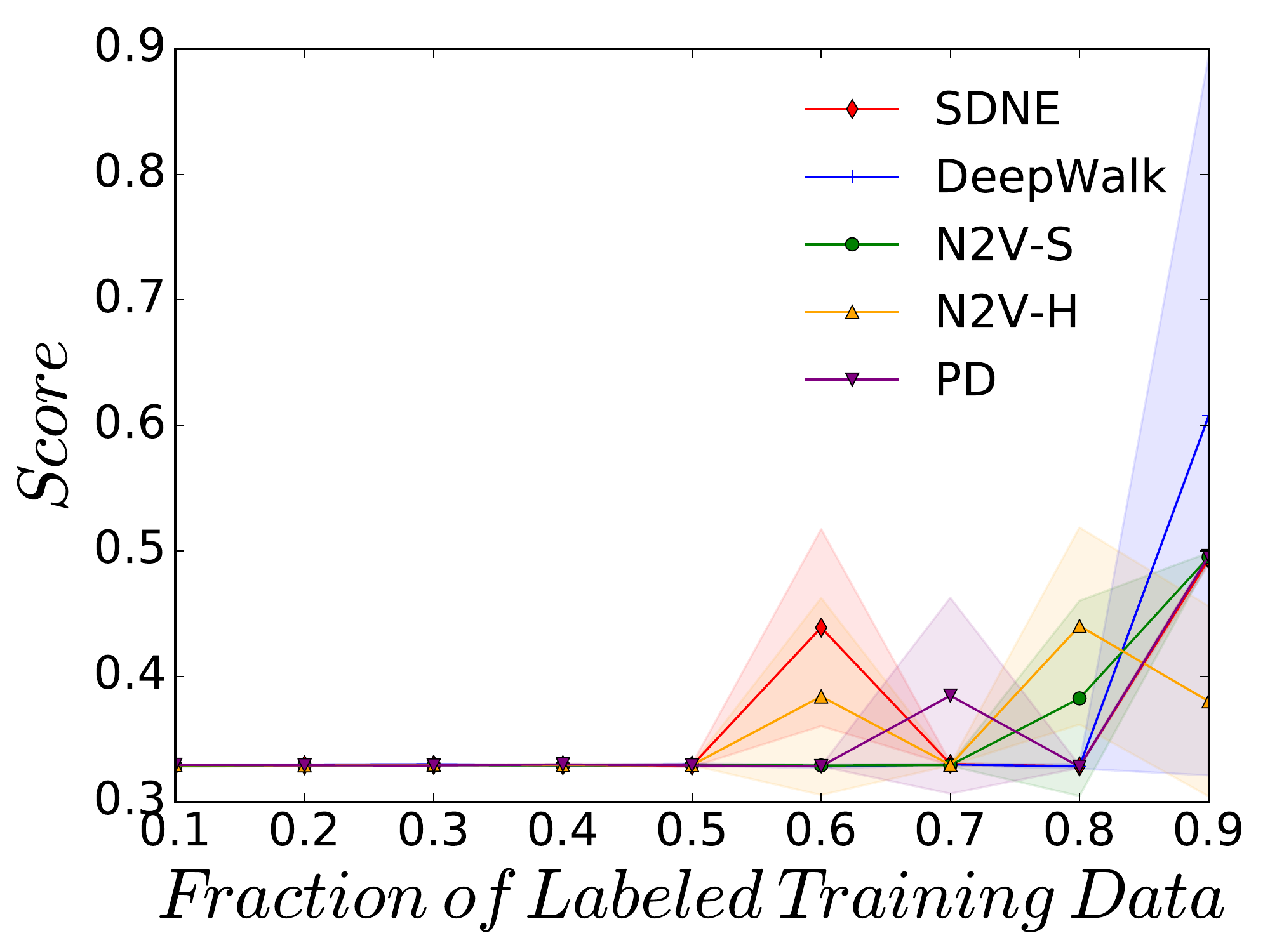}}
  \label{DCmaCA}\hfill
\subfloat[Micro Email-EU]{%
    \includegraphics[width=0.25\linewidth]{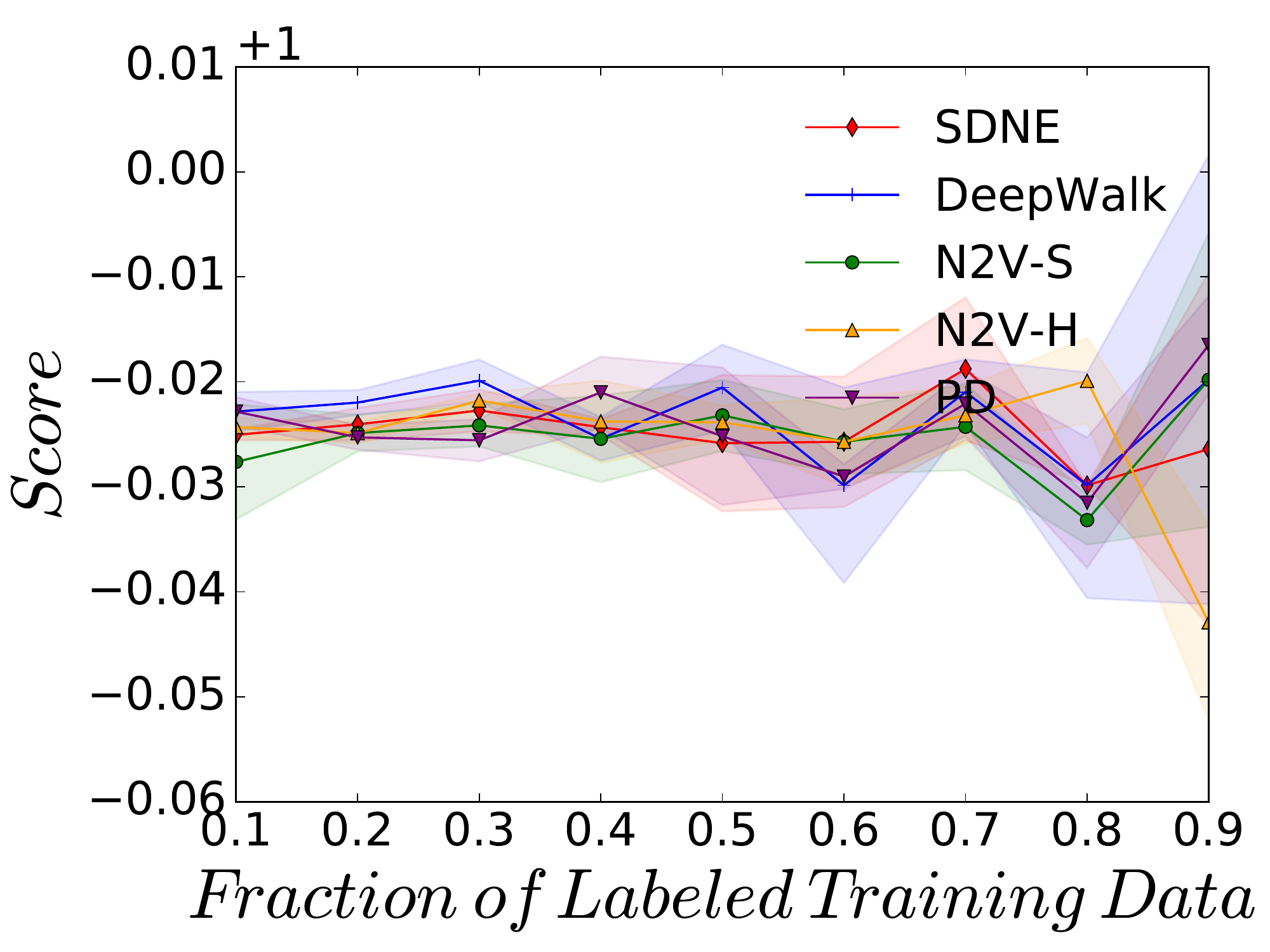}}
  \label{DCmaFB}\hfill
\subfloat[Macro Facebook]{%
    \includegraphics[width=0.25\linewidth]{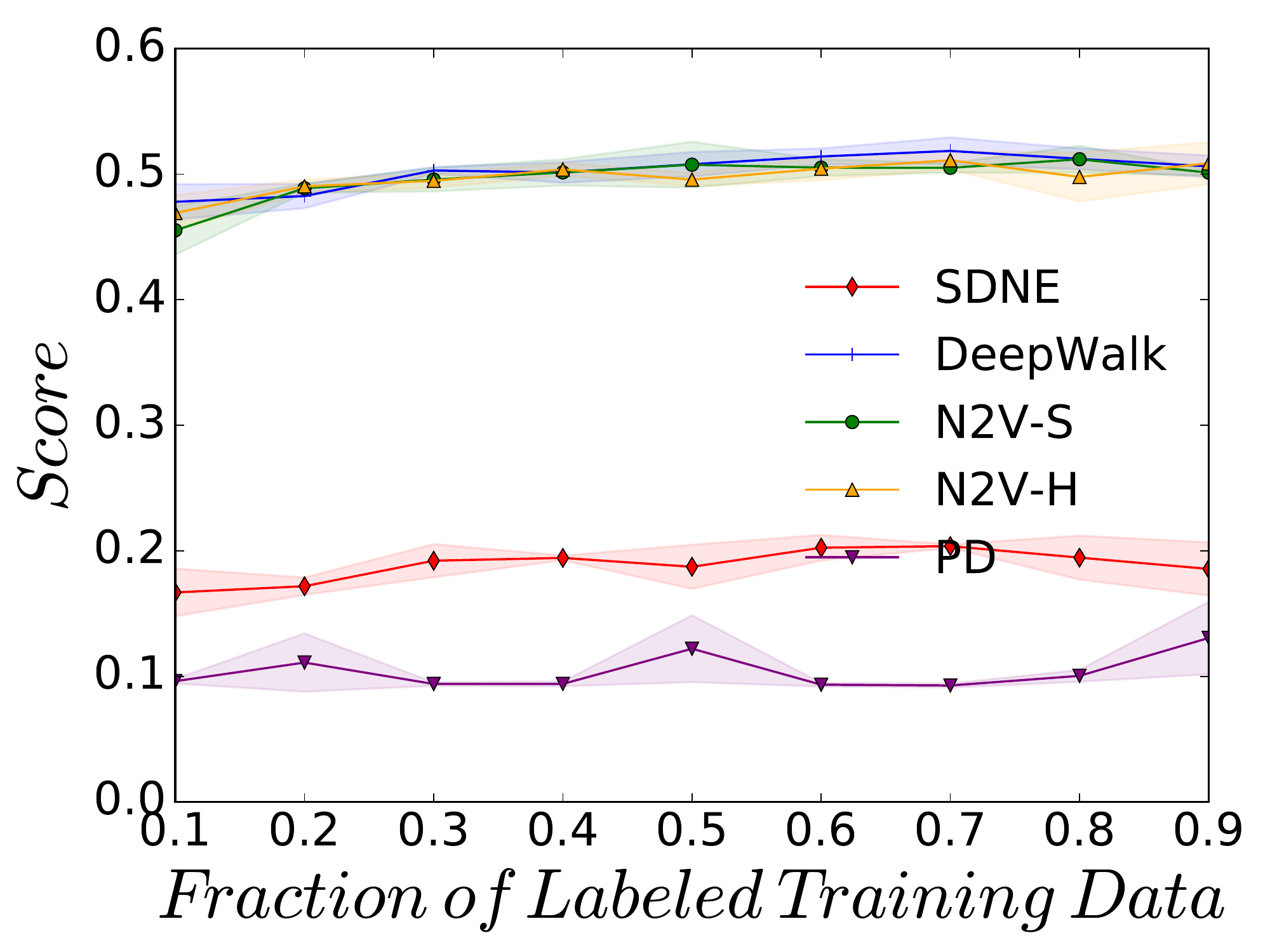}}
  \label{DCmaGN}\hfill
\subfloat[Micro Facebook]{%
    \includegraphics[width=0.25\linewidth]{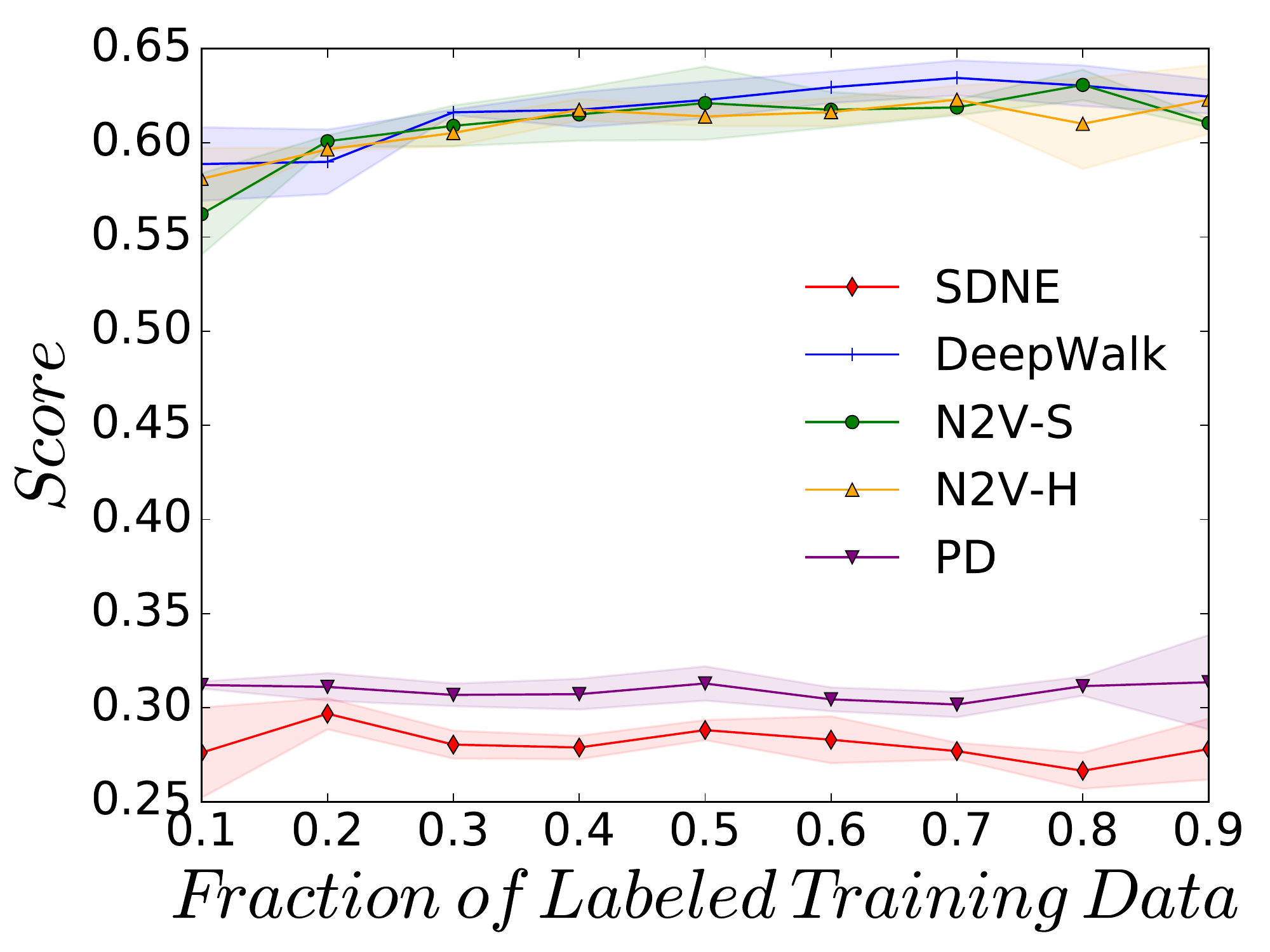}}
  \label{DCmaWI} \\ 

  \subfloat[Macro Openflights]{%
    \includegraphics[width=0.25\linewidth]{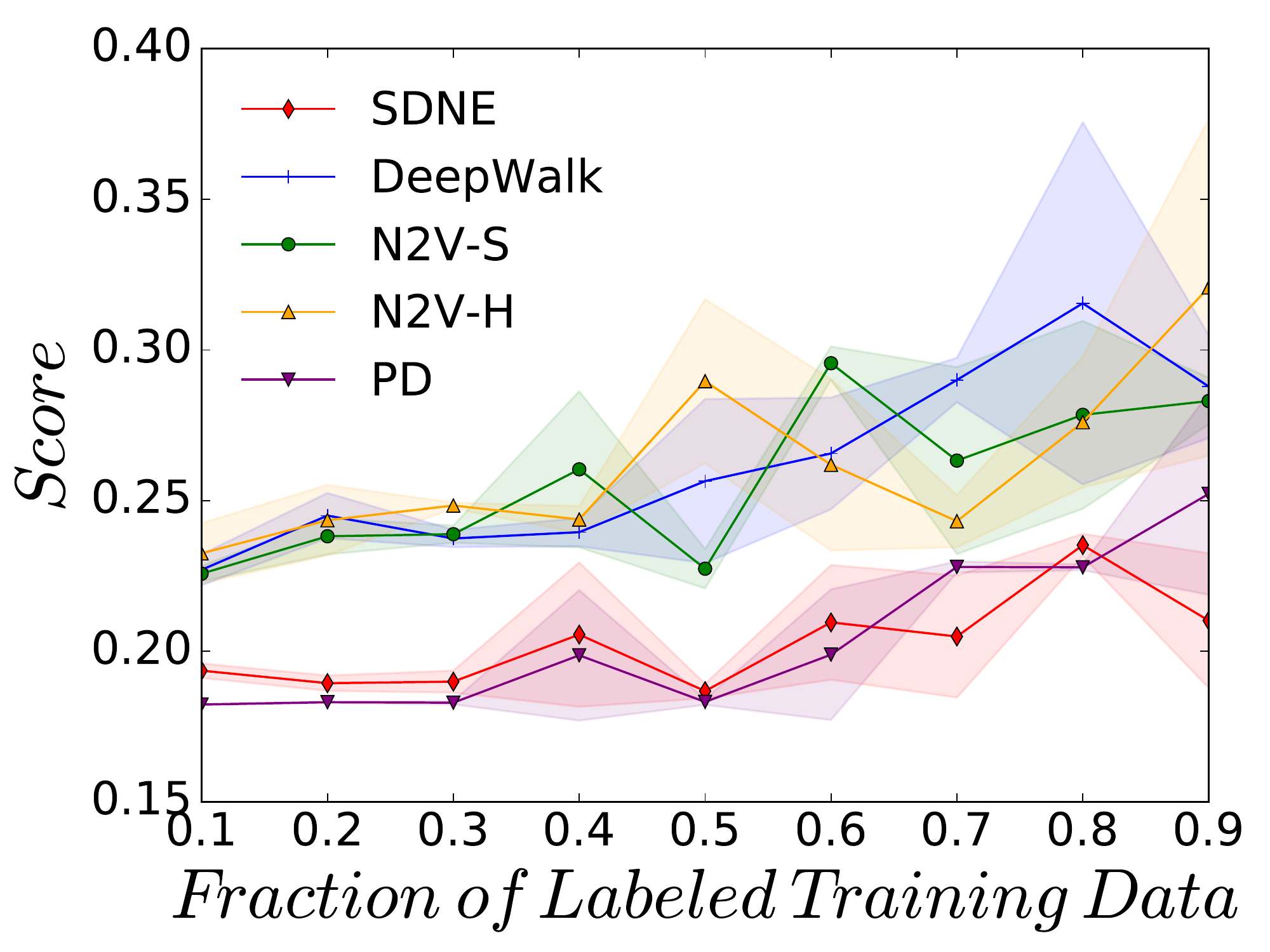}}
  \label{DCmaCA}\hfill
\subfloat[Micro Openflights]{%
    \includegraphics[width=0.25\linewidth]{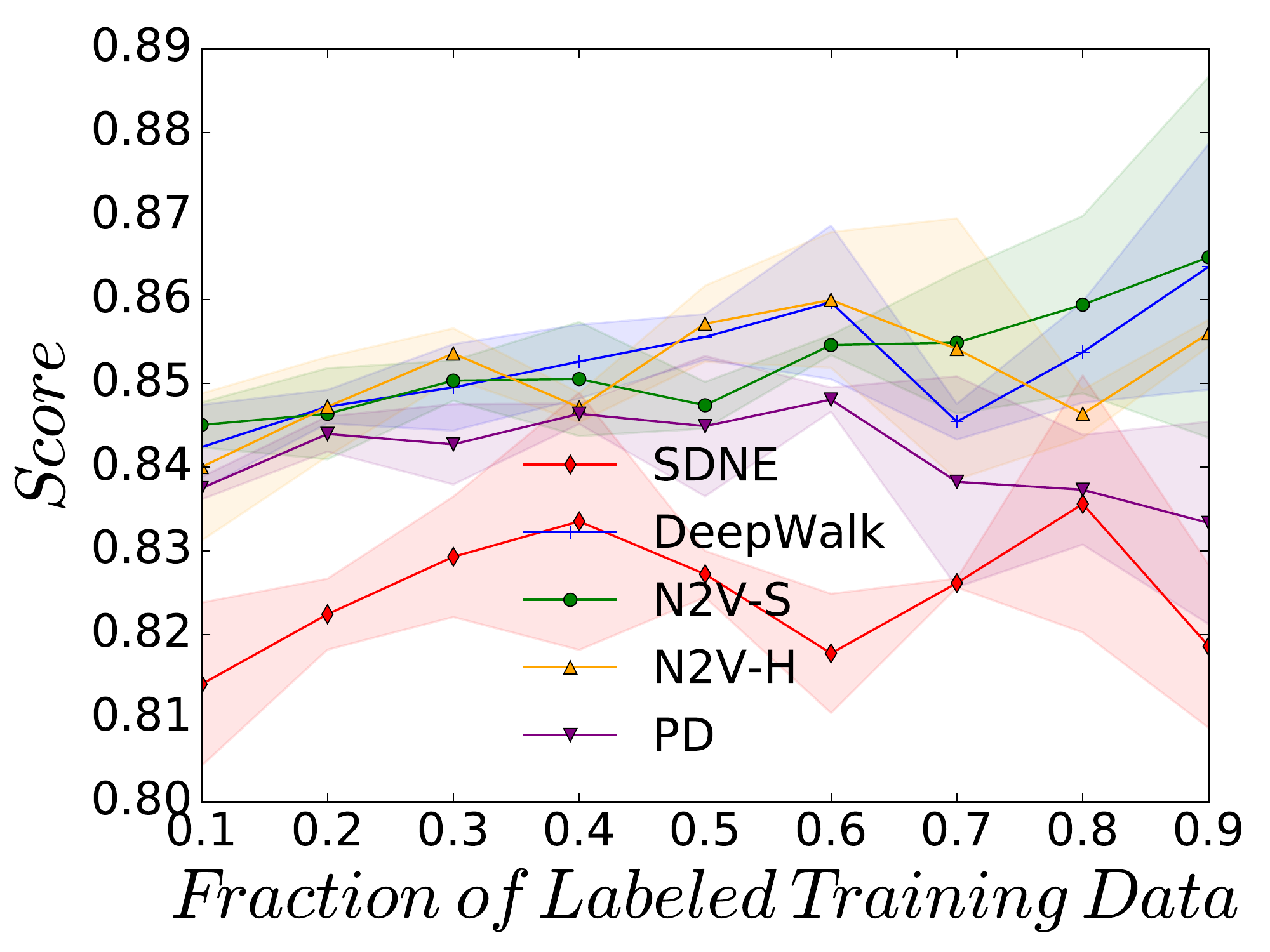}}
  \label{DCmaFB}\hfill
\subfloat[Macro Bitcoinotc]{%
    \includegraphics[width=0.25\linewidth]{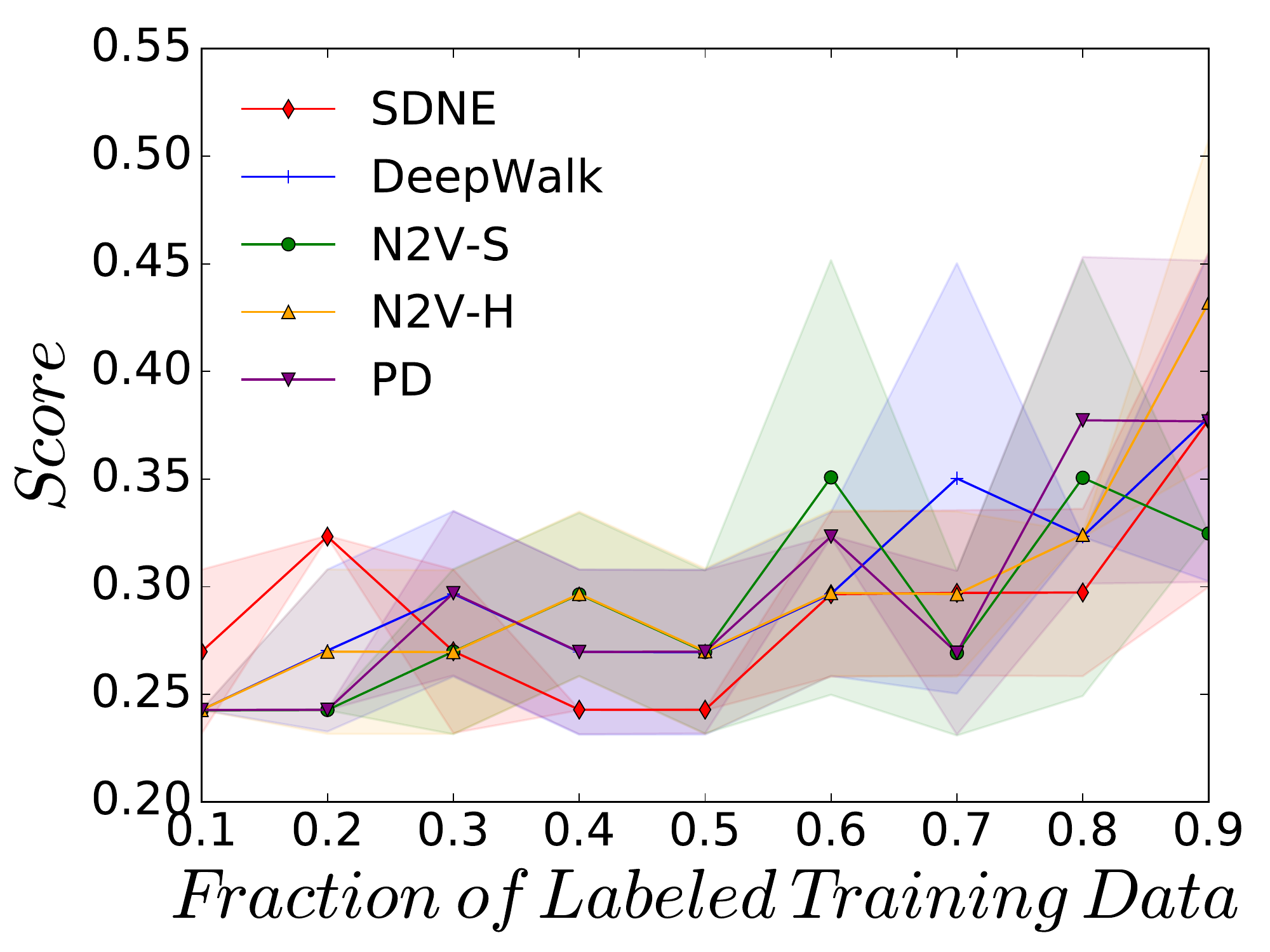}}
  \label{DCmaGN}\hfill
\subfloat[Micro Bitcoinotc]{%
    \includegraphics[width=0.25\linewidth]{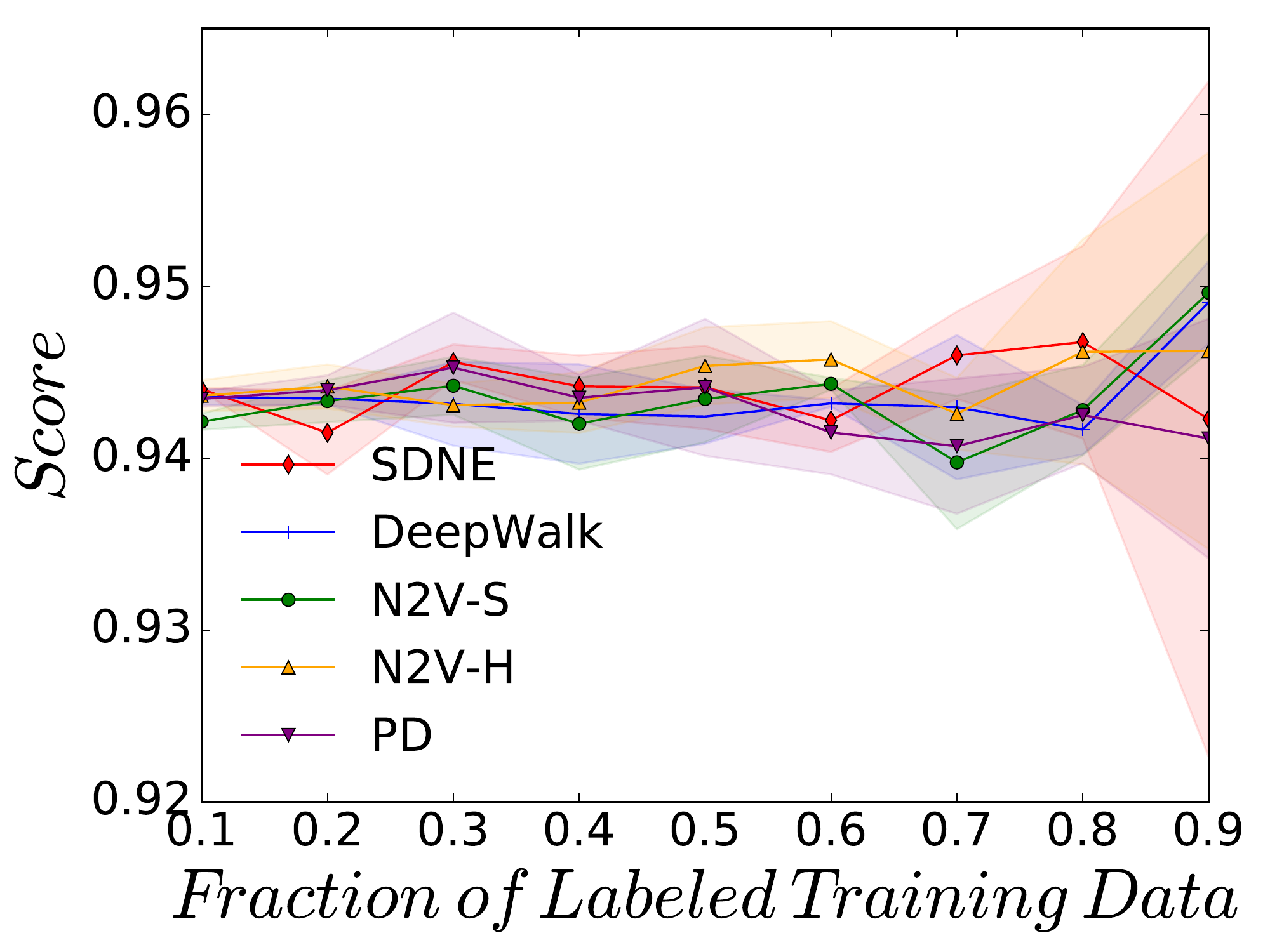}}
  \label{DCmaWI}
\caption{Micro-f1 and Macro-f1 Scores, across a range of labelling fractions, for all approaches when predicting a vertex's Eigenvetor Centrality value across all datasets.}
\label{fig:EC_FIG}
\end{figure*}

\begin{figure*}
  \centering
\subfloat[Macro Drosophila]{%
    \includegraphics[width=0.25\linewidth]{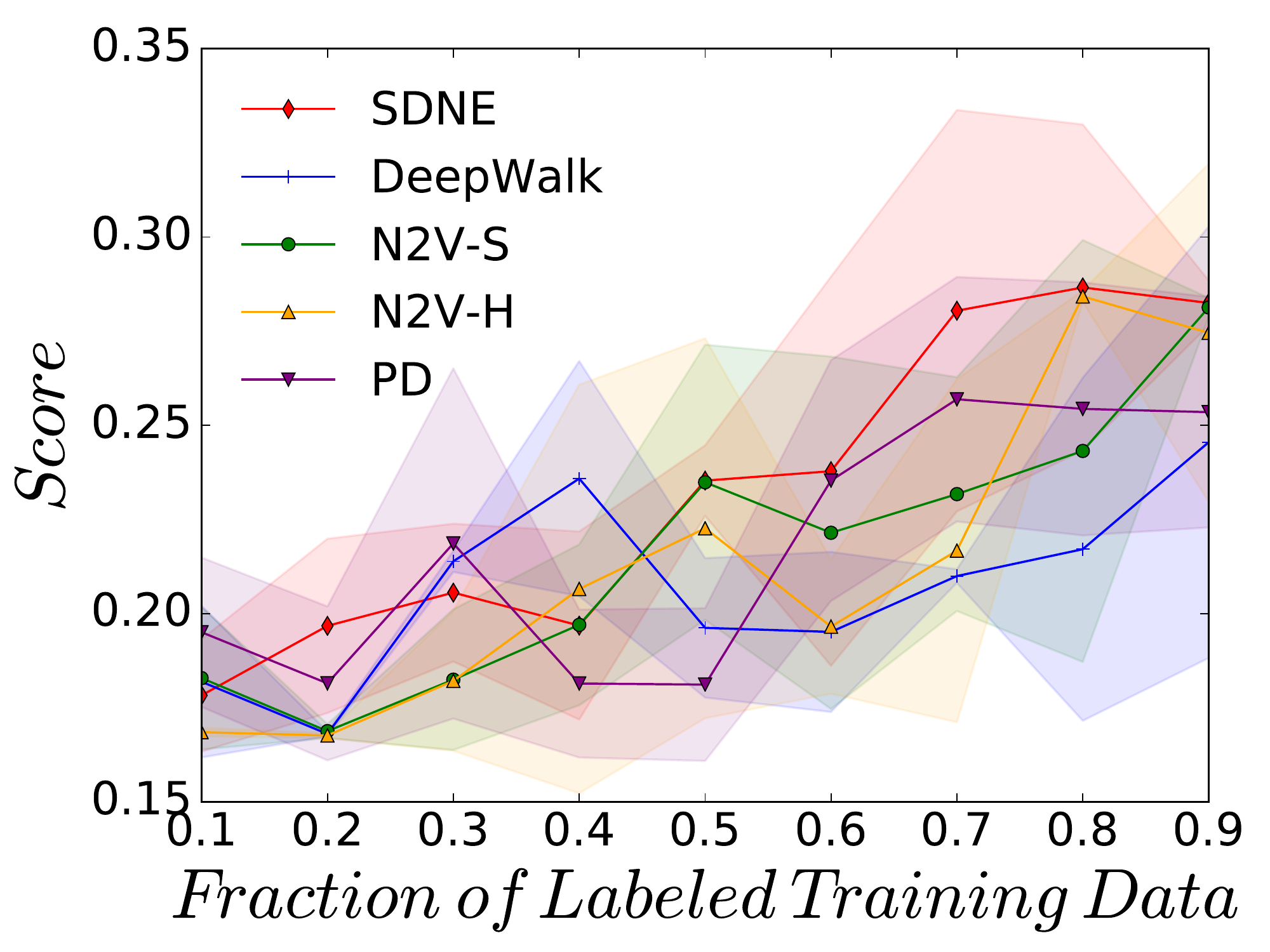}}
  \label{DCmiCA}\hfill
\subfloat[Micro Drosophila]{%
    \includegraphics[width=0.25\linewidth]{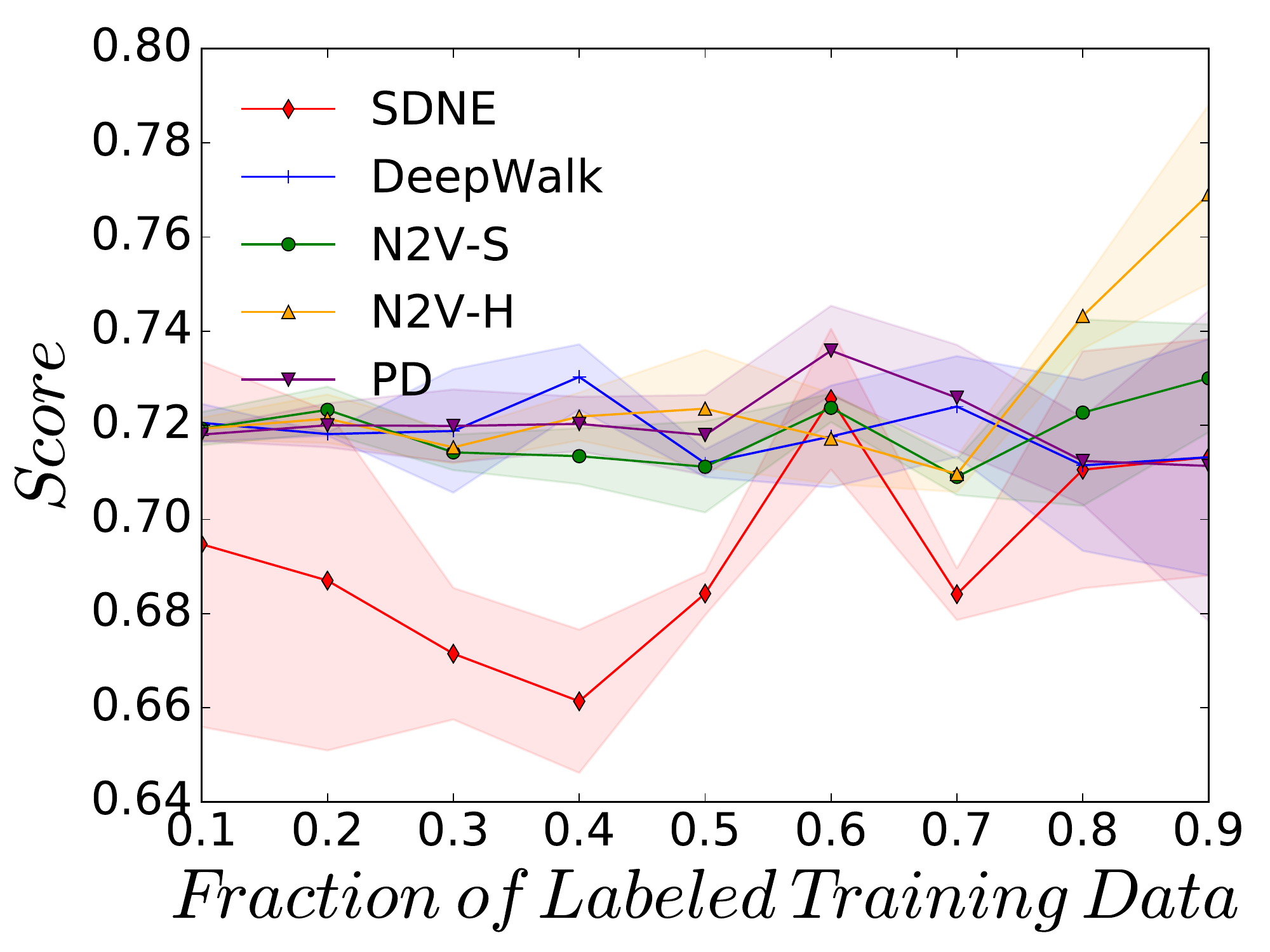}}
  \label{DCmiFB}\hfill
\subfloat[Macro HepTh]{%
    \includegraphics[width=0.25\linewidth]{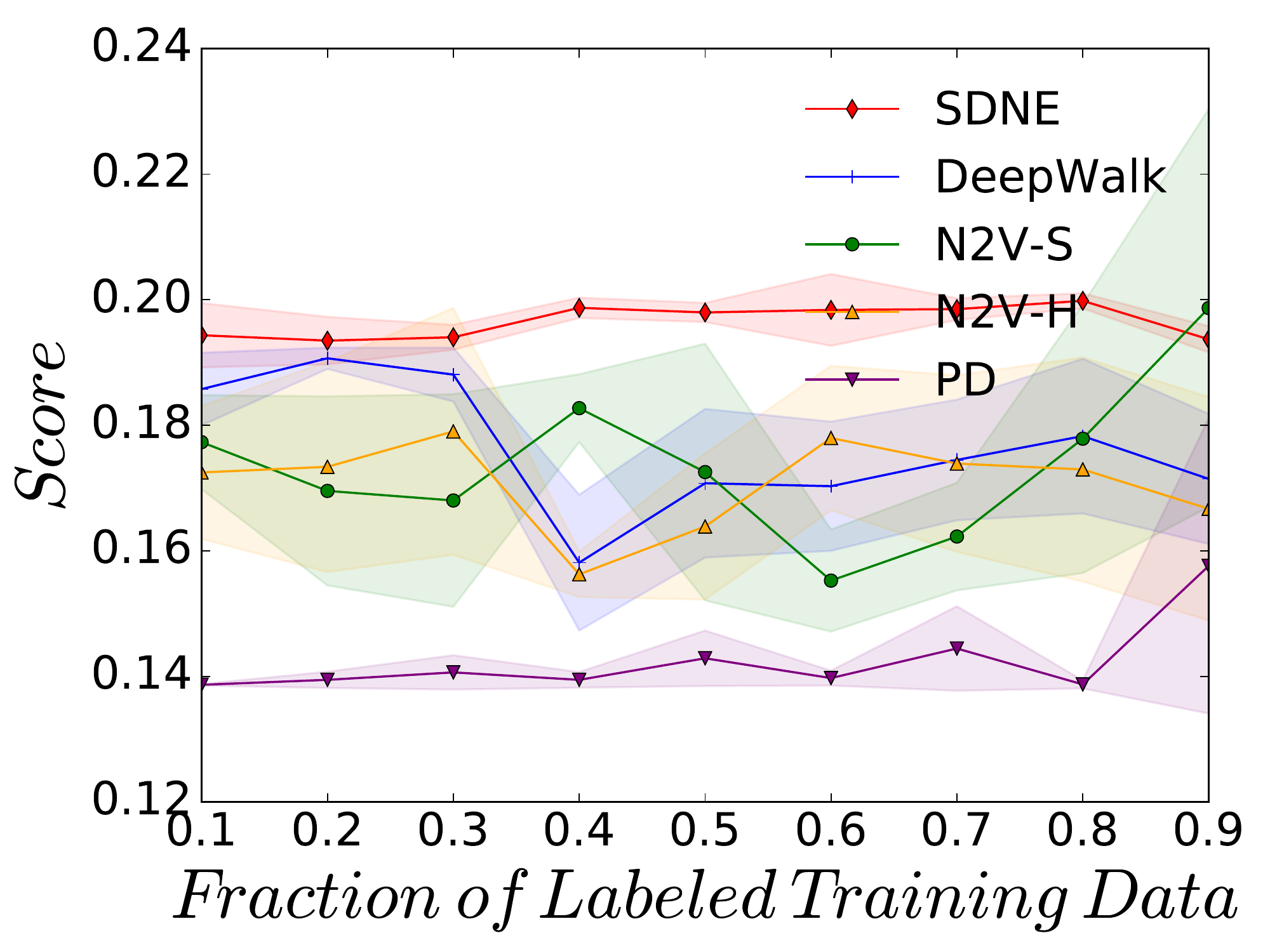}}
  \label{DCmiGN}\hfill
\subfloat[Micro HepTh]{%
    \includegraphics[width=0.25\linewidth]{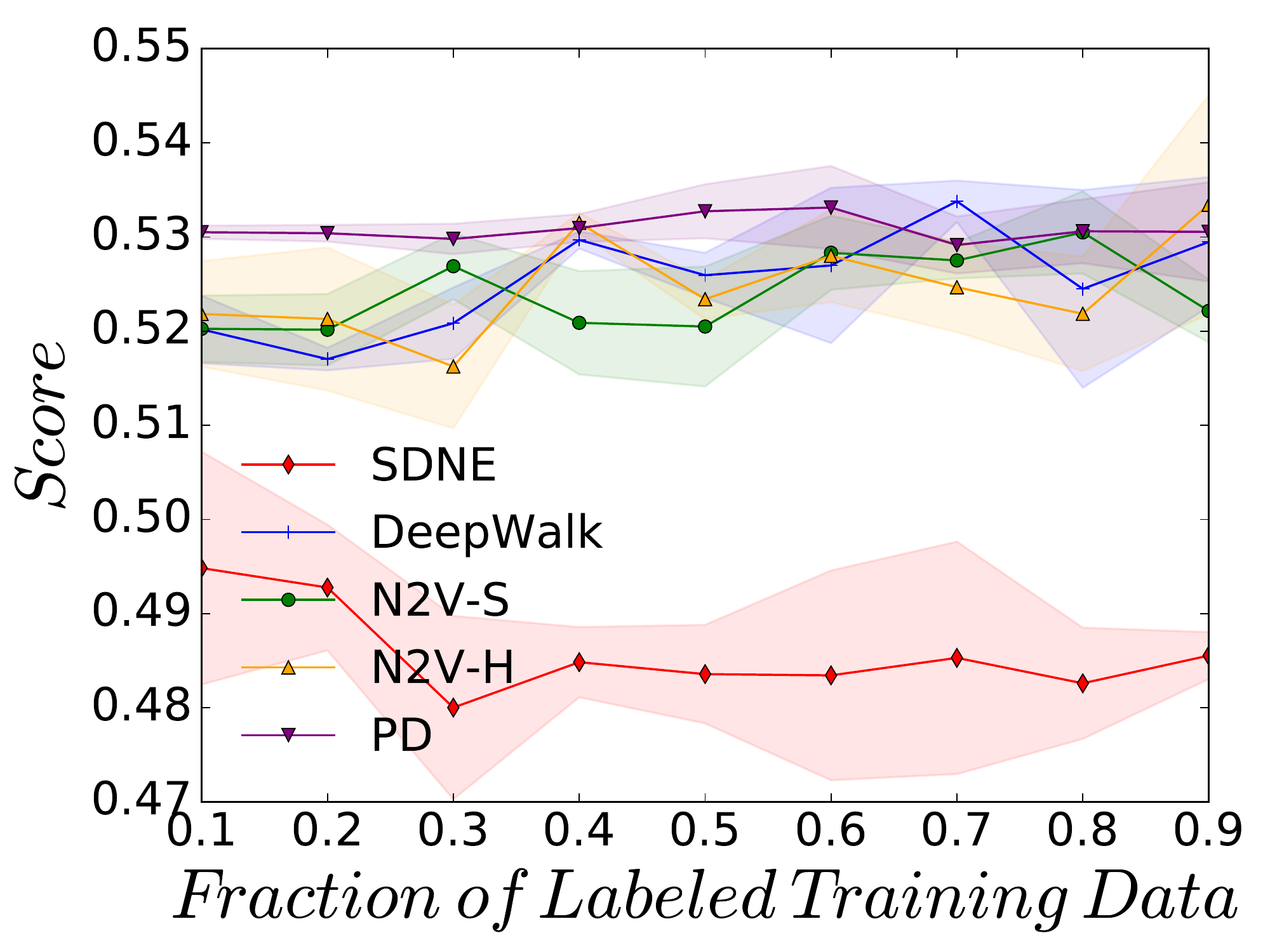}}
  \label{DCmiWI}\\

\subfloat[Macro Email-EU]{%
    \includegraphics[width=0.25\linewidth]{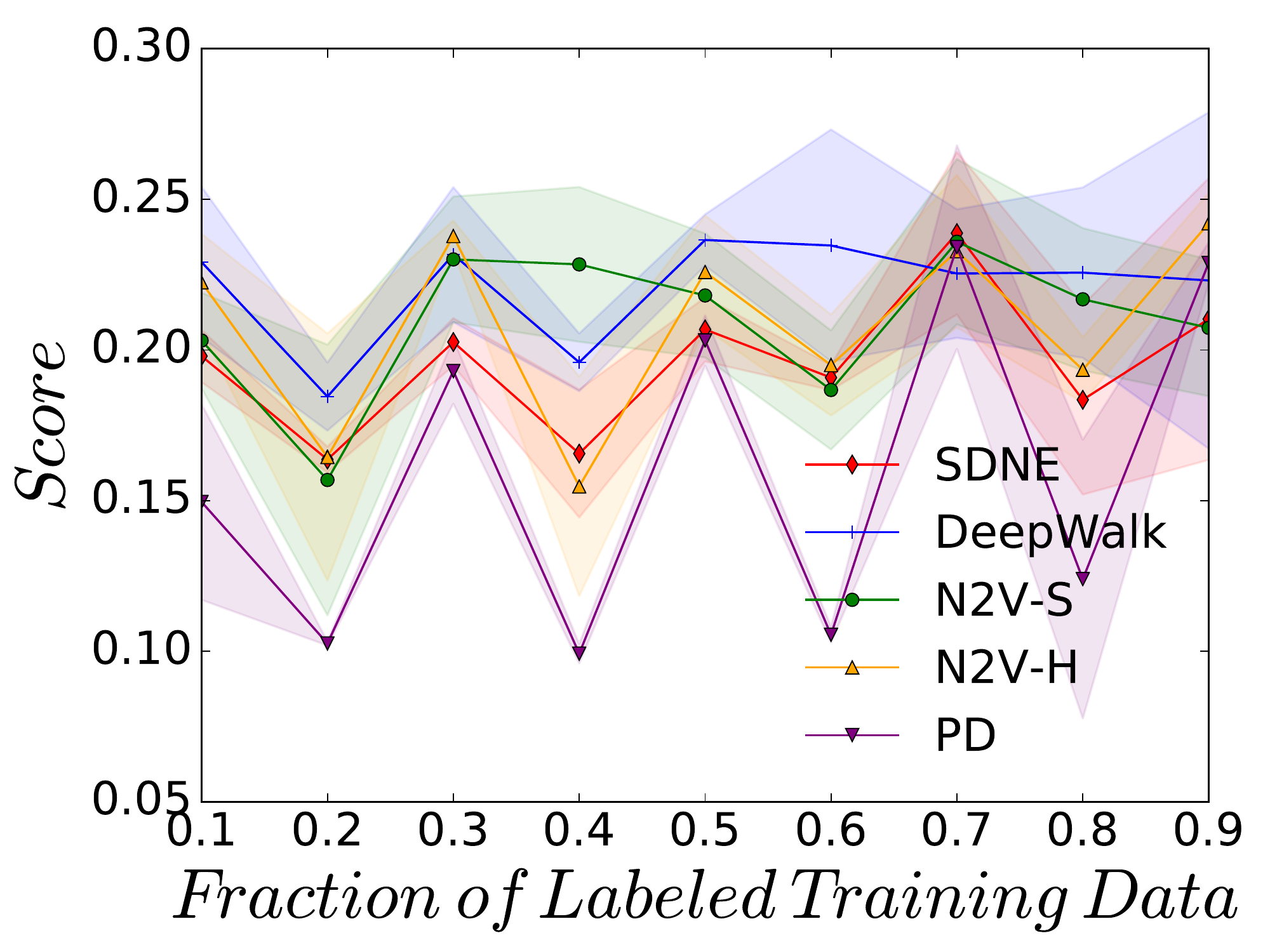}}
  \label{DCmaCA}\hfill
\subfloat[Micro Email-EU]{%
    \includegraphics[width=0.25\linewidth]{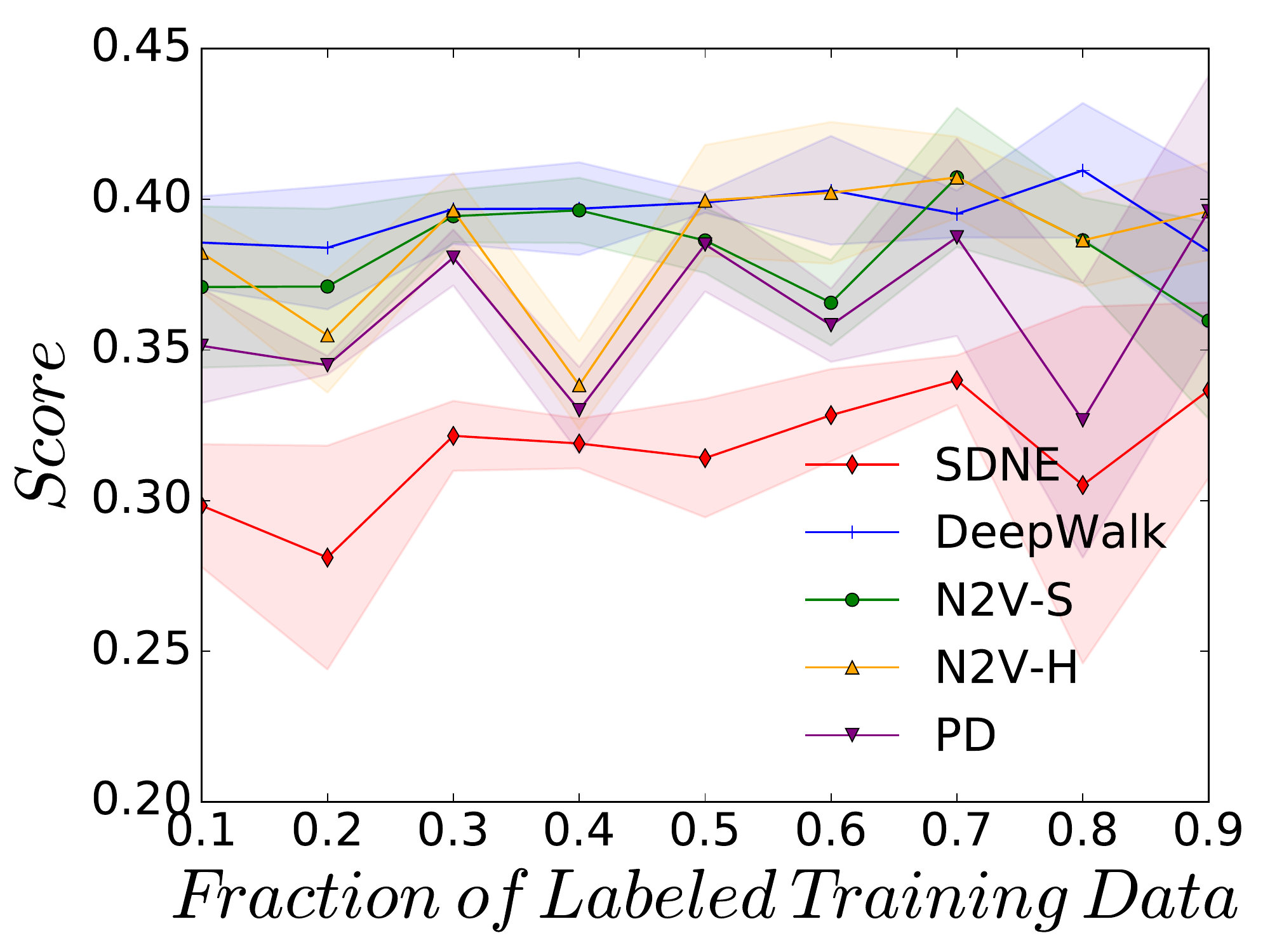}}
  \label{DCmaFB}\hfill
\subfloat[Macro Facebook]{%
    \includegraphics[width=0.25\linewidth]{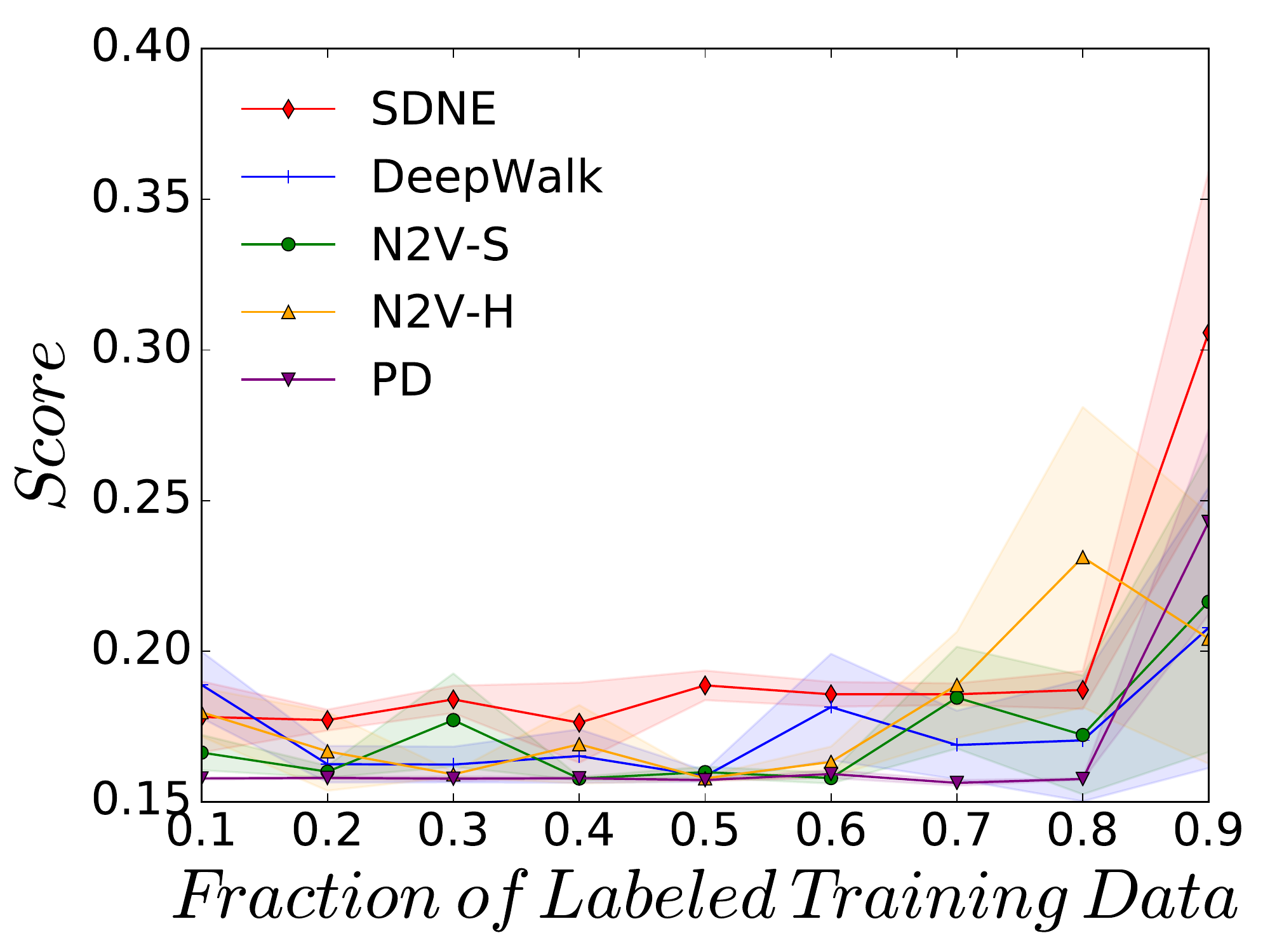}}
  \label{DCmaGN}\hfill
\subfloat[Micro Facebook]{%
    \includegraphics[width=0.25\linewidth]{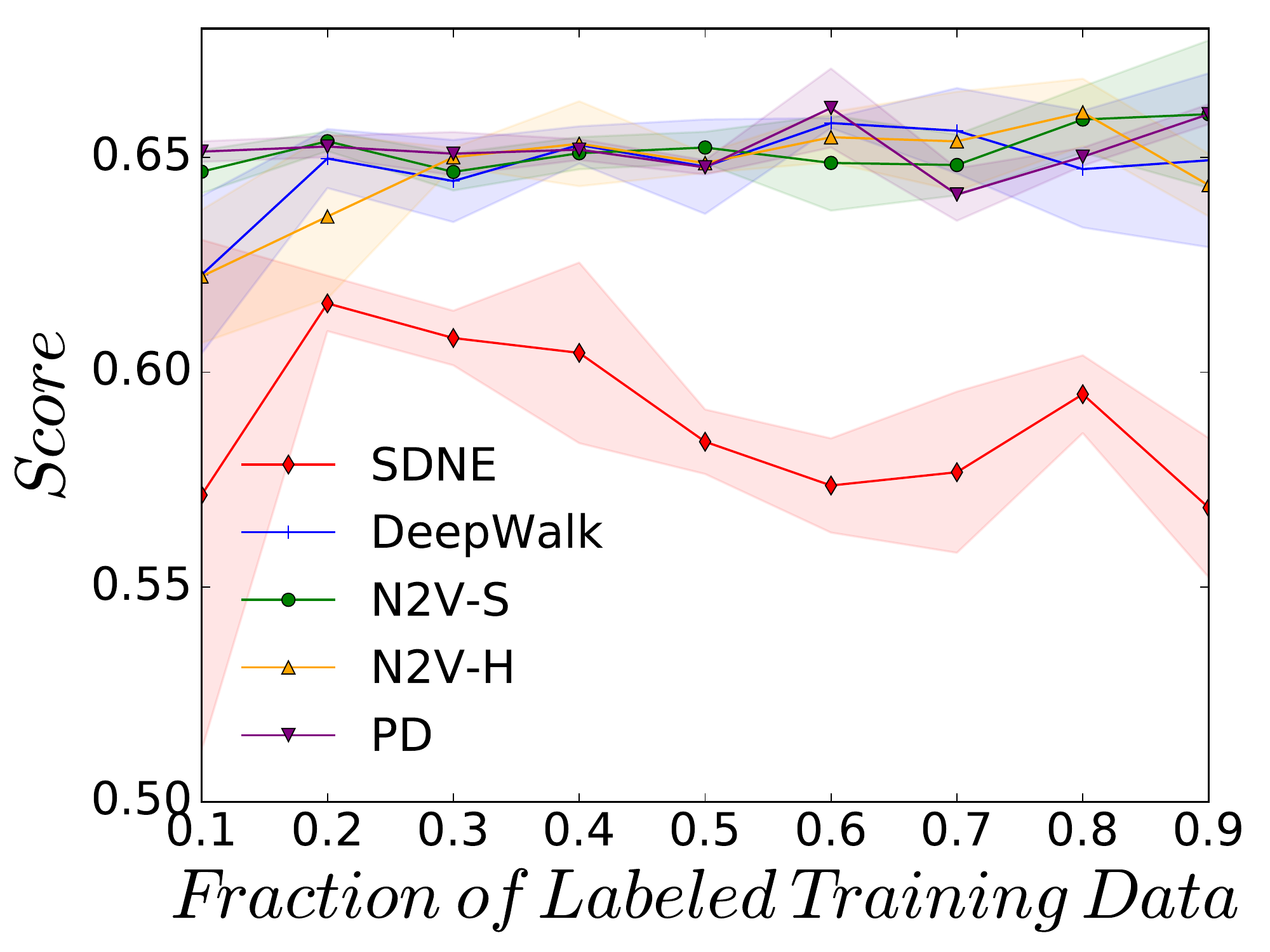}}
  \label{DCmaWI} \\ 

  \subfloat[Macro Openflights]{%
    \includegraphics[width=0.25\linewidth]{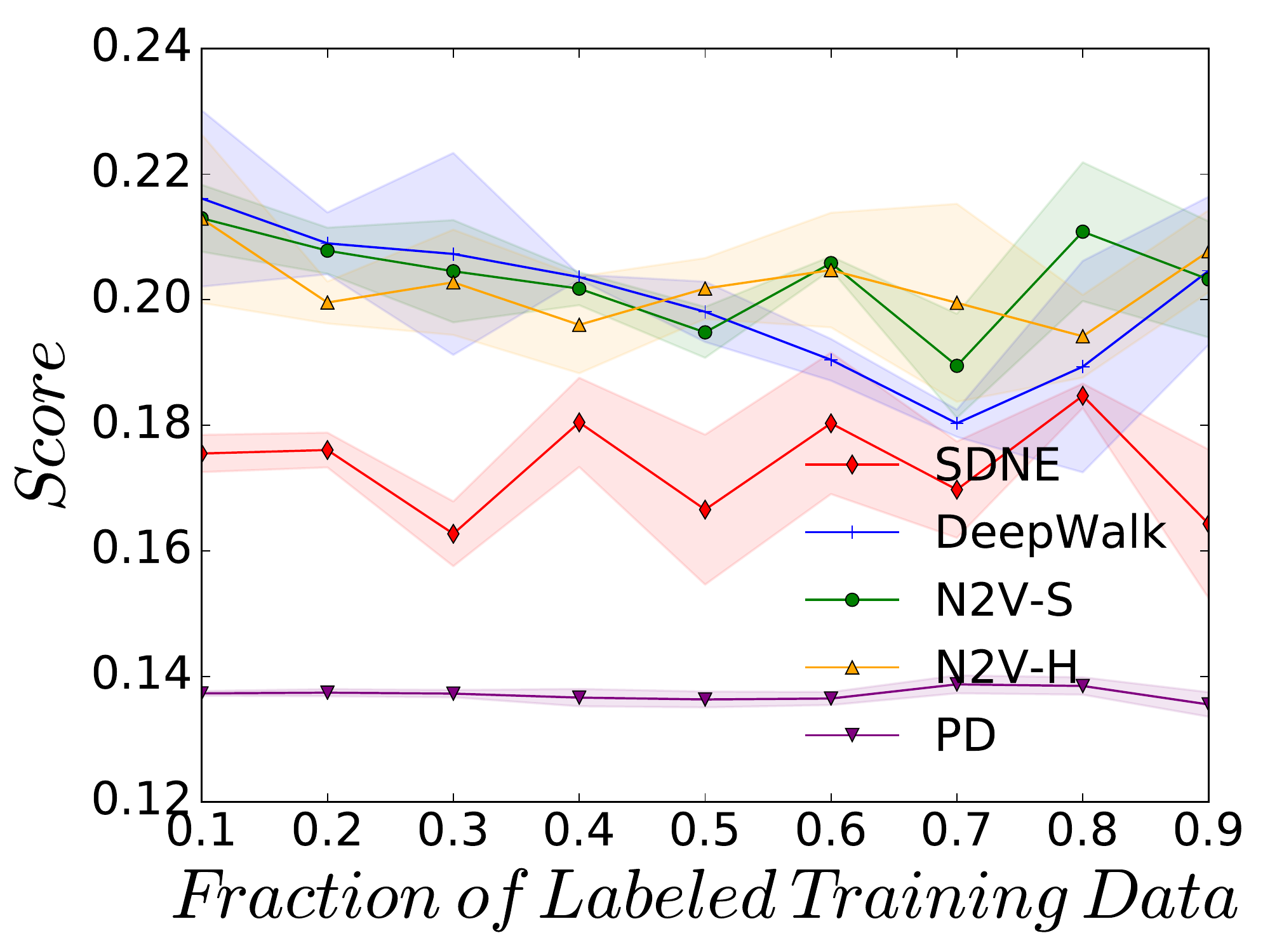}}
  \label{DCmaCA}\hfill
\subfloat[Micro Openflights]{%
    \includegraphics[width=0.25\linewidth]{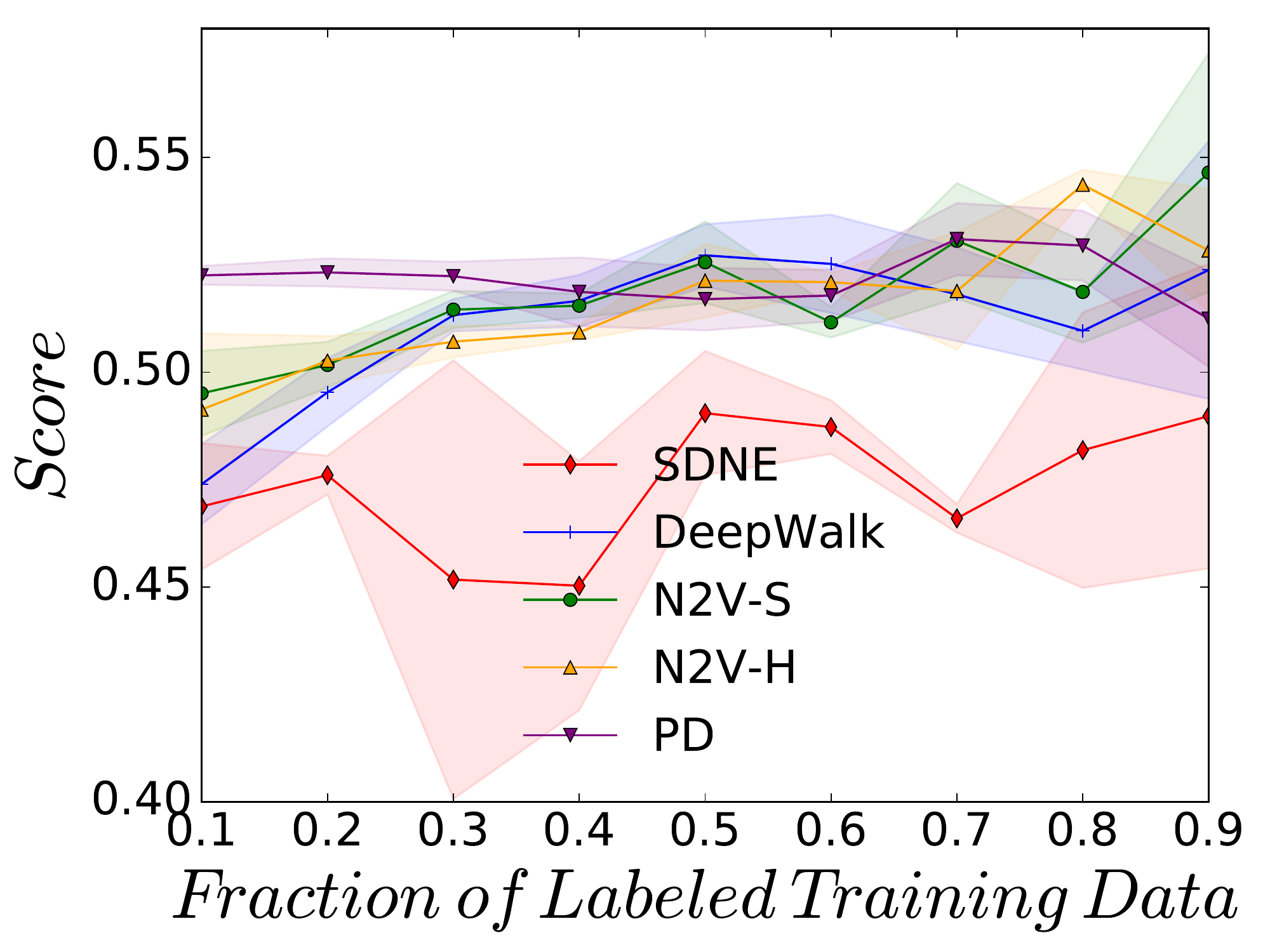}}
  \label{DCmaFB}\hfill
\subfloat[Macro Bitcoinotc]{%
    \includegraphics[width=0.25\linewidth]{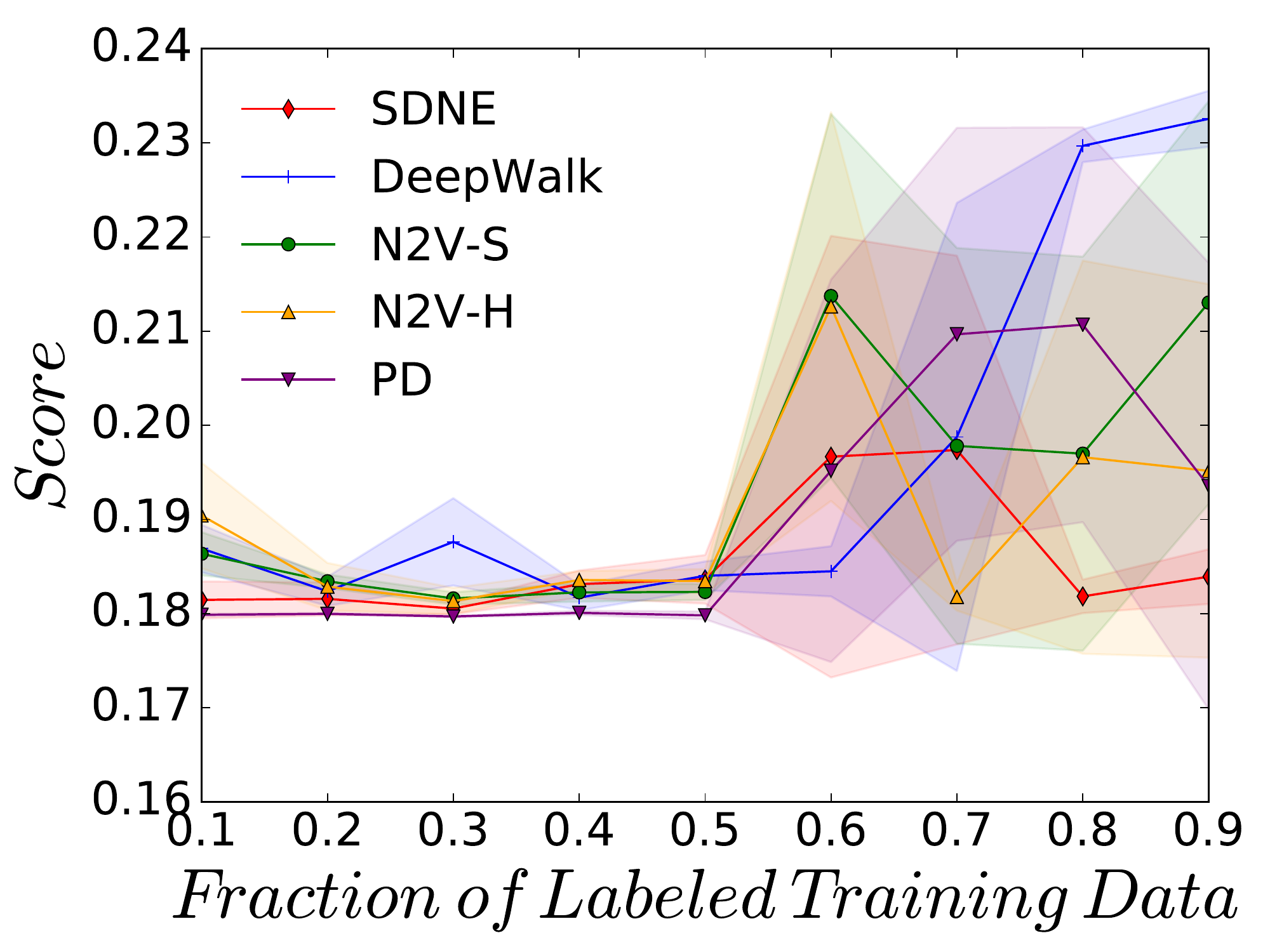}}
  \label{DCmaGN}\hfill
\subfloat[Micro Bitcoinotc]{%
    \includegraphics[width=0.25\linewidth]{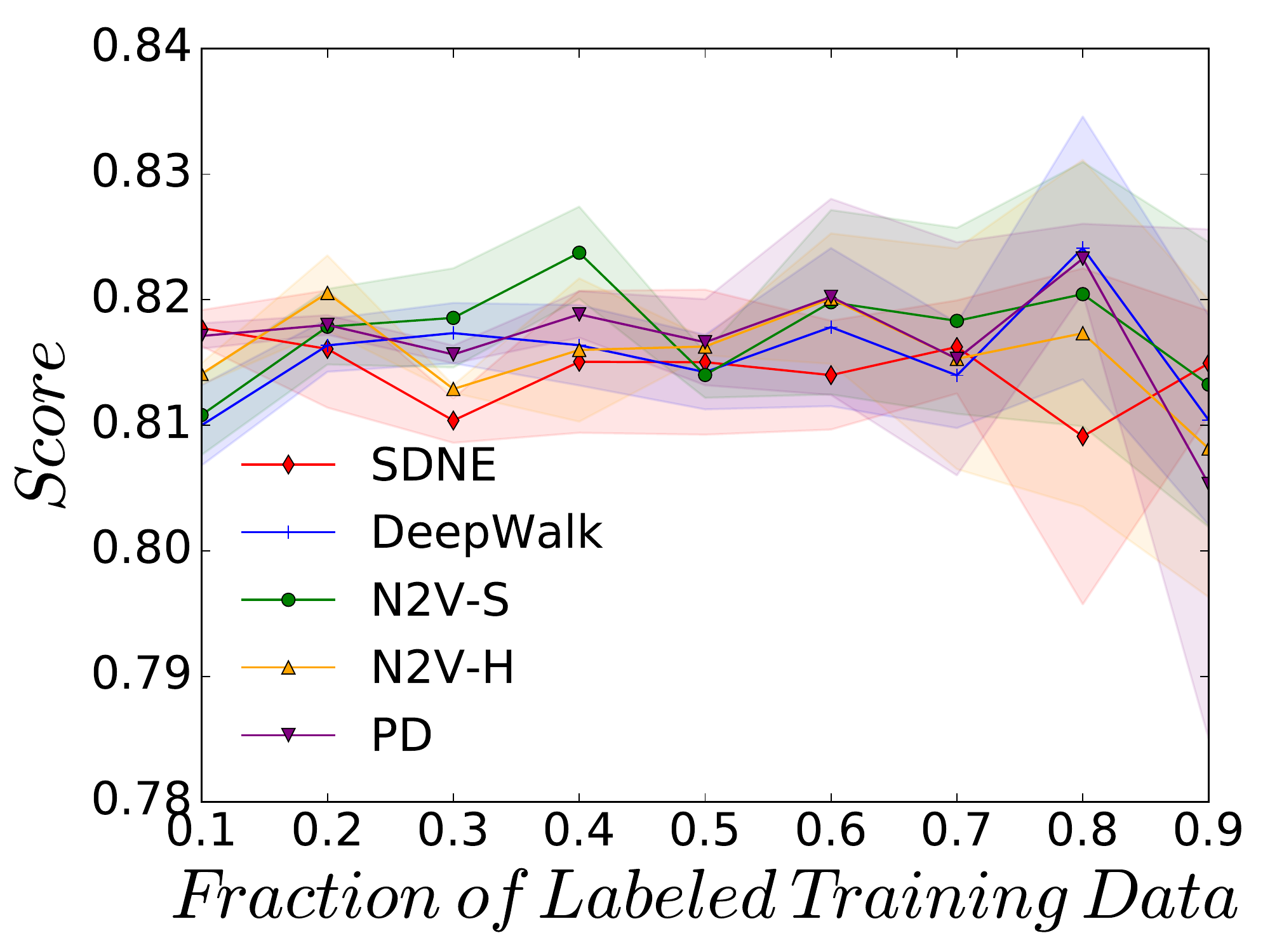}}
  \label{DCmaWI}
\caption{Micro-f1 and Macro-f1 Scores, across a range of labelling fractions, for all approaches when predicting a vertex's PageRank value across all datasets.}
\label{fig:PR_FIG}
\end{figure*}

\begin{figure*}
  \centering
\subfloat[Macro Drosophila]{%
    \includegraphics[width=0.25\linewidth]{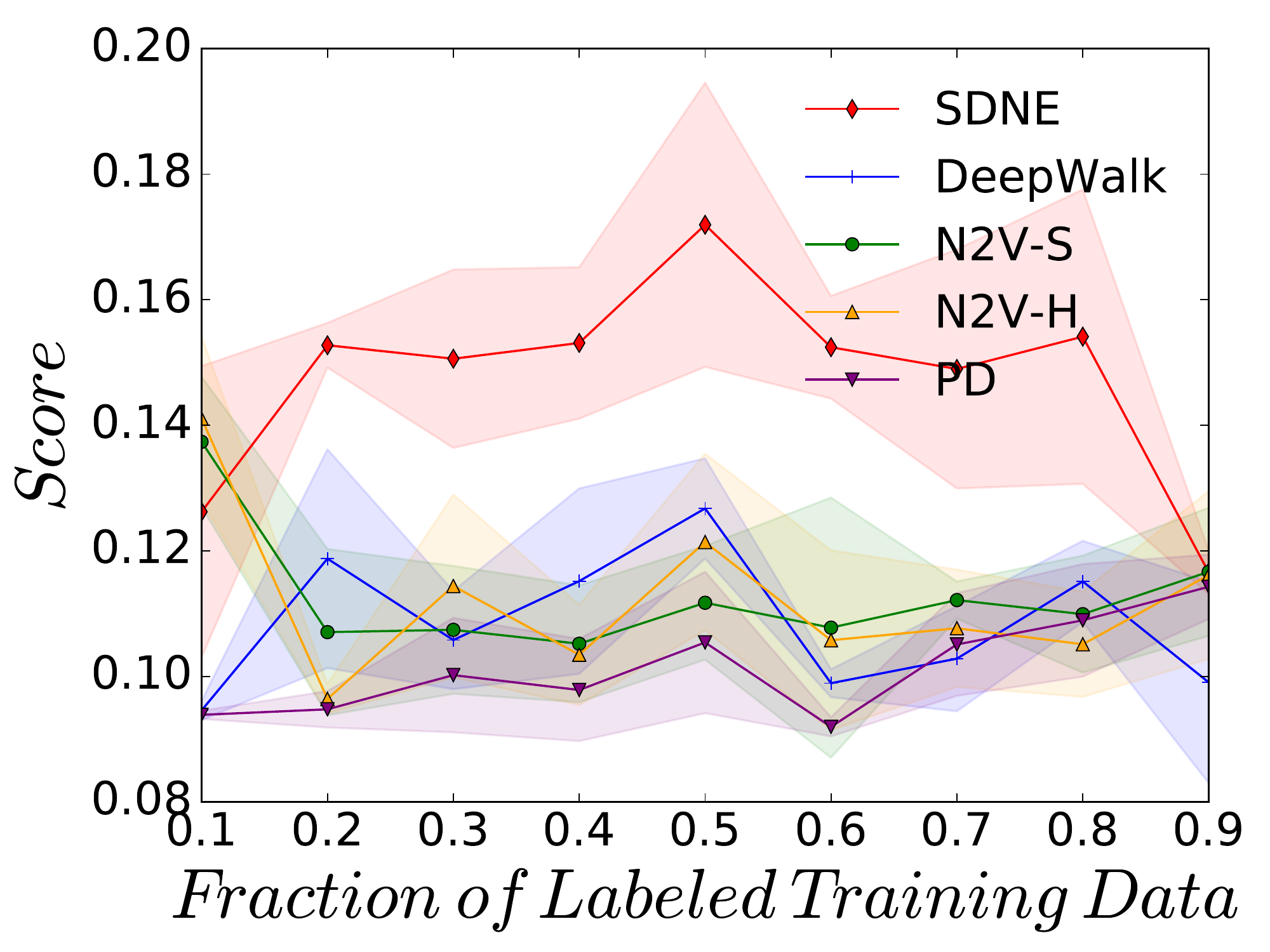}}
  \label{DCmiCA}\hfill
\subfloat[Micro Drosophila]{%
    \includegraphics[width=0.25\linewidth]{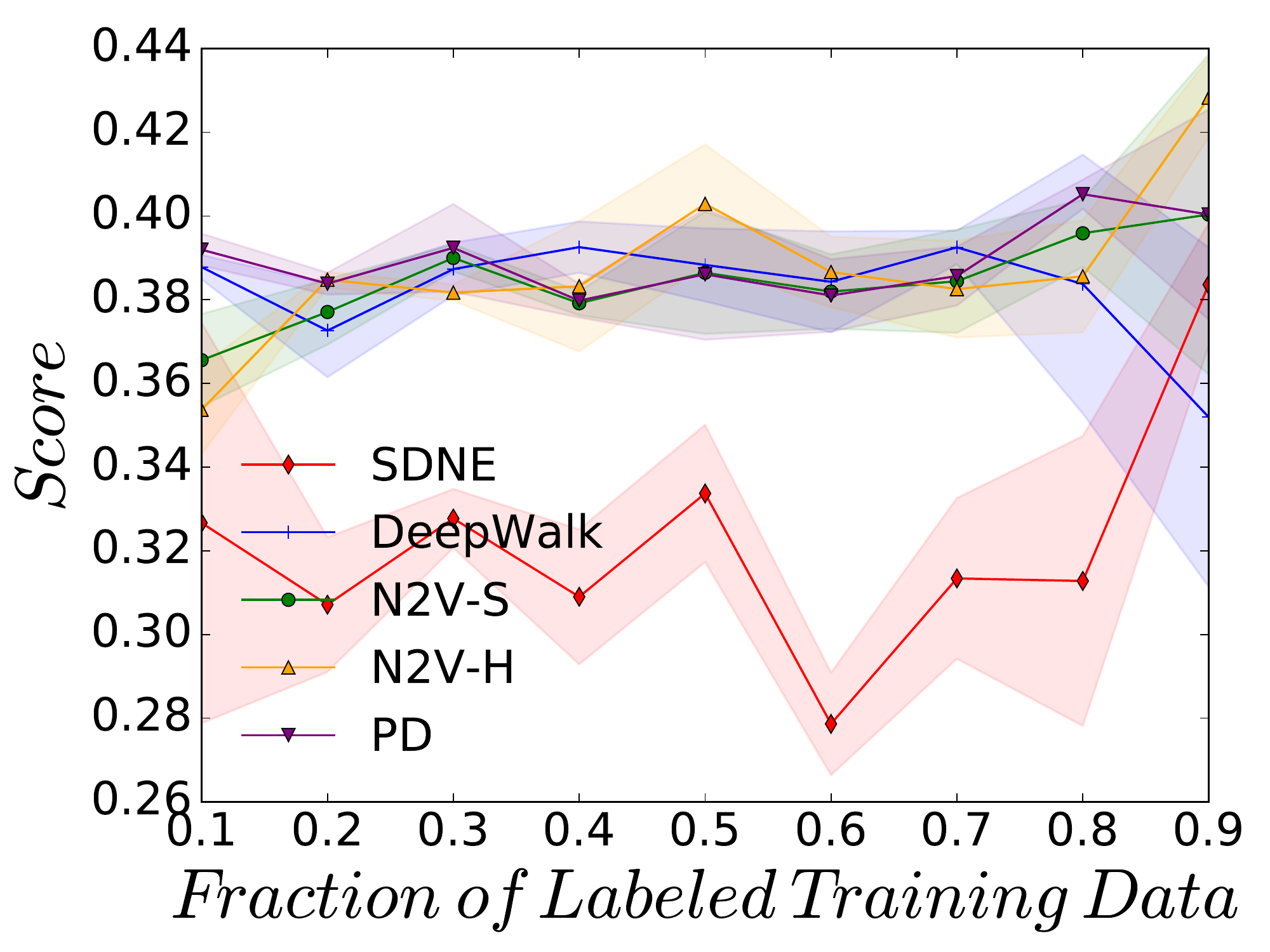}}
  \label{DCmiFB}\hfill
\subfloat[Macro HepTh]{%
    \includegraphics[width=0.25\linewidth]{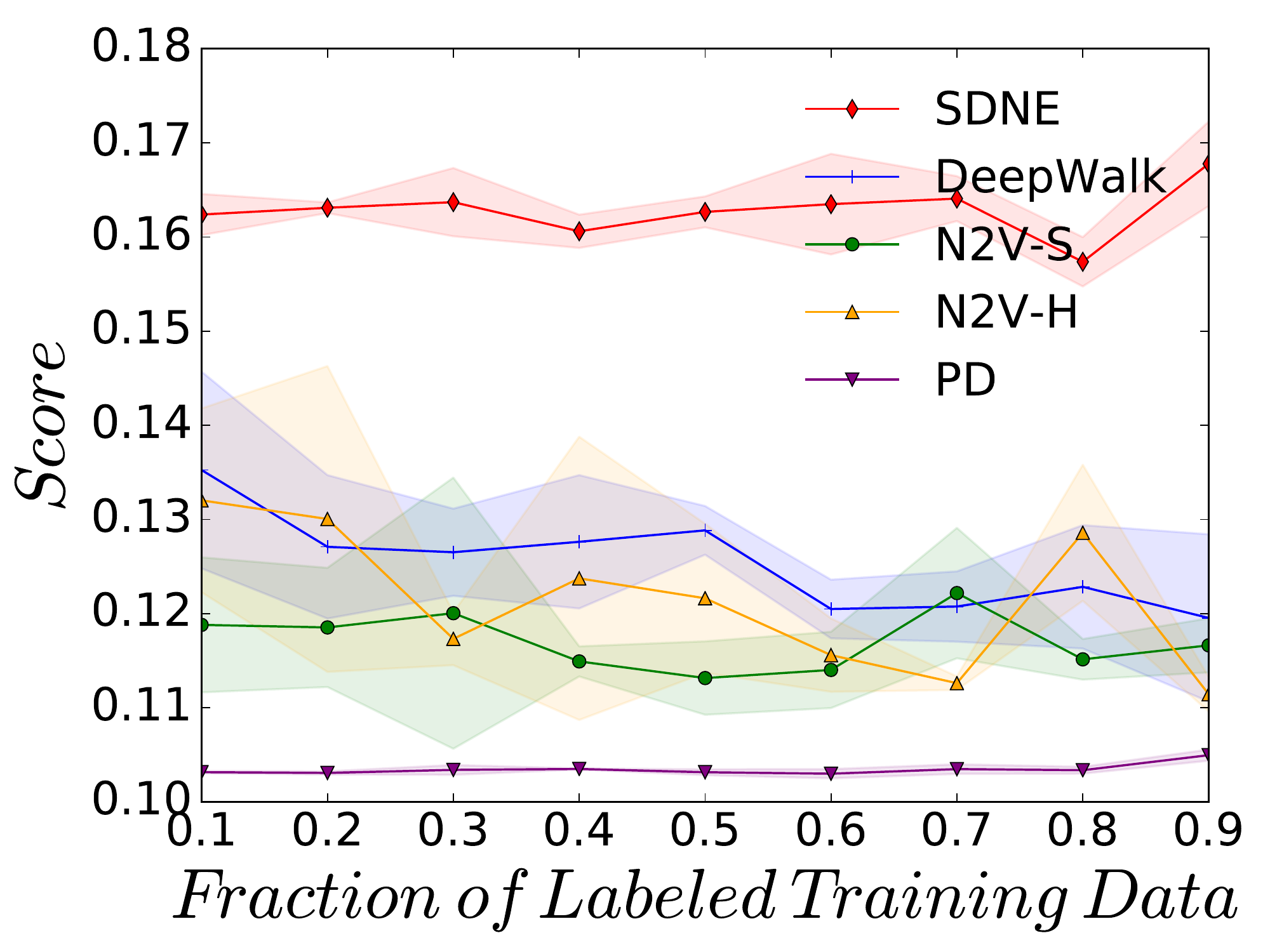}}
  \label{DCmiGN}\hfill
\subfloat[Micro HepTh]{%
    \includegraphics[width=0.25\linewidth]{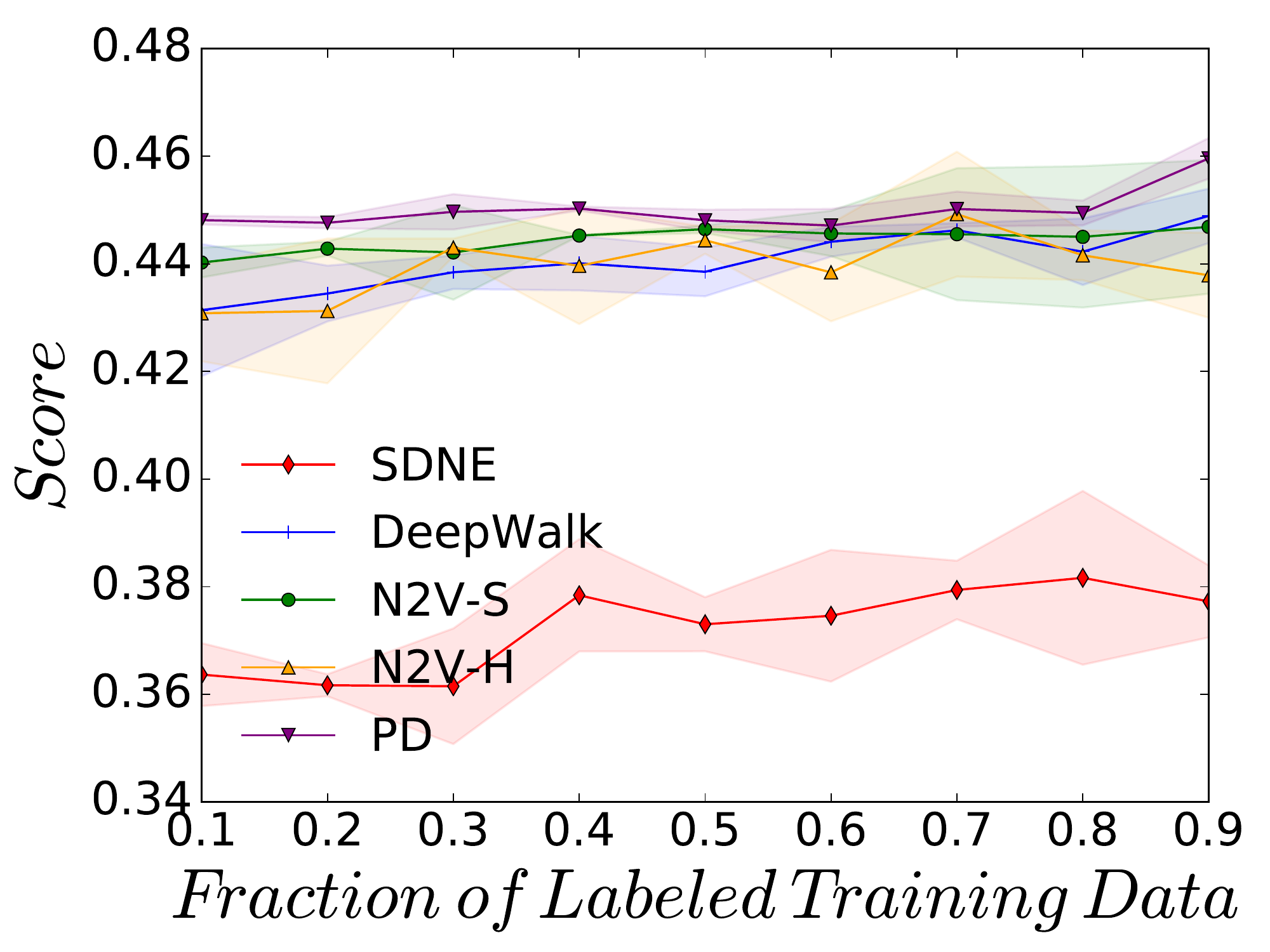}}
  \label{DCmiWI}\\

\subfloat[Macro Email-EU]{%
    \includegraphics[width=0.25\linewidth]{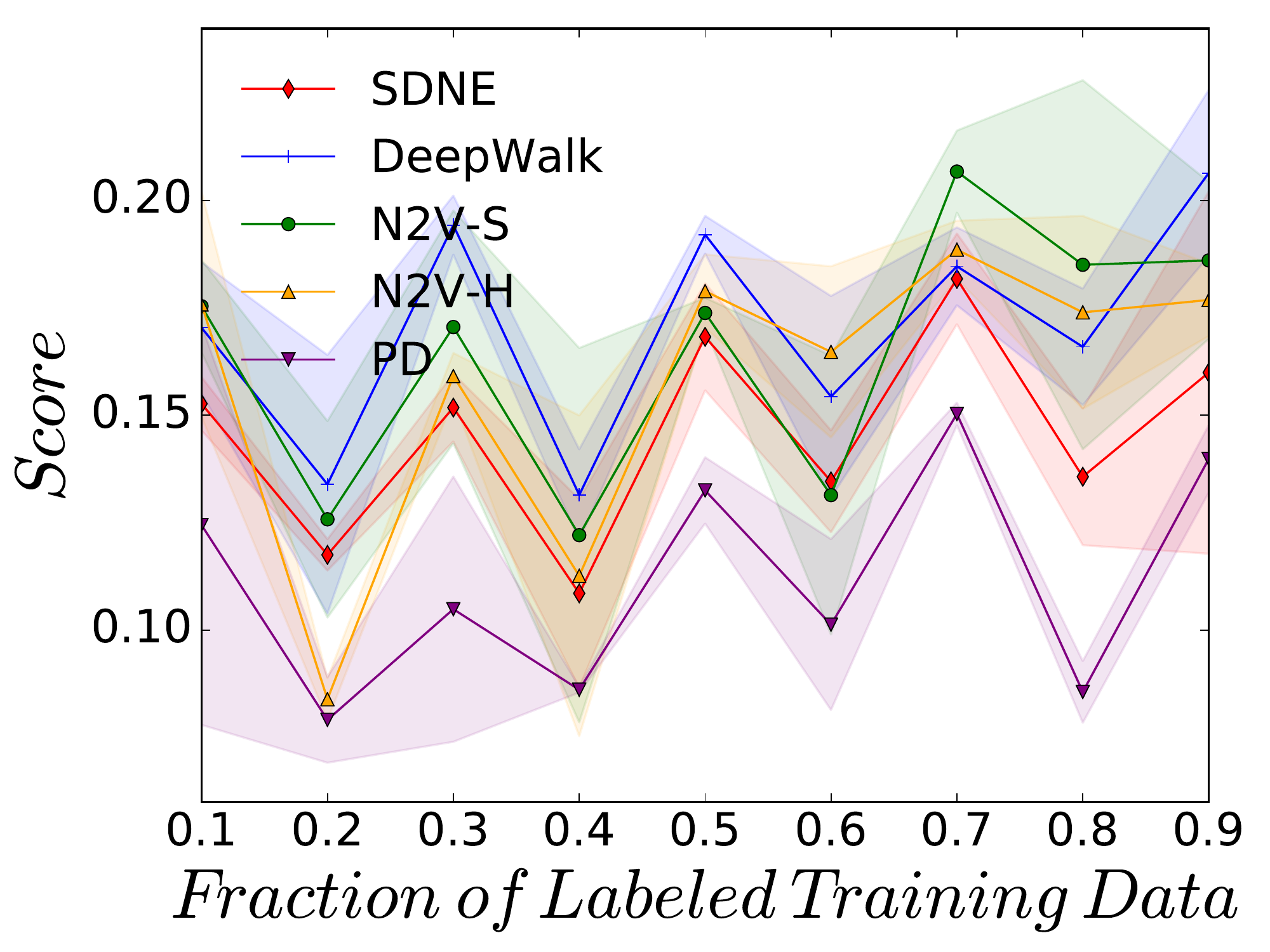}}
  \label{DCmaCA}\hfill
\subfloat[Micro Email-EU]{%
    \includegraphics[width=0.25\linewidth]{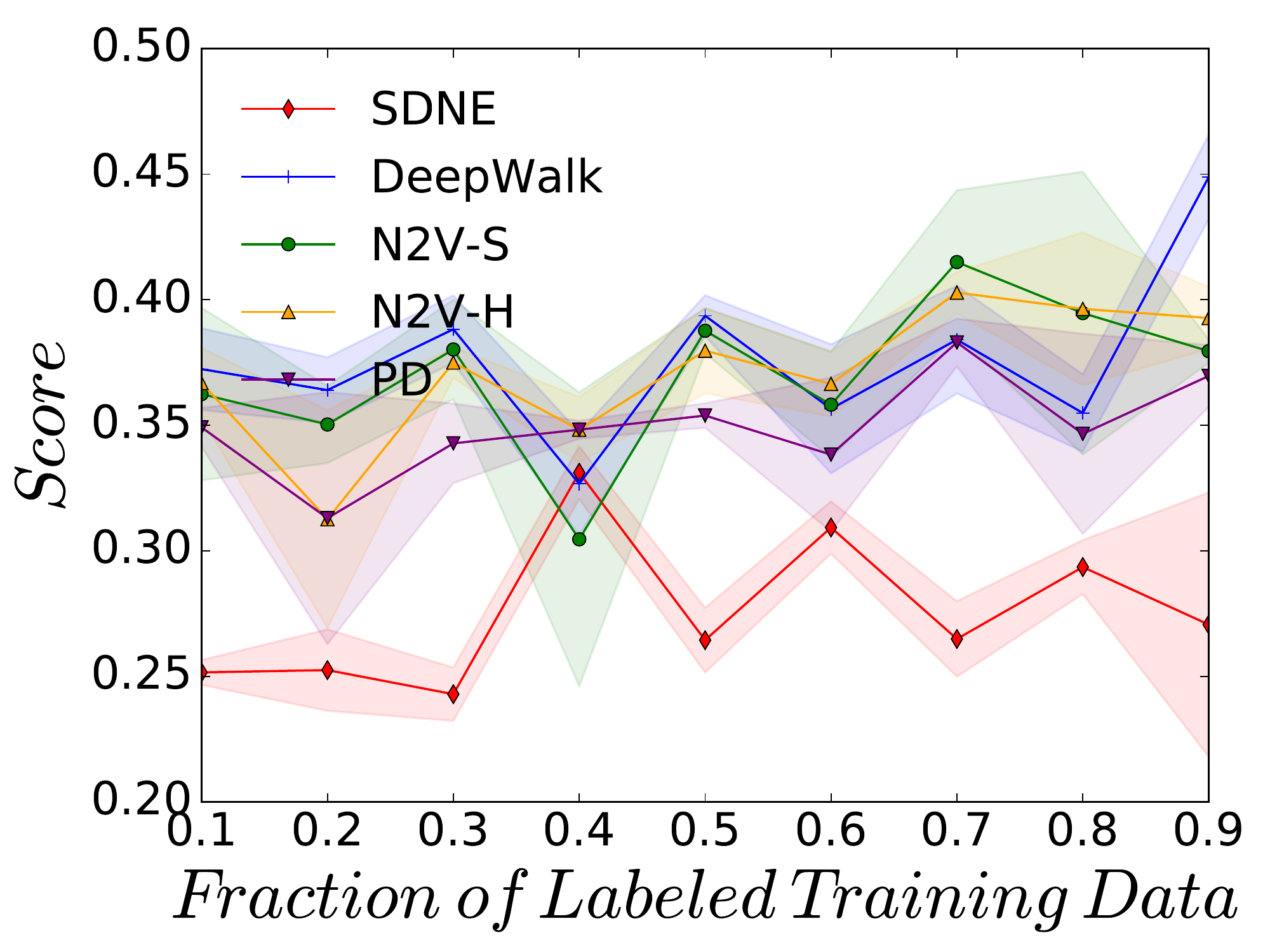}}
  \label{DCmaFB}\hfill
\subfloat[Macro Facebook]{%
    \includegraphics[width=0.25\linewidth]{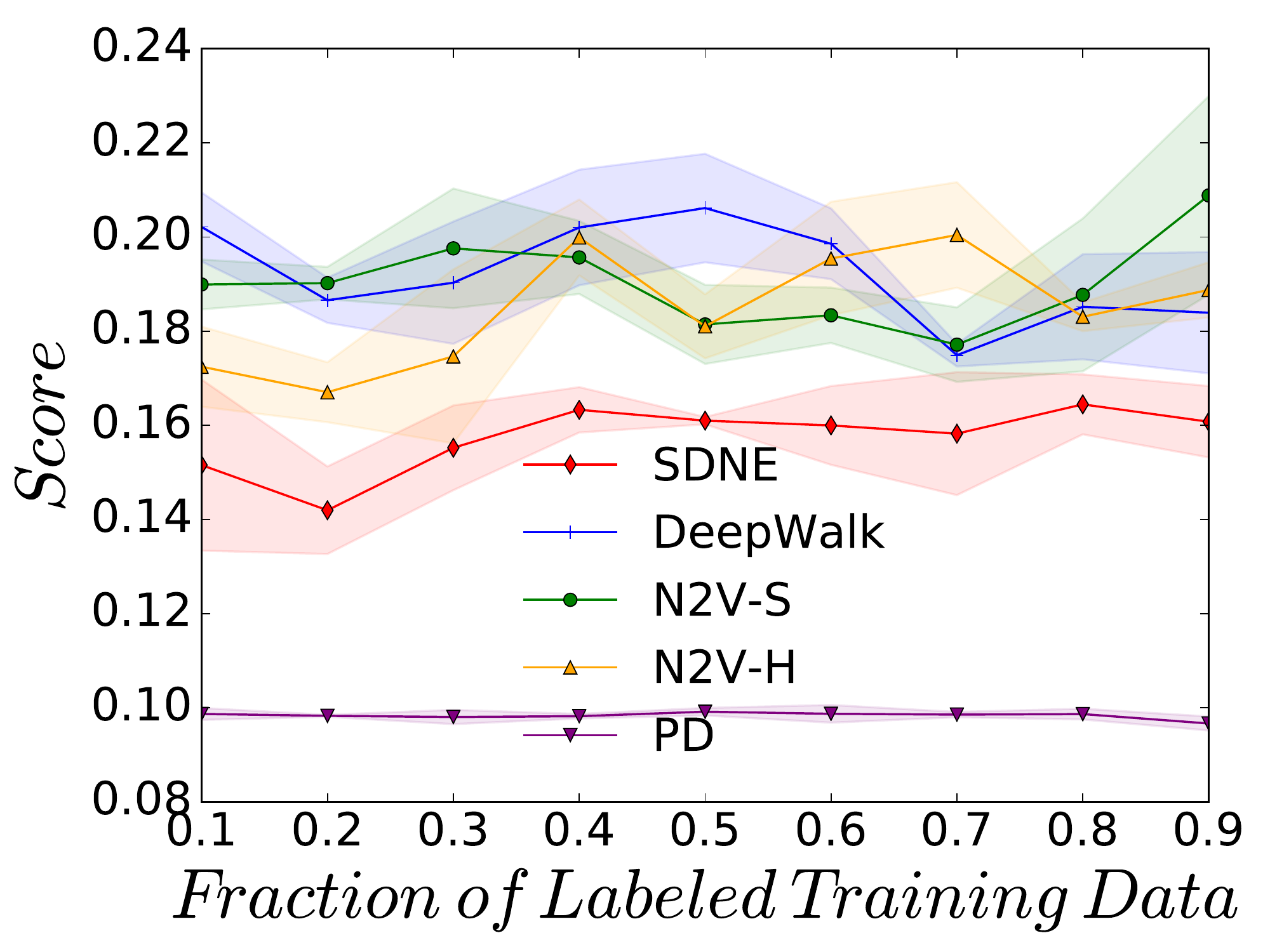}}
  \label{DCmaGN}\hfill
\subfloat[Micro Facebook]{%
    \includegraphics[width=0.25\linewidth]{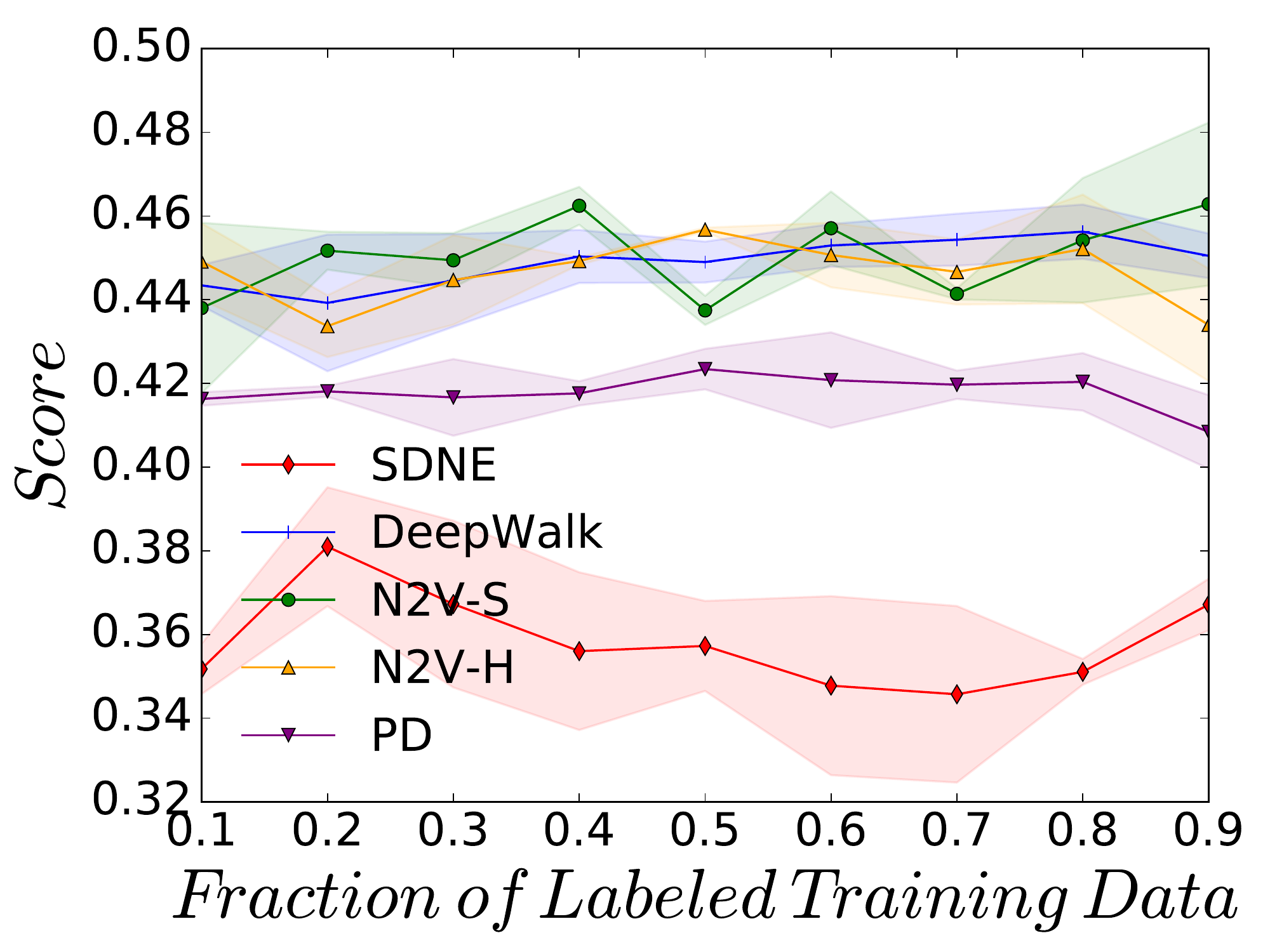}}
  \label{DCmaWI} \\ 

  \subfloat[Macro Openflights]{%
    \includegraphics[width=0.25\linewidth]{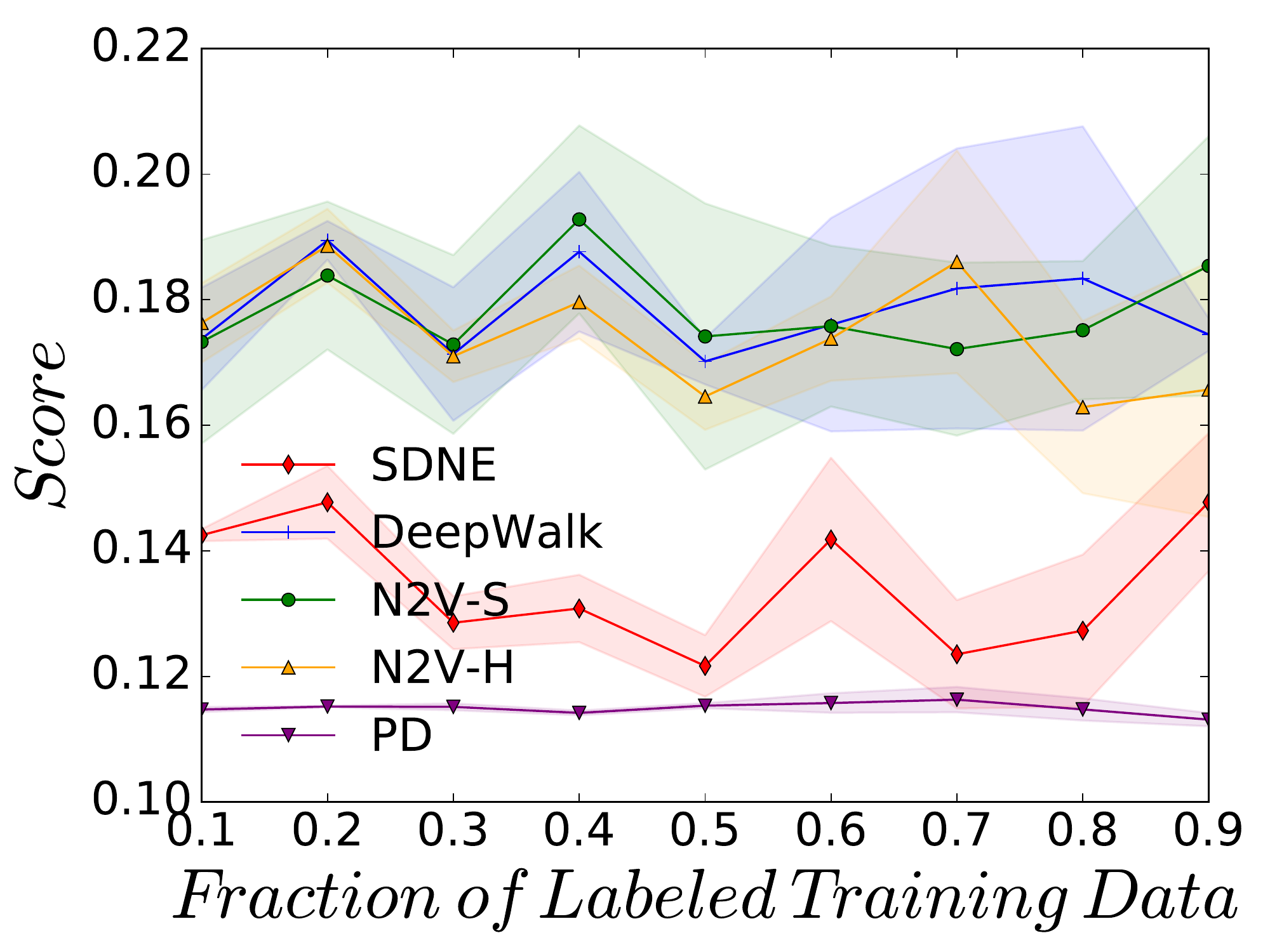}}
  \label{DCmaCA}\hfill
\subfloat[Micro Openflights]{%
    \includegraphics[width=0.25\linewidth]{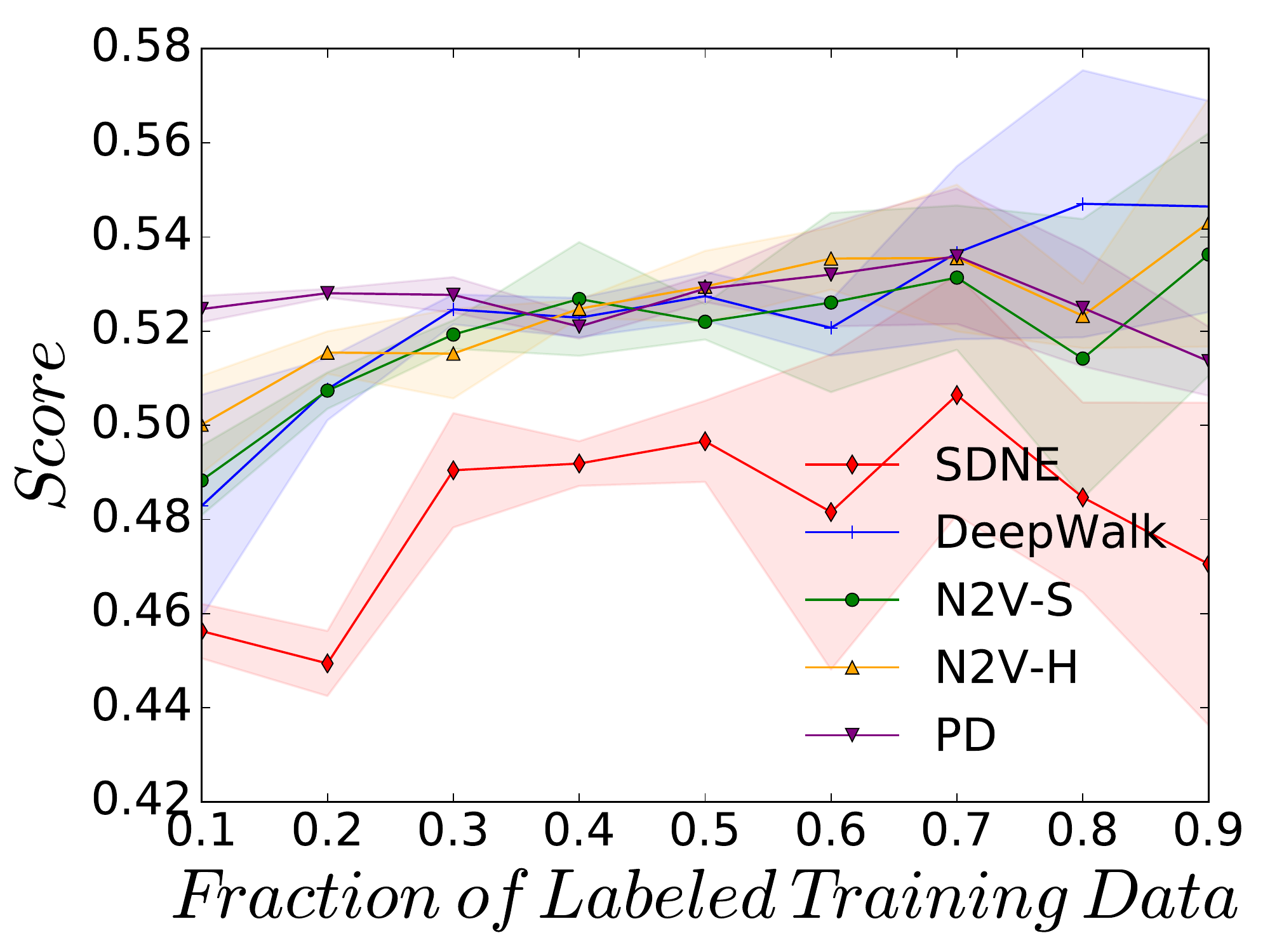}}
  \label{DCmaFB}\hfill
\subfloat[Macro Bitcoinotc]{%
    \includegraphics[width=0.25\linewidth]{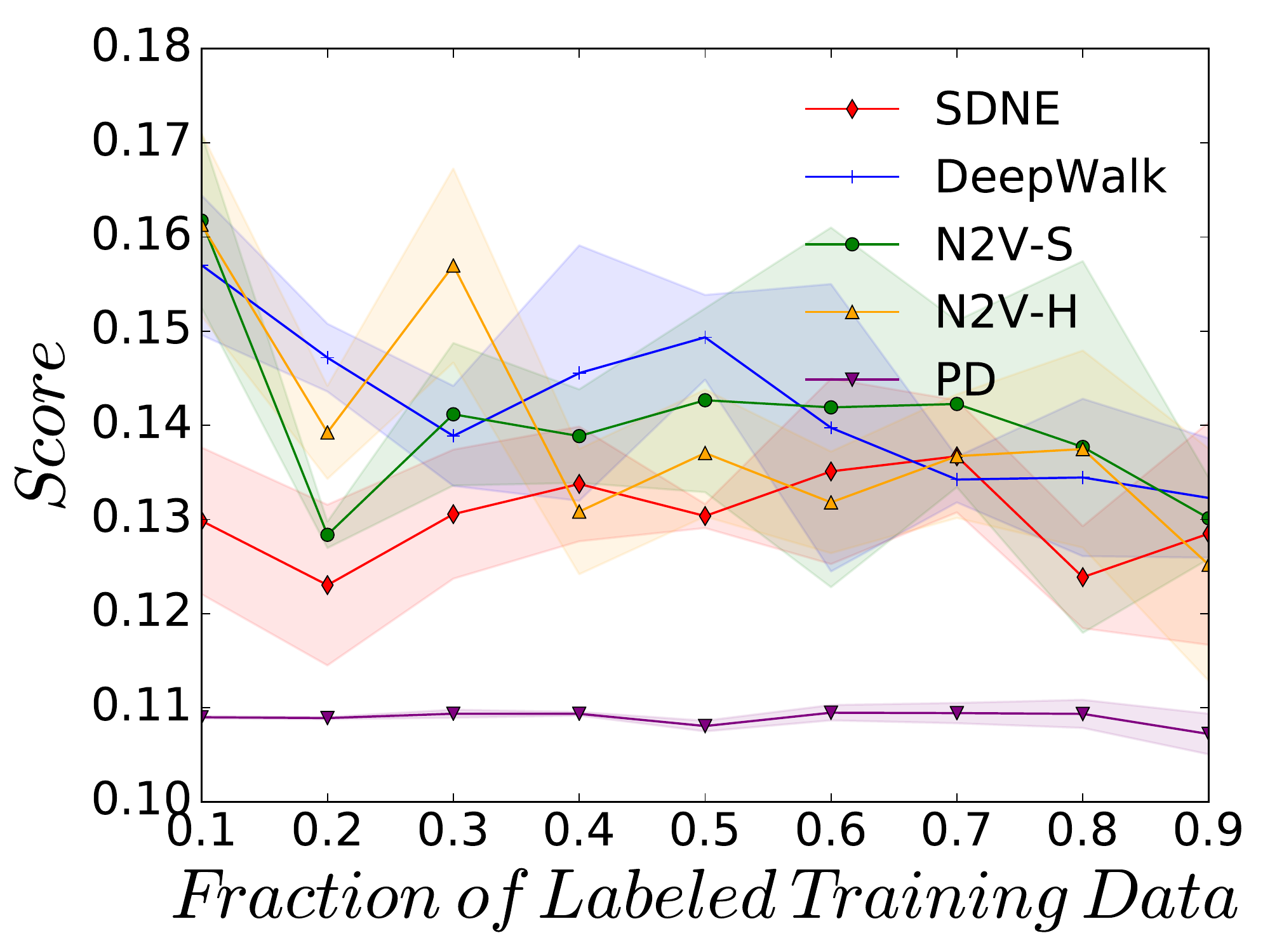}}
  \label{DCmaGN}\hfill
\subfloat[Micro Bitcoinotc]{%
    \includegraphics[width=0.25\linewidth]{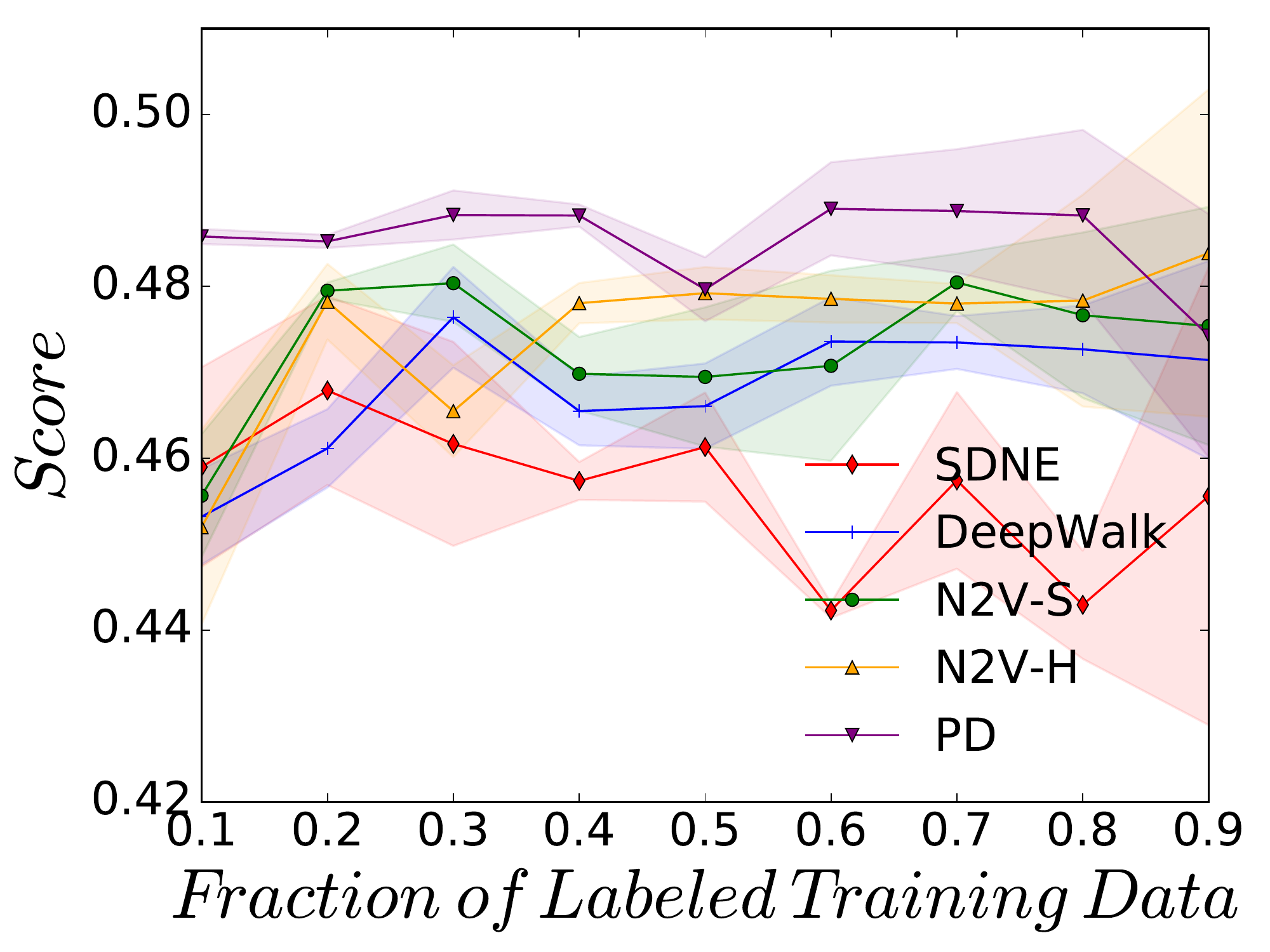}}
  \label{DCmaWI}
\caption{Micro-f1 and Macro-f1 Scores, across a range of labelling fractions, for all approaches when predicting a vertex's Betweenness Centrality value across all datasets.}
\label{fig:BC_FIG}
\end{figure*}

In this section, we present the experimental evaluation of the classification of topological features using the embeddings generated from the five approaches (DeepWalk, Node2Vec-H, Node2Vec-S, SDNE and PD) on the datasets detailed in Table \ref{tab:datasets}. We present both the macro-f1 and micro-f1 scores plotted against a varying amount of labelled data available during the training process. Where a higher score equates to a better classification result -- with a score of one meaning a perfect classification of every example in the data.

Figure \ref{fig:DG_FIG} displays the classification f1 scores for predicting the simplest feature we are measuring: the degree of the vertices. Interestingly we see a large spread of results across the datasets and between approaches, with no clear pattern emerging in this Figure. On certain datasets, it is possible to see a high micro-f1 score, for example in the bitcoinotc dataset, suggesting that an approximation of the degree value is present in the embedding. The figure also shows that SDNE and PD often have a lower score when compared with the stochastic approaches. 

Figure \ref{fig:DC_FIG} highlights the macro-f1 and micro-f1 scores for the classification of the Degree Centrality value. As the Degree Centrality of a given vertex is strongly influenced by it's degree, it is perhaps unsurprising to observe largely similar patterns to those in Figure \ref{fig:DG_FIG}, which again shows the dataset bitcoinotc to be the dataset with the highest accuracies. As was seen in the previous figure, generally the three stochastic approaches have a similar score for both macro-f1 and micro-f1.

The results for the classification of Triangle Counts for the vertices are presented in Figure \ref{fig:TR_FIG}. This is a more complex feature than the previous two, as it requires more information than is available from just the immediate neighbours of a given vertex. The Figure shows again that, to some degree of accuracy, the feature is able to be reconstructed from the embedding space, with bitcoinotc having the highest micro-f1 accuracy of all the datasets. SDNE and PD continue to have, on average, the lowest accuracies. 

Classifying a vertex's local clustering score across the datasets is explored in Figure \ref{fig:CLU_FIG}. The figures shows that this features, although more complicated to compute than a vertices triangle count, appears to be easier for a classifier to reconstruct from the embedding space. With this more complicated feature, some interesting results regrading SDNE can be seen in the Email-EU and HepTh datasets, where the approach has the highest macro-f1 score -- perhaps indicating that the more complex model is better able to learn a good representation for this more complicated feature. 

Figure \ref{fig:EC_FIG} displays the result for the classification of a vertices Eigenvector centrality. This figure is perhaps the most interesting one so far as it shows high classification accuracies across many of the empirical datasets, even though this feature is of greater complexity than previous ones. This figure further supports the results presented in Table \ref{tab:class-dw}, which showed Eigenvector centrality having not only the highest accuracies, but also the highest lifts in accuracy over the rule-based predictors. Interestingly SDNE does not demonstrate higher macro-f1 scores in this experiment.

In Figure \ref{fig:PR_FIG}, the approaches ability to correctly classify the PageRank score of the vertices is considered. Here we see generally lower classification accuracies than the last figure, perhaps owing to the more complicated nature of the PageRank algorithm. Although high classification accuracies can still be seen, particularly on the on the Bitcoinotc and Drosophila datasets.

Finally, Figure \ref{fig:BC_FIG} highlights the ability of the graph embeddings to predict betweenness centrality. Here, the figure shows that this feature is on average, harder to predict from the embeddings than the previous two centrality measures as evidenced by the lower accuracies scores. Again SDNE shows the highest macro-f1 scores on the Drosophila and HepTh datasets, indicating it's embedding capture something akin to this structural information better than the other approaches. 

\subsection{Confusion Matrices}

One consideration that must be made is that the binning process, used to transform the features into targets for classification, removes the inherent ordering present in continuous values. As an example, a vertex with a degree of 8 would still be classified incorrectly if the prediction was 10 or 100, but clearly one is more incorrect than the other. To address this, we present a selection of error matrices, to explore how `wrong' an incorrect prediction is. This is made possible as the labels used for classification have consecutive ordering, as a result of a histogram binning function, meaning that a prediction of 2 for a true label of 1, is more correct than a prediction of 5. 

\begin{figure*}[!h]
  \centering
  \subfloat[SDNE]{%
  \includegraphics[width=0.25\linewidth]{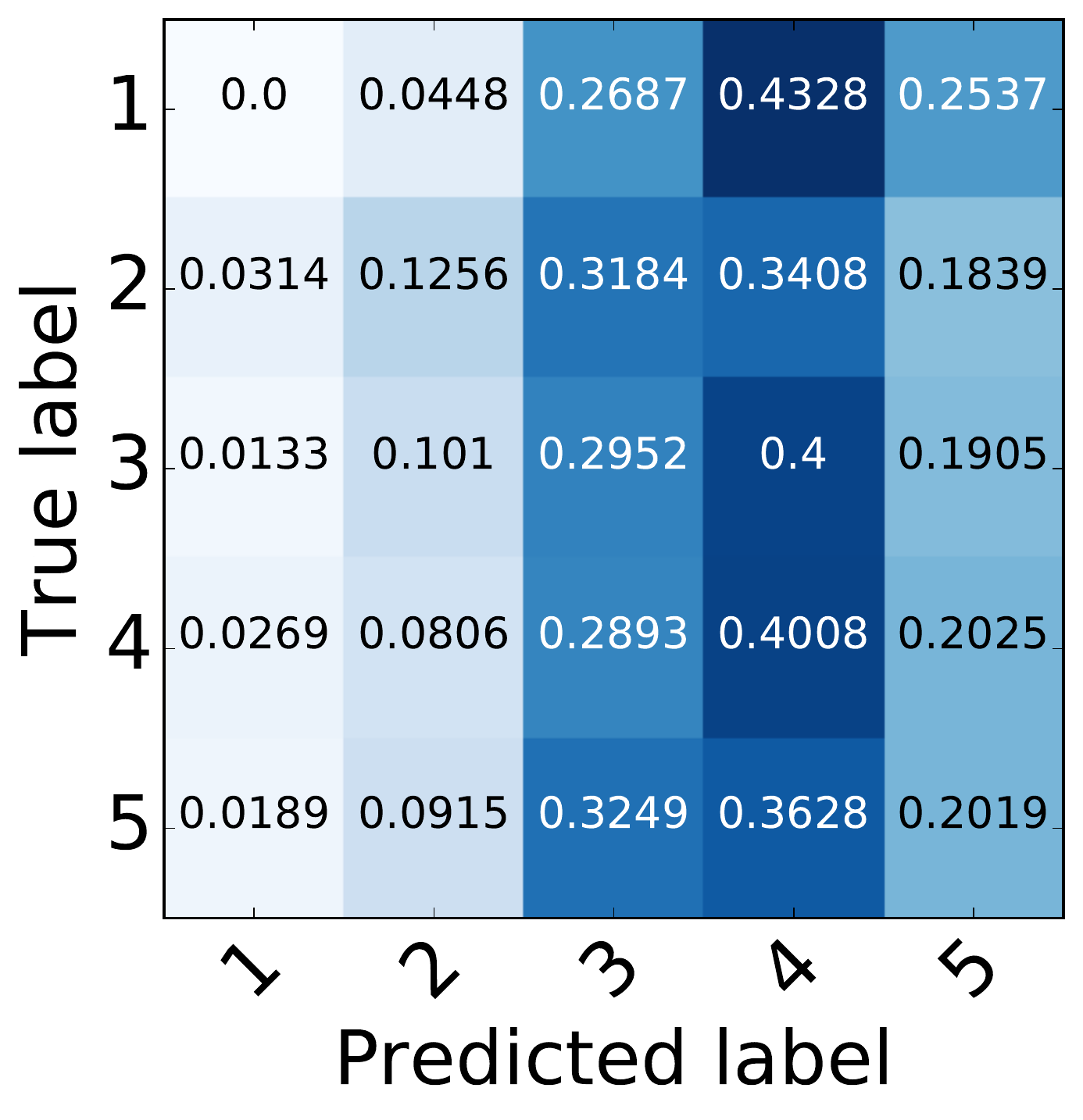}}
\label{SDNE-EM}\hfill
\subfloat[DW]{%
    \includegraphics[width=0.25\linewidth]{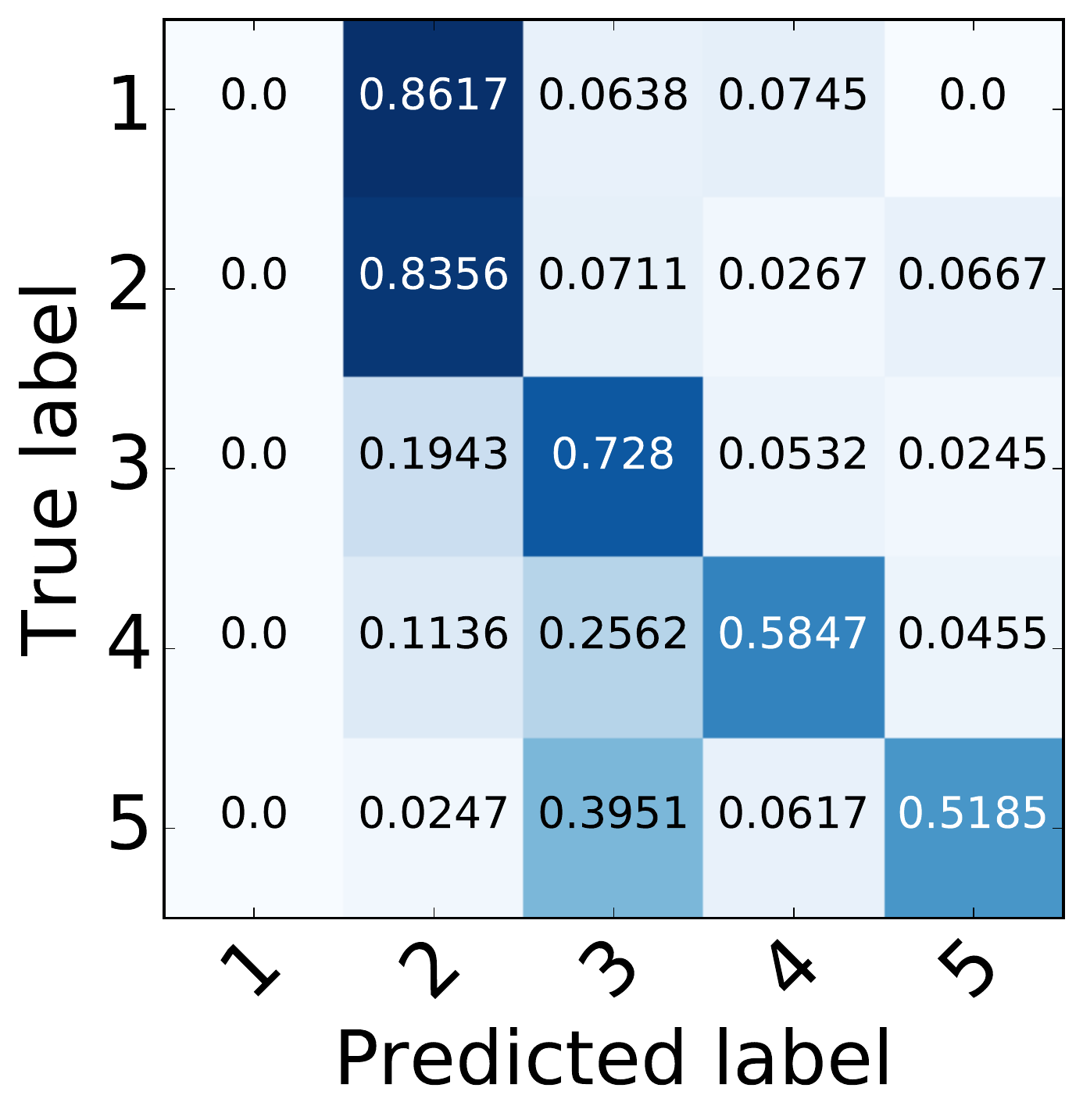}}
  \label{DW-EM}\hfill
\subfloat[N2V-H]{%
    \includegraphics[width=0.25\linewidth]{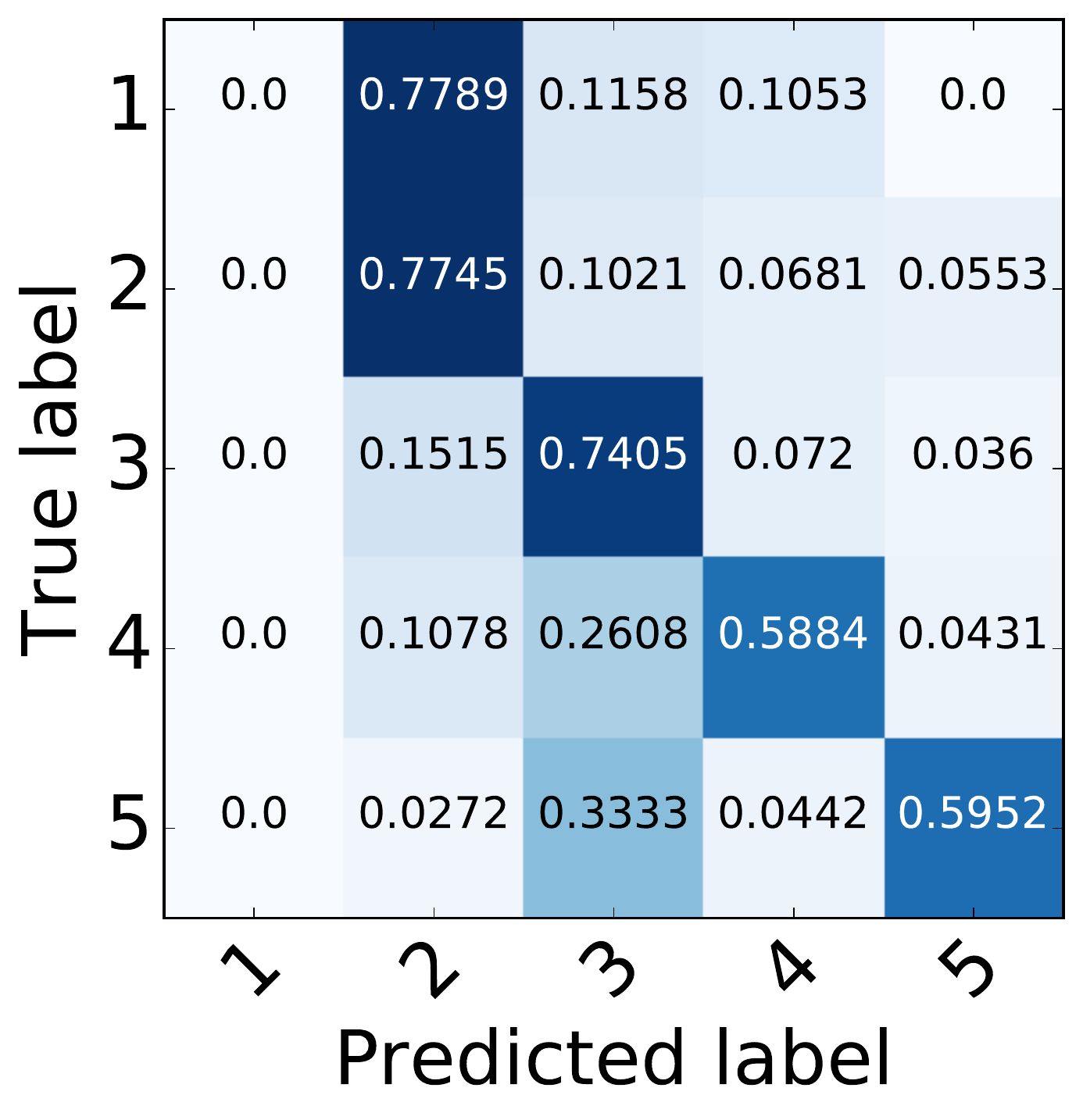}}
  \label{N2VH-EM}\hfill
  \subfloat[N2V-S]{%
    \includegraphics[width=0.25\linewidth]{figs/EM-facebook_combined-node2vec-stru.pdf}}
  \label{TRmiFB}\hfill
\caption{Error matrices for neural network classification of Eigenvector Centrality for the ego-Facebook dataset.}
\label{fig:EM}
\end{figure*}

For brevity, Figure \ref{fig:EM} displays the error matrices for a selection of the tested embedding approaches when classifying Eigenvector Centrality in the ego-Facebook dataset, although similar patterns were found across all datasets. With error matrices, the diagonal values represent correctly classified label, thus a good prediction will produce an error matrix with a higher concentration of diagonal values. Figure \ref{fig:EM} shows that, for the stochastic walk approaches DeepWalk and Node2Vec, the error matrices have a higher clustering of values around the diagonals. Interestingly, when the classification is incorrect for these approaches, the incorrect prediction tends to be close to the true label. This phenomenon can clearly be seen in these approaches for labels 1 and 2, meaning that embeddings for vertices with these particularly Eigenvector Centrality are similar. The Figure also shows that, for this particular vertex feature, the embeddings produced via SDNE seemingly do not contain the same topological information. This is highlighted by the lack of structure on the diagonals of it's error matrix. 

\subsection{Unsupervised Low-Dimensional Projections}
\label{sec:unsuperisedproj}

Another way to explore assessing the semantic content of the graph embeddings, we utilised an embedding visualisation technique entitled t-SNE \cite{maaten2008visualizing}. This techniques allows relatively high dimensional data, such as the graph embeddings we are dealing with, to be projected into a low dimensional space in such a way as to preserve the inter-spatial between points that were present in the original space. Thus, we utilise t-SNE to project the embeddings down into two dimensions so they can be easily visualised. This process is performed without the need for any classification to be performed upon the embeddings, removing the problems of classifying unbalanced datasets.

\begin{figure*}[!h]
  \centering
  \subfloat[SDNE]{%
  \includegraphics[width=0.25\linewidth]{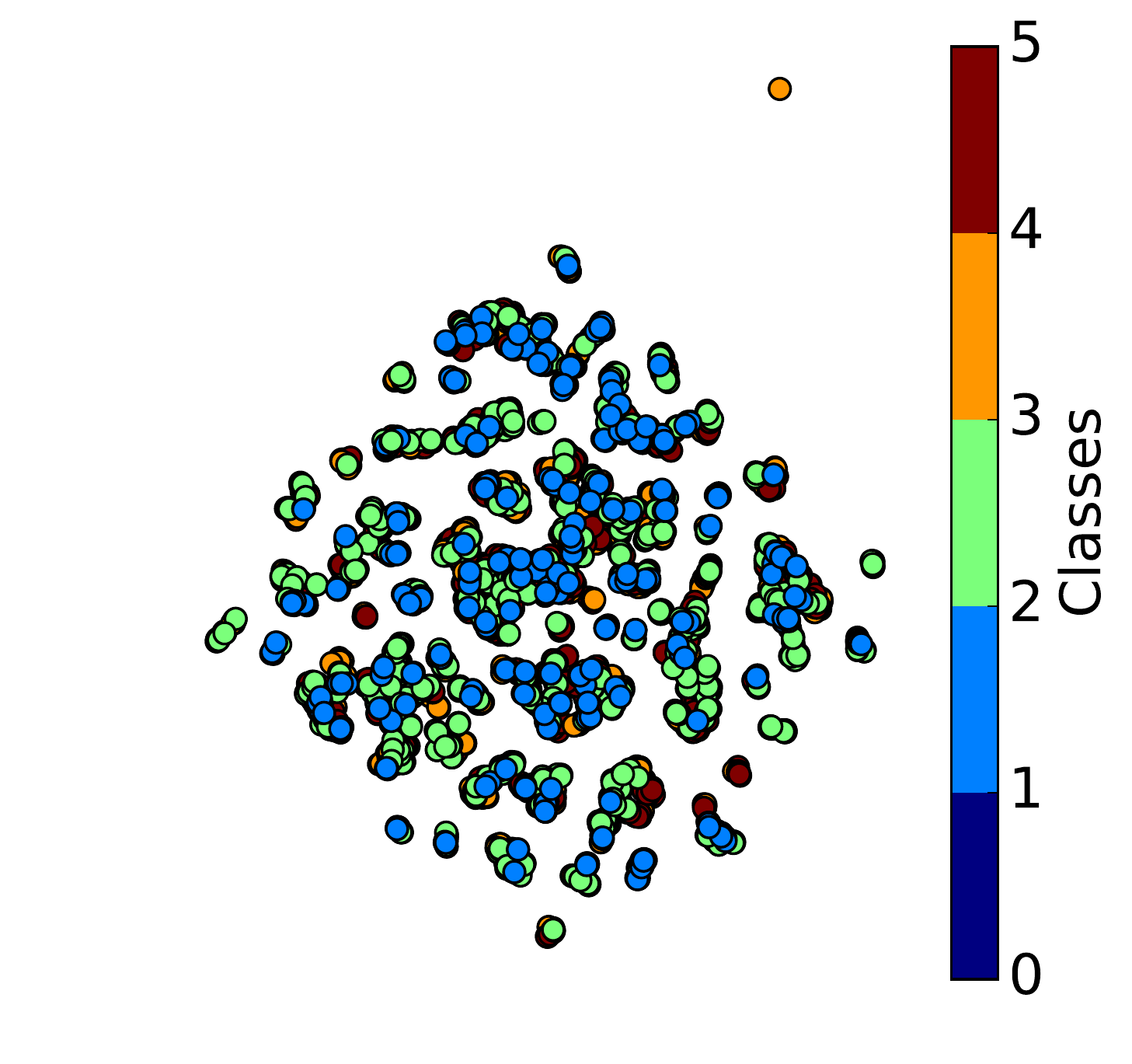}}
\label{SDNE-tsne}\hfill
\subfloat[DW]{%
    \includegraphics[width=0.25\linewidth]{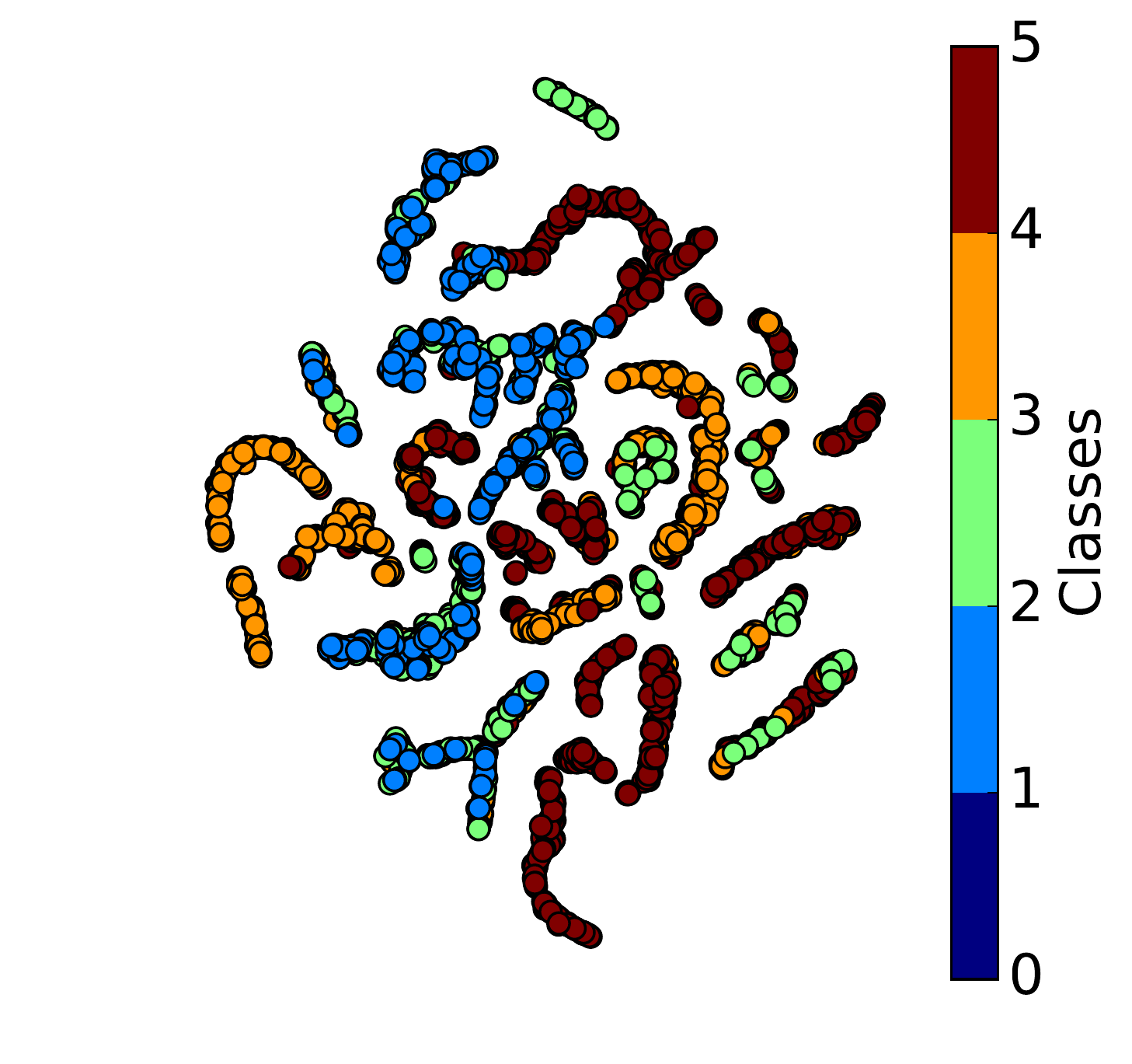}}
  \label{DW-tsne}\hfill
\subfloat[N2V-H]{%
    \includegraphics[width=0.25\linewidth]{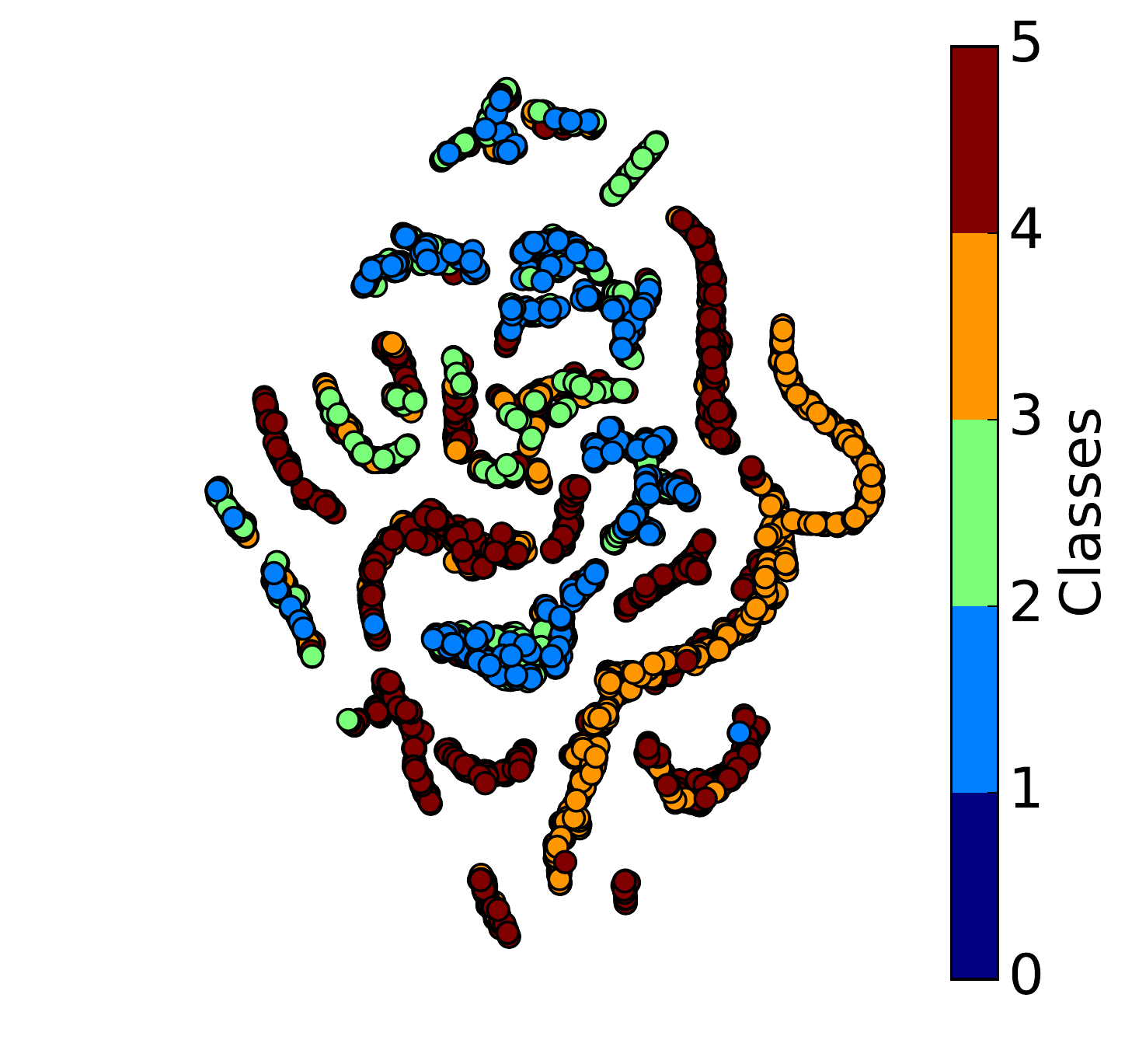}}
  \label{N2VH-tsne}\hfill
  \subfloat[N2V-S]{%
    \includegraphics[width=0.25\linewidth]{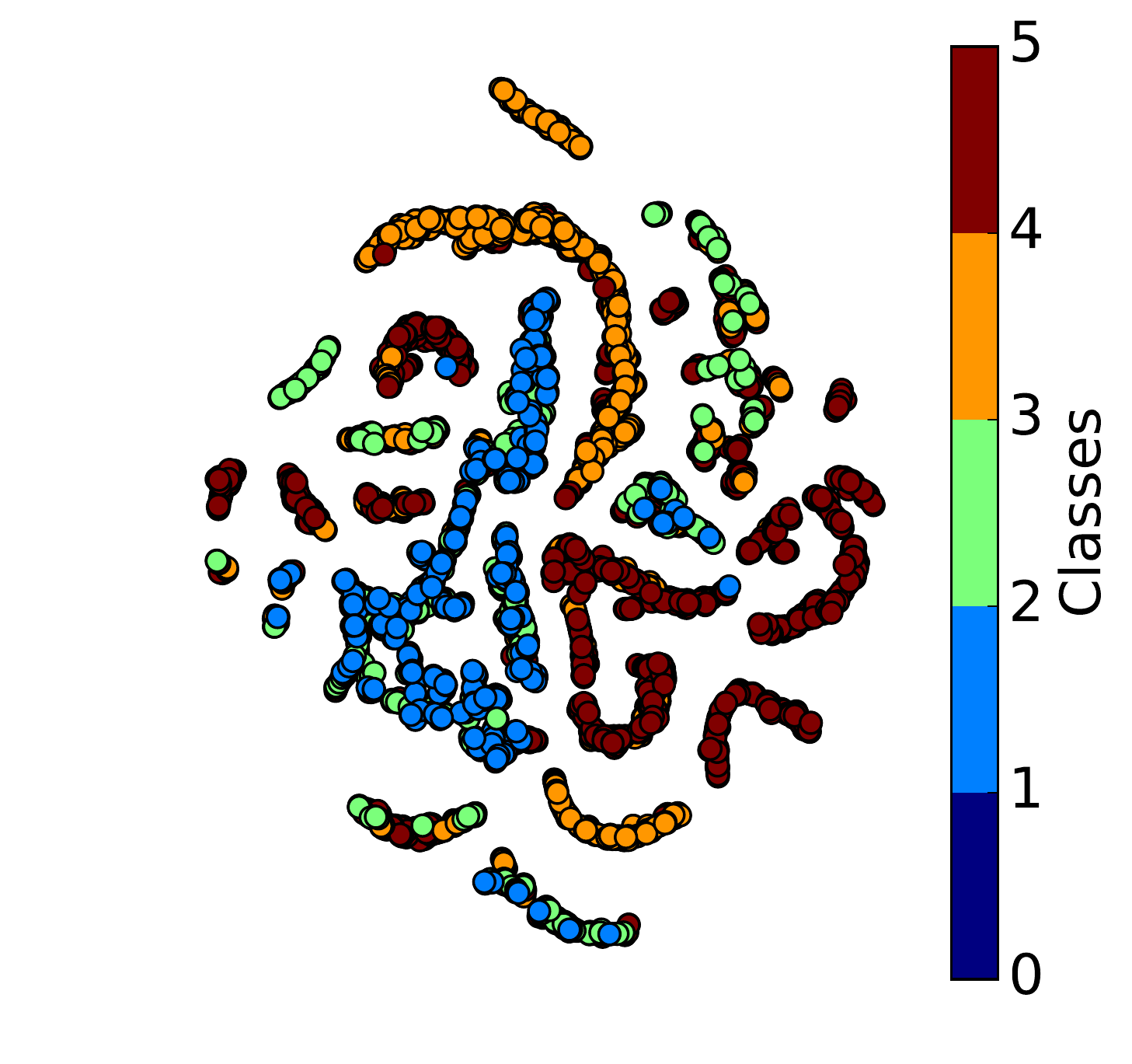}}
  \label{N2VS-tsne}\hfill
\caption{t-SNE plots of the embeddings taken from the ego-Facebook dataset, where the points are coloured according to their Eigenvector Centrality}
\label{fig:tsne}
\end{figure*}

Figure \ref{fig:tsne} displays a selection of t-SNE plots taken from the ego-Facebook data, where the points are coloured according to the Eigenvector centrality value after being passed through the binning process. The figure shows that the SDNE embeddings seemingly have no clear structure in the low dimensional space which correlates strongly with the Eigenvector centrality, as points in the same class are not clustered together. However, with the other embedding approaches, it is possible to see a clear clustering of points belonging to  the same class. For example, in both the Node2Vec approaches, there is very clear clustering of classes 1, 4 and 5. This result provides further evidence for our observation that, even when exploring the embeddings using an unsupervised method, it is possible to find correlations between known topological features and the embedding space.

\subsection{Discussion}


This section has provided extensive experimentation evaluation to explore the questions raised in Section \ref{sec:EEQ}. Specifically, we investigated if a broad range of topological features can be predicted from the embedding created from a range of unsupervised graph embedding techniques. Across all the features and datasets tested, it can be seen that many topological features can be approximated by the different embedding approaches, with varying degrees of accuracy. The results which show the increase in accuracy over the rule based predictions (Section \ref{sec:classalgsec}) give strong indication that the approaches are able to overcome the inherent unbalanced nature of graph datasets and a mapping from the embedding space to features is happening. It is also interesting to observe that numerous features can be approximated from the graph embeddings, suggesting that several structural properties are being captured to create the best representation for a vertex automatically. Of all the topological features measured in the experimentation section, the one which consistently gave the best results was Eigenvector centrality. Particularly for the stochastic approaches, Eigenvector centrality was predicted with a high degree of accuracy, suggesting that the topological structure represented by this feature is captured extremely well in the embedding space and indicates this is a useful feature for the minimising the objective functions of the approaches. This is further reinforced by the unsupervised projections (Figure \ref{fig:tsne}), which shows clear and distinct clustering between classes, even without the use of a classification algorithm.

Another interesting observation from this study is that no one approach strongly out performs the other when classifying a particular feature -- seemingly all the approaches are approximating similar topological structures. The figures show that the stochastic approaches (DeepWalk and Node2Vec) are the most consistent across all features and datasets, often having the highest macro-f1 and micro-f1 scores. SDNE demonstrates a more inconsistent performance profile for feature classification, this is in contrast to other studies which have found it to have the best performance in vertex labelling problems \cite{Goyal2017}. The performance of SDNE demonstrated in this work could be explained by it being the only deep model tested, meaning that it contains many more parameters. This increase in complexity means that SDNE could be very sensitive to the correct selection of hyper-parameters or possibly that more complex topological features are being approximated by the embeddings, or even that entirely novel features are being learned. Finally, it is interesting to note the performance of Hyperbolic approach PD, as it has far fewer latent dimensions in which to capture topological information due to its limitation in modelling the space as a 2D disk. Empirically, PD shows largely similar performance to the other approaches on most datasets, providing strong evidence that the hyperbolic space is an appropriate space in which to represent graphs.

\section{Conclusion}
\label{sec:C}

Graph embeddings are increasingly becoming a key tool to solve numerous tasks within the field of graph mining. They have demonstrated state-of-the-art results by reporting to automatically learn a low dimensional, but highly expressive, representation of vertices, which captures the topological structure of the graph. However to date, there has been little work providing a theoretical grounding detailing why they have been so successful. In this paper, we explore making a step in this direction by investigating which traditional topological graph features can be reconstructed from the embedding space. The hypothesis being that if a mapping from the embedding space to a particular topological feature can be found, then the topological structure encapsulated by this feature is also captured by the embedding. We present an extensive set of experiments exploring this issue across five unsupervised graph embedding techniques (detailed in Section \ref{sec:approachescompared}), classifying seven graph features (detailed in Section \ref{sec:topoligicalfeatures}), across a range of empirical datasets (detailed in Table \ref{tab:datasets}). We find that a mapping from many topological features to the embedding space of the tested approaches is indeed possible, using both supervised and unsupervised techniques. This discovery suggests that graph embeddings are indeed learning approximations of known topological features, with our experiments showing that Eigenvector centrality is best reconstructed by many of the approaches. This could allow key insight into how graph embedding learn to create high quality representations.

For future research, we plan to see if other Eigenvector based topological features, know to be representative of a graph's topology \cite{Li2012}, are also captured as well by the embedding approaches. We plan to perform more experimentation with synthetically created graphs with artificially balanced degree distributions. This will remove the unbalanced nature of empirical datasets, and allow us to explore the structure of the embeddings in more detail. Further more, we plan to investigate if directly predicting topological features during the embedding training process, perhaps in the form of a regularisation term, can produce embeddings which generalise better across other tasks.

\subsubsection*{Acknowledgments}

We gratefully acknowledge the support of NVIDIA Corporation with the donation of the Tesla K40 GPU used for this research. Additionally we thank the Engineering and Physical Sciences Research Council UK (EPSRC) for funding. We also thank the authors of papers \cite{Grover2016}, \cite{Chamberlain2017} and \cite{Wang2016a} for making implementations of their code publicly available. For invaluable feedback and comments during this research, we also thank Nik Khadijah Nik Aznan, Philip Jackson and Amir Atapour-Abarghouei.

\bibliographystyle{plain}
\bibliography{ref}

\end{document}